\documentclass[9pt]{article}

\usepackage[utf8]{inputenc}
\usepackage[small,compact]{titlesec}
\usepackage{amssymb, framed}
\usepackage{amscd}
\usepackage{graphicx, hyperref}
\usepackage{amsmath}
\usepackage{mathtools}
\usepackage[math]{cellspace}
\usepackage[top=1in,bottom=1in,left=1in,right=1in]{geometry}
\usepackage{moreverb}
\usepackage{epstopdf}
\usepackage{indentfirst}
\usepackage{xcolor,colortbl}
\usepackage{array}
\usepackage{xspace}
\usepackage{bbm}
\usepackage[normalem]{ulem}
\usepackage{multirow}
\usepackage{ulem}
\usepackage{cleveref}
\usepackage{multibib}
\usepackage{multicol}
\usepackage{siunitx}
\usepackage[font=small,labelfont=bf,tableposition=top]{caption}
\usepackage{soul}
\usepackage[nonumberlist]{glossaries}
\usepackage{wrapfig}
\usepackage[ruled,boxed]{algorithm}
 \usepackage{algpseudocode}
 \usepackage{algorithmicx}
 \usepackage{diagbox}

\floatname{algorithm}{Algorithm}
\algrenewcommand\algorithmicrequire{\textbf{Input:}}
\algrenewcommand\algorithmicensure{\textbf{Output:}}

\newenvironment{mat}{\left[\begin{array}{ccccccccccccccc}}{\end{array}\right]}
\newcommand\bcm{\begin{mat}}
\newcommand\ecm{\end{mat}}

\usepackage{tikz}

\newenvironment{rmat}{\left[\begin{array}{rrrrrrrrrrrrr}}{\end{array}\right]}
\newcommand\brm{\begin{rmat}}
\newcommand\erm{\end{rmat}}
\newcommand\beq{\begin{equation}}
\newcommand\eeq{\end{equation}}

\def\eps{{\epsilon}}

\def\mR{{\bf R}}

\def\mY{{\bf Y}}
\def\mR{{\bf R}}

\def\cD{{\mathcal{D}}}
\def\cC{{\mathcal{C}}}
\def\cM{{\mathcal{M}}}
\def\cN{{\mathcal{N}}}
\def\cO{{{O}}}

\def\cX{{\mathcal{X}}}

\def\bR{{\mathbb{R}}}
\def\bE{{\mathbb{E}}}

\def\mb{{\bf b}}
\def\mw{{\bf w}}
\def\mx{{\bf x}}
\def\my{{\bf y}}
\def\mz{{\bf z}}
\def\mzeta{{\bf \zeta}}
\def\mzA{{\bf z^{\mathcal{A}}}}
\def\mzAt{{{\bf z}_t^{\mathcal{A}}}}
\newcommand\mzAtk[1]{{{\bf z}_t^{\mathcal{A}_{#1}}}}

\def\mm{{\bf m}}

\def\rd{{\rm d}}

\def\bv {{\bf v}}
\def\bmx {{\bar{\fvar}}}
\DeclareMathOperator{\arctantwo}{arctan2}
\newcommand{\IMmth}{\ensuremath{\xspace\cM_{\eps}}\xspace}
\newcommand{\IMxmth}{\ensuremath{\xspace\cM^{\mathbf{x}}_{\eps}}\xspace}
\newcommand{\hatIMmth}{\ensuremath{\xspace\hat{\cM}_{\eps}}\xspace}
\newcommand{\hatIMmthk}[1]{\ensuremath{\xspace\hat{\cM}_{\eps}^{#1}}\xspace}
\newcommand{\IMmthepshatk}[1]{\ensuremath{\xspace\hat{\cM}_{\eps}^{#1}}\xspace}
\newcommand{\SMmth} {\ensuremath{\xspace\cM_{0}}\xspace}
\newcommand{\SMxmth} {\ensuremath{\xspace\cM^{\mathbf{x}}_{0}}\xspace}

\def\fvari{{\boldsymbol\xi}}
\def\cov{{\text{cov}}}

\def\Ito{{\text{It\^o}}}
\def\tr{{\mathrm{tr}}}

\usepackage{ntheorem}\theoremseparator{.}
\newtheorem{myassum}{assumption}\theoremseparator{.}

\newcommand{\lndmrkidx}[1]{^{{#1}}}
\newcommand{\lndmrk}[1]{\hat{\mz}\lndmrkidx{#1}}
\newcommand{\fvar}{\my}
\newcommand{\svar}{\mx}

\newcommand{\mzIC}{\mz_0}
\newcommand{\fovar}{\mz^{\text{fst}}}
\newcommand{\sovar}{\mz^{\text{slw}}}
\newcommand{\sovarbar}{\bar{\mz}^{\text{slw}}}
\newcommand{\fovarbar}{\overline{\mz}^{\text{fst}}}
\newcommand{\fsubspace}{\mathbb{V}^{\text{fst}}}

\newcommand{\mzlslow}{\mz^{l,\text{slw}}}
\newcommand{\blN}{\hat\mb\lndmrkidx{l}}
\newcommand{\bl}{\hat\mb\lndmrkidx{l}}
\newcommand{\ClN}{\hat C\lndmrkidx{l}}
\newcommand{\ClNbar}{\bar C\lndmrkidx{l}}
\newcommand{\trueexp}{\mm\lndmrkidx{l}_t}

\newcommand{\truecov}{C(\mz\lndmrkidx{l}_t|\mz\lndmrkidx{l}_0)}

\newcommand{\trueexpslp}{\mb\lndmrkidx{l}}

\newcommand{\truecovslp}{\Lambda\lndmrkidx{l}}

\newcommand{\HldN}{\ensuremath{\hat H\lndmrkidx{l}_d}\xspace}

\newcommand{\LambdaldN}{{\hat\Lambda\lndmrkidx{l}_d}}
\newcommand{\LambdalN}{\hat\Lambda\lndmrkidx{l}}

\newcommand{\GammalN}{\hat\Gamma\lndmrkidx{l}}
\newcommand{\Lambdal}{\hat\Lambda\lndmrkidx{l}}

\newcommand{\hatmetric}{\ensuremath{\hat\rho}\xspace}
\newcommand{\metricneighborhood}{B}
\newcommand{\PNlop}{\ensuremath{\hat P\lndmrkidx{ l}}\xspace}

\newcommand{\PNl}[1]{\PNlop{({#1})}\xspace}

\newcommand{\PNlz}{\ensuremath{\PNl{\mz}}\xspace}

\newcommand{\Plz}{\ensuremath{\hat P\lndmrkidx{ l}(\mz)}\xspace}
\newcommand{\mmlNt}[1]{{\hat{\mm}\lndmrkidx{l}_{#1}}}
\newcommand{\ClNt}{\hat{C}\lndmrkidx{l}_t}
\newcommand{\lndmrklN}{\lndmrk{l}}
\newcommand{\barmmlNM}{\bar\mm\lndmrkidx{l}_M}
\newcommand{\ProjA}{\ensuremath{\hat{P}^{\mathcal{A}}}}
\newcommand{\bA}{\hat{\mb}^{\mathcal{A}}}
\newcommand{\LambdaA}{\hat\Lambda^{\mathcal{A}}}
\newcommand{\HA}{\ensuremath{{\hat H^{\mathcal{A}}}}\xspace}
\newcommand{\Projrkd}{\ensuremath{\mathrm{Proj}_{\mathrm{rk}(d)}}\xspace}

\newcommand{\tanPlaneldN}{\ensuremath{\hat T_{\lndmrk{l}}\IMmth}\xspace}

\newcommand{\hattildemetric}{\ensuremath{\hat{\tilde \rho}}\xspace}
\newcommand{\Uslowl}{\ensuremath{\hat U\lndmrkidx{l,\mathrm{slw}}_{d}}\xspace}
\newcommand{\Uslowtruel}{\ensuremath{ U\lndmrkidx{l,\mathrm{slw}}_{d}}\xspace}
\newcommand{\Vfastl}{\ensuremath{\hat V\lndmrkidx{l,\mathrm{fst}}_{D-d}}\xspace}
\newcommand{\Vfasttruel}{\ensuremath{ V\lndmrkidx{l,\mathrm{fst}}_{D-d}}\xspace}
\newcommand{\hatEl}{{\hat{E}^{l}}}

\newcommand{\Vfasttrue}{\ensuremath{ V\lndmrkidx{\mathrm{fst}}_{D-d}}\xspace}
\newcommand{\Uslowtrue}{\ensuremath{ U\lndmrkidx{\mathrm{slw}}_{d}}\xspace}

\definecolor{Gray}{gray}{0.8}
\definecolor{MyGray}{gray}{0.6}
\definecolor{LightCyan}{rgb}{0.88,1,1}
\definecolor{LightRed}{rgb}{1,0.8,0.8}
\newcolumntype{a}{>{\columncolor{Gray}}c}
\newcommand{\IM}{invariant manifold\xspace}
\newcommand{\SM}{slow manifold\xspace}
\newcommand{\peanut}{pinched sphere\xspace}
\newcommand{\Peanut}{Pinched sphere\xspace}

\newcommand{\relerrb}{\ensuremath{{\text{relErr}}({\hat{\mathbf{b}}})}\xspace}
\newcommand{\relerrLambda}{\ensuremath{{\text{relErr}}({{\hat\Lambda}})}\xspace}
\newcommand{\abserrIMmth}{\ensuremath{{\text{AbsErr}}({{\hatIMmth}})}\xspace}
\newcommand{\abserrTangle}{\ensuremath{{\text{AbsErr}}({{\hat{T}_{\lndmrk{l}}}})}\xspace}
\graphicspath{{./fig/main/}{./fig/SI/}}

\numberwithin{equation}{section}

\usepackage{authblk}

\title{
Nonlinear model reduction for slow-fast stochastic systems near unknown invariant manifolds
}

\author[a]{Felix X.-F. Ye \thanks{Corresponding author. xye2@albany.edu}}
\author[b]{Sichen Yang} 
\author[b,c]{Mauro Maggioni}

\affil[a]{Department of Mathematics \& Statistics, University at Albany}
\affil[b]{Department of Applied Mathematics and Statistics, Johns Hopkins University}
\affil[c]{Department of Mathematics, Johns Hopkins University} 

\date{} 

\begin{document}
\vspace{-6mm}
\maketitle
\vspace{-10mm}
\begin{abstract}

We introduce a nonlinear stochastic model reduction technique for high-dimensional stochastic dynamical systems that have a low-dimensional invariant effective manifold with slow dynamics, and high-dimensional, large fast modes. Given only access to a black box simulator from which short bursts of simulation can be obtained, we design an algorithm that outputs an estimate of the invariant manifold, a process of the effective stochastic dynamics on it, which has averaged out the fast modes, and a simulator thereof. This simulator is efficient in that it exploits of the low dimension of the invariant manifold, and takes time steps of size dependent on the regularity of the effective process, and therefore typically much larger than that of the original simulator, which had to resolve the fast modes.
The algorithm and the estimation can be performed on-the-fly, leading to efficient exploration of the effective state space, without losing consistency with the underlying dynamics. 
This construction enables fast and efficient simulation of paths of the effective dynamics, together with estimation of crucial features and observables of such dynamics, including the stationary distribution, identification of metastable states, and residence times and transition rates between them. 

\end{abstract}

\section{Introduction}
Many mathematical models of dynamical systems, across the sciences, are based on Ordinary and Stochastic Differential Equations (ODEs and SDEs, respectively), with a large number of degrees of freedom, often with dynamics at very different timescales. 
These systems offer multiple significant challenges to their simulation and understanding, which often require collecting a large number of long trajectories to capture a wide variety of possible behaviors of the system. These challenges include:
(i) the high dimension of the state space and the corresponding large number of equations;
(ii) many fast/stiff modes, corresponding to very rapid fluctuations (e.g. solvent molecules around a protein);
(iii) metastability, with trajectories dwelling for large times in certain regions of state space (metastable states), with rare transitions between them.

These challenges often compound in a single system, making the large scale (in state space and in time) phenomena of the system, which are often of interest in applications, difficult to capture and study. Some of the properties that one often wishes to capture include the \IM, around which trajectories lie; the stationary distribution, describing the large-time distribution of the system in state space; the residence times describing the distribution (or its expectation) of the time spent in a metastable state $M$ before leaving it, once started in $M$; the transition rates and transition paths, containing information about the expected time and most likely paths followed by the system when transitioning between metastable states; the reaction coordinates, representing low-dimensional observables whose dynamics is approximately Markovian and predictive of transitions between metastable states; the leading eigenvalues and eigenvectors of the generator of the process, related to transition rates and reaction coordinates respectively \cite{DiffusionPNAS,CKLMN:DiffusionMapsReductionCoordinates,MSM-review, jns-review, clementiARPC2013,Andreas2018, DMD-textbook, Eric-book, leimkuhler2015molecular,Legoll1, Legoll2,Givon-review,ALEXANDER2020132520}.

These objects of interest intertwine geometry and dynamics, and are our focus in this work: we aim at jointly estimating crucial geometric objects, such as a low-dimensional invariant manifold \IMmth, and effective dynamics of the system at and above a given timescale, via estimated stochastic equations, that average out complex, high-dimensional aspects of the dynamics below that timescale. This {{reduced model}} can be more amenable to faster simulation, with low-dimensional equations and time steps much larger than those needed by a simulator of the original system. 

As with any type of model reduction, loss of information is in general unavoidable, and it has possibly dramatic consequences, among which loss of Markovianity, and to a loss of accuracy in predictions by the reduced model, especially at large times. Our approach aims at reducing these problems, at least on a suitable class of systems. We consider the problem of nonlinear model reduction for stochastic systems that, while presenting the above challenges, have features that are possibly redeeming, if appropriately exploited: fast and slow modes of evolution of the system, with a non-negligible {\it{separation of timescales}};  {\it{a low-dimensional \IM}}, onto which the dynamics may be projected by averaging the fast modes, while preserving information about the large-scale/time phenomena; {\it{fast modes that may be linearized}}, but may be high-dimensional, and have large magnitude, with varying direction relative to the invariant manifold. The objects we need to estimate are nonlinear: the invariant manifold, the corresponding reduction map onto it, and the effective stochastic equations on it.

The modality in which we have access to ground-truth trajectory data is important for algorithmic, statistical and computational considerations. We assume to be given access to the system via a black-box simulator $\mathcal{S}$ taking time-steps $\delta t\lesssim\epsilon$ (to resolve the fast modes, whose derivative has the order $1/\epsilon$), that we can call to obtain only short trajectories, of length-in-time of order $\tau\gg\epsilon$, where $\tau$ is typically of the order of the relaxation timescale of the fast modes. From a given initial condition we use the simulator $\mathcal{S}$ to obtain a burst of $N$ short paths, each of time-length $\cO(\tau)$. This now classical set-up \cite{yannis2009jcp} enables trivial parallelization, across initial conditions and paths from each initial conditions, and is well-suited to applications \cite{yannis2015mmnp, yannis2015jcp, leimkuhler2015molecular, dietrich2021learning}. A crucial problem is how initial conditions for these bursts are made available. When they cannot be chosen by the estimation procedure, they may be modeled as randomly drawn, ideally from a probability measure on state space that is reasonably well-distributed over the state space: it is then straightforward to estimate how many initial conditions needs to be sampled to guarantee, with high probability, coverage (see e.g. \cite{CM:ATLAS}). When the initial conditions may be chosen by the estimation procedure, a natural exploration-exploitation dilemma arises: refining the estimates in a region of space already populated by initial conditions by sampling more initial conditions and paths, or generate paths from initial conditions ``outside'' from the parts of state space already visited? And how to generate the latter? Especially when the state space is high-dimensional, and the dynamics of interest is along a low-dimensional invariant manifold, it is not trivial to sample new initial conditions. Even more so in applications, where many physical constraints are often extremely complex and unknown. We develop a simple approach, called ``exploration mode'', discussed below, that addresses both the exploration-exploitation dilemma and the problem of generating new initial conditions ``outside'' the already-visited state-space; all  needed is a small number (e.g. $1$) of initial conditions; our numerical examples will be run in ``exploration mode''.

From the observations of $N$ paths from an initial condition, we estimate locally the \IM \IMmth, the effective local directions of the fast modes and an oblique projection along them onto \IMmth, and an effective drift and diffusion coefficient to be used for an effective It\^o diffusion on \IMmth. In these steps, it is crucial to avoid the curse of dimensionality, which would demand a number of observations exponential in the high dimension $D$ of the state space. We achieve this by using simple parametric models for these local estimators, and prove that the sampling requirement scale favorably linearly in $D$. We then piece together these local estimators of the effective dynamics at timescales $\tau$, to obtain a global estimate of \IMmth, of a nonlinear projection onto \IMmth, and of a process on \IMmth called ATLAS.

The ATLAS stochastic process $(\mzAt)_{t\ge0}$ takes place on the estimated \IMmth, and aims at reproducing, at timescale $\tau$ and above, the dynamics of the original process on \IMmth, after averaging out the fast modes. The ATLAS process may be simulated much more efficiently than the original process, as it is lower dimensional and amenable to be simulated with time-steps of size $\tau\gg\eps$, instead of size $\lesssim\epsilon$ as in $\mathcal{S}$. We demonstrate this construction numerically on several systems that display several different salient features: nonlinear slow manifolds, lack of a global map to globally linear slow and fast variables, linear and curvilinear fast modes, with and without a clear separation of timescales between fast and slow modes.

As we mentioned above, in many applications (e.g. in molecular dynamics) a ``large-enough'' set of initial conditions, at which to collect bursts of paths, is not known. A key contribution of this work is to introduce a construction for ATLAS in {\it{``exploration mode''}}, where we initially construct ATLAS from a very small number of initial conditions, and update it on the fly by collecting new bursts of simulations from $\mathcal{S}$, started at automatically well-chosen initial conditions, whenever ATLAS trajectories leave an ever-increasing ``domain of competency'' of the current ATLAS. This yields an increasing family of ATLASes, {\it{each consistent with the previous ones and with the original dynamics, on ever-increasing subsets of the state space}}, without over-sampling already explored regions. To efficiently and consistently explore the effective state space for the system is a crucial ability of ATLAS, achieved with techniques very different from existing ones, which typically are based on biasing the dynamics, and trading exploration with fidelity to the original dynamics \cite{ChiavazzoE5494, yannis2009, plumed2,clementiJPC2013, Chen3235}. 

All our numerical experiments are run exclusively in exploration mode, and demonstrate that ATLAS accurately reproduces features of the dynamics at medium and  large time scales, and enables the efficient construction of Markov State Models (MSMs), and of approximations of important observables, such as eigenvalues and eigenvectors of the generator of the dynamics. These in turn may be used for further reduction of the dynamics at very large time scales, estimating transition rates, and yielding low-dimensional embeddings of \IMmth.

\section{Fast-slow SDEs with slow nonlinear manifolds}
\label{s:FastSlowSDEs}
A classical model of fast-slow SDEs is
\begin{equation}
\left\{
\begin{aligned}
&\rd \svar_t = g(\svar_t, \fvar_t)   \rd t +  G(\svar_t, \fvar_t)\, \rd U_t \\
&\rd \fvar_t = \frac{1}{\eps} f(\svar_t, \fvar_t) \rd t + \frac{1}{\sqrt{\eps}}F(\svar_t, \fvar_t)\, \rd V_t
\end{aligned}
\right. ,
\label{e:master}
\end{equation}
where $\eps>0$ is a small parameter, determining the separation of timescales,  $\fvar_t\in \bR^{D-d}$ and $\svar_t\in \bR^{d}$ are, respectively, the fast and slow variables, and $(U_t)_{t\ge 0}$, $(V_t)_{t\ge 0}$ are independent Wiener processes in $\bR^d$ and $\bR^{D-d}$ respectively. The drift coefficients $f,g$ and the diffusion coefficients $F,G$ are assumed to be regular, e.g. twice-differentiable. Systems governed by this type of equations have been extensively studied \cite{pavliotis2008multiscale,NB-book,vanKampen, gardiner2009}. We are interested in the situation, common in applications, where the ambient dimension $D$ is much larger than ``intrinsic'' dimension $d$ of the slow variables. 
The time-step $\delta t$ in the original simulator $\mathcal{S}$ of \cref{e:master} is typically $\lesssim\eps$ to ensure accuracy and stability of the numerical scheme, making it computationally onerous. 
This constraint on the time-step is generally applicable only to explicit schemes, motivating continued research in implicit schemes, which however in high-dimensions still appear to be computationally prohibitive. While the techniques we introduce are also applicable to ODEs with minor changes, including for example substituting bursts of stochastic paths by bursts of deterministic paths started by stochastically perturbed initial conditions, we focus on SDEs to streamline the presentation. Also, recall that fast-slow ODEs may be approximated by SDEs, at least in the limit $\eps\rightarrow0$ \cite{pavliotis2008multiscale}.

\subsection{The slow manifold, and averaged equations on it}
\label{slow-manifold-latent}

The fast variable $\fvar$ is assumed to relax, at a timescale $O(\tau)$, and stay close to the \SM $\SMxmth:=\{(\svar,\fvar^\star(\svar)) : f(\svar,\fvar^\star(\svar))=0 \}$ of the corresponding deterministic system (with $F,G\equiv0$), if the \SM is asymptotically stable. Geometric singular perturbation theory implies the existence of an \IM \IMxmth, close to \SMxmth, see \cite{NB-book,NB-jde, Kuehn-book} and \cref{appx:assumption}. Under suitable further conditions, one then obtains a reduced set of equations on \IMxmth, with the drift and diffusion coefficients on \IMxmth obtained by locally {\it{averaging}}, at each $\svar_0\in\IMxmth$, those in \cref{e:master} against the conditional invariant measure $\nu(\fvar|\svar=\svar_0)$ of the fast modes \cite{pavliotis2008multiscale,NB-book}.  The technical assumptions needed can be far from trivial, e.g. often $G$ is assumed independent of $\fvar$ \cite{Veretennikov_1991,Givon_2006,Givon_2007}. The {\it{reduced equations}} are
\begin{equation}\label{e:reduced-par}
\begin{aligned} 
    \rd \bar{\svar}_t = \bar{g}(\bar{\svar}_t)\rd t + \bar{G}(\bar{\svar}_t) \rd U_t\,,\qquad
    \bar{\fvar}_t = \bmx(\bar{\svar}_t,\eps)\,,
\end{aligned}
\end{equation}
which define a process on the \IM \IMxmth, for small $\epsilon$ (see \cref{appx:averaging} for details). Having averaged out the fast variables, the reduced dynamics deliberately lose information about the details of the dynamics of the fast variables and phenomena below the timescale $\tau$, but it yields a low-dimensional process (in the regime of interest $d\ll D$), that reproduces the effective dynamics of the original system on \IMxmth at timescale of order $O(\tau)$ and, ideally, beyond.

\subsection{Nonlinear observations, unknown slow/fast variables}
\label{s:nonlinearobs}
In the model in \cref{e:master} the slow and fast variables are the given, linear, and orthogonal coordinates $\svar$ and $\fvar$. In applications, however, the slow and fast variables are typically not known a priori and need to be identified, and in general they are neither linear nor orthogonal \cite{GSPT-beyond}. 

This motivates the following observation model. We view the system in \cref{e:master} as as a {{black-box latent local model}}. {\it{Black-box}} because equations are not available to us. {\it{Latent}} because we do not have access to $\svar$ and $\fvar$, but to observations $\mz$, ranging in $\Omega\subseteq\mathbb{R}^{D}$, which can be mapped to latent variables $(\svar,\fvar)\in\mathbb{R}^D$,  satisfying \cref{e:master}. {\it{Local}} because such a map is not a global map, but is in fact realized by a collection of charts $\{(\mathcal{U}_\alpha,\varphi_\alpha)\}_\alpha$, consisting of open neighborhoods $\{\mathcal{U}_\alpha\}_\alpha$ covering $\Omega$ and smooth maps $\varphi_\alpha:\mathcal{U}_\alpha\rightarrow\mathbb{R}^D$, each invertible on its range and such that $\varphi_\alpha\circ\varphi_{\alpha'}^{-1}$ is smooth where defined, so that for every point $\mz\in\Omega$ the local latent variables $(\svar,\fvar)=\varphi_\alpha(\mz)$ satisfy \cref{e:master}, for $\mathcal{U}_\alpha\ni\mz$ (there exists one such $\mathcal{U}_\alpha$ since $\{\mathcal{U}_\alpha\}_\alpha$ covers $\Omega$).
Geometrically, this is of course the natural set-up for expressing that the observations $\mz$ are on a manifold parametrized by a system of charts (called an atlas, in differential geometry). This geometric perspective is here merged with the dynamics, through the condition that the local parametrizations map the dynamics of the observed variables to a dynamics where the latent variables follow the model equations \eqref{e:master}.
This model is inspired and generalizes that of \cite{Singer2009, GSPT-beyond}, where the aim was to discover an embedding of the underlying slow variables, via the lowest frequency eigenfunctions of an estimated generator of the whole process, and not necessarily a parametrization of the invariant manifold, nor effective equations on it. 
That approach is broadly applicable to a larger class of processes than ours, for example when the fast modes are highly nonlinear; however this comes at the price of falling victim of the curse of dimension, requiring sampling paths from an exponentially large number of initial conditions (this is not discussed in \cite{Singer2009}, but it could be derived). We shall estimate the slow variables and effective equations directly, without first learning the detailed behavior of the high-dimensional fast variables, and subsequently reduce the dynamics via eigen-decompositions.

In the observed variables $\mz$, the process $(\svar_t,\fvar_t)_{t\ge0}$ maps to a process $(\mz_t)_{t\ge0}$, where slow and fast variables are in general nonlinearly mixed, instead of being linear and orthogonal as in \cref{e:master}. The slow variables $\sovar$ will lie on a nonlinear \IM \IMmth of dimension $d$; trajectories will lie in a domain of concentration around \IMmth, that we model as a non-self-intersecting tube around \IMmth. 
$\IMmth$ is close to a slow manifold \SMmth (which in general is {\it{not}} the image of $\SMxmth$ under the maps $\varphi_\alpha^{-1}$). Locally around the initial point $\mzIC\in\IMmth$, one may linearize the equations for $(\mz_t)_{0\le t\lesssim\tau(\mzIC)}$ to a form similar to \cref{e:master}, with $\mx$ replaced by slow variables $\sovar$ and $\my$ replaced by fast variables $\fovar$. Under the same linearization, $\sovar$ is approximated as lying in the tangent space $\smash{T_{\mzIC}\IMmth}:=\mathrm{span(col(}\Uslowtrue))$ to \IMmth at $\mzIC$, and   $\fovar$ is approximated as lying in $\smash{\fsubspace_{\mzIC}:=\mathrm{span(col(}\Vfasttrue))}$. The slow and fast directions $\smash{\Uslowtrue, \Vfasttrue}$ in general vary, smoothly, with $\mzIC$. 
$(\IMmth-\mzIC)$ is locally a graph of a function $\fovarbar_\eps(\cdot;\mzIC): \smash{T_{\mzIC}\IMmth\rightarrow\fsubspace_{\mzIC}}$ over the slow variables $\sovar$. 
We can then proceed to the reduction to equations in the slow variables $\sovar$ only, in a form similar to \cref{e:reduced-par}, by averaging the fast variables at a prescribed timescale $\tau$, obtaining a reduced process on \IMmth.

\begin{figure*}
\centering
\includegraphics[width=0.75\textwidth]{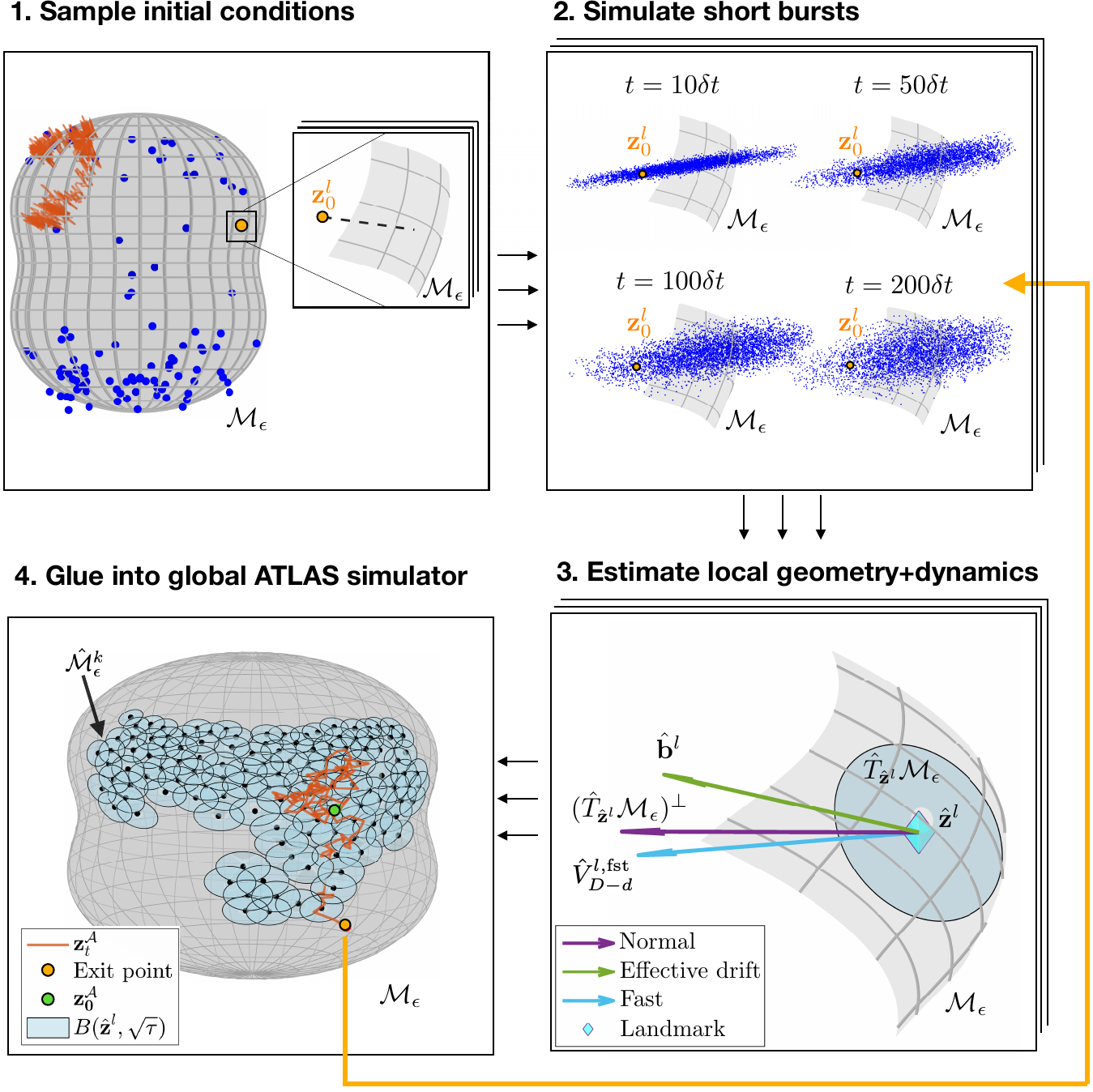}
\caption{High-level overview of the steps in the construction of ATLAS, for a system exhibiting fast large oscillations around an invariant manifold \IMmth shaped as a pinched sphere. {\textbf{1. Sample initial conditions}}, according to a given probability measure. Typically initial conditions (ICs) do not lie on \IMmth (see zoom-in portion: IC represented by orange dot is not on \IMmth), nor are they well-distributed throughout state space. A sample path of the original system is shown, which oscillates around \IMmth with large amplitude. {\textbf{2. Simulate short bursts}}: from each IC, in parallel, a burst of short trajectories, of time length comparable to the relaxation time scale $\tau$ of the fast modes, is obtained from a black-box simulator of the original system (if $\tau$ is not given, it may be estimated from these trajectories; in this example $\tau=200\delta t$). In this example the fast modes have large amplitude, but our technique will correctly determine fast and slow directions. {\textbf{3. Estimate local geometry and dynamics:}} from the trajectories in each burst, the local geometry of \IMmth (including a landmark $\lndmrk{l}$ on \hatIMmth, an affine tangent space $\smash{\tanPlaneldN}$, and an oblique projection $\PNlz$ onto it with kernel $\smash{\mathrm{span(col(}\Vfastl}$), as well as parameters of the reduced effective dynamics (drift $\blN$ and diffusion coefficient $\LambdaldN$), are estimated. {\textbf{4. Glue into global ATLAS simulator:}} the local geometry and dynamics estimators are glued together into a global ATLAS, with an associated simulator of a reduced effective ATLAS process $(\mzAt)_{t\ge0}$. We display estimated local tangent ellipses corresponding to balls in a diffusion-induced local metric, a sample path from the ATLAS simulator. However, at this stage ATLAS may still not cover \IMmth: in ``exploration mode'' it will simulate paths that stop when reaching the boundary of the current ``domain of expertise'' (orange point), collect new bursts at those locations (as in step 2), update ATLAS with the estimators from the new region (as in step 3), and then resume the simulation.}
\label{f:ATLASmain}
\end{figure*}

\subsection{Structure of the local reduced effective equations}    
\label{struc-local-eq}
We assumed that the deviation of the fast variable from the \IM \IMmth lies, exactly or approximately, in $\fsubspace_{\mzIC}$; in this subspace let it be given by  ${\fvari_t:= \fovar_t- \fovarbar_\eps(P(\sovar_t);\mzIC)}$ with $t\lesssim\tau(\mzIC)$, where $P$ is the projection onto $\smash{T_{\mzIC}\IMmth}$ with kernel $\fsubspace_{\mzIC}$. The idea of averaging in fast-slow systems \cite{pavliotis2008multiscale,freidlin2012random,Givon-review,NB-book, vanKampen} exploits the timescale separation between slow and fast variables: at the separation timescale $\tau(\mzIC)$, the dynamics of $\fvari_t$ conditioned at $\sovar_t=\mzIC$ reaches its quasi-equilibrium distribution $\nu(\fvari|\sovar_t=\mzIC)$, which we approximate by a $(D-d)$-dimensional Gaussian distribution $\cN(0, \Xi(\mzIC))$. If the trace of $\Xi(\mzIC)$ is large, the fast oscillations of $\fvari_t$ around \IMmth have large expected amplitude.

We discuss in sec.\ref{s:ATLASconstruction} how $\tau(\mzIC)$ may be estimated from observations of short trajectories. We assume throughout, but only in order to simplify the presentation, that $\tau(\mzIC)$ can be chosen to be the same at all locations $\mzIC$: we simply denote it as $\tau$, and is assumed as given. By {\it{stochastic averaging}}, in these coordinates, the {\bf{reduced}} stochastic dynamics on the slow variables is obtained by averaging the drift and diffusion terms by this quasi-equilibrium distribution, leading to reduced SDEs
\begin{equation} \label{e:master-atlas}
     \rd\sovarbar_t  = b(\sovarbar_t)\rd t + H(\sovarbar_t)\rd U_t\,,
\end{equation}
similar to those in \cref{e:reduced-par}. These SDEs may be viewed in intrinsic coordinates, or in Cartesian coordinates in the ambient space $\mathbb{R}^D$, with $\sovar_t\in\mathbb{R}^D$ but on \IMmth, $b\in\mathbb{R}^D$ a vector field on \IMmth, and $H\in\mathbb{R}^{D\times d}$ acting on a Wiener process $U_t$ in $\mathbb{R}^d$.

In classical stochastic averaging, it considers $\epsilon\rightarrow0$ and the separation timescale $\tau$ is typically of order $\eps$, see e.g. \cite{Givon_2006,Givon_2007,cms/1288725269,Zhang_2018}. Here, instead, motivated by applications, we consider $\epsilon$ fixed, unknown and unused, and $\tau$ fixed, known or estimated, and larger than $\eps$, and independent thereof. The dynamics of $\sovar_t$ is low-dimensional, taking place on \IMmth, and represents the reduced {\bf{effective}} dynamics at timescales $\tau$ and beyond, having averaged out the fast transients of the high-dimensional process $(\fvari_t)_{t\ge0}$ at timescales $\lesssim\tau$. Simulating $\sovar_t$ requires a {\it{time-step independent of $\epsilon$}}, often much larger than $\eps$, and only dependent on the regularity of \IMmth and of the regularity of the effective drift $b$ and diffusion coefficient $H$ on \IMmth.

Our goal is to estimate a process, called ATLAS, that approximates $\sovarbar_t$, on an estimated \hatIMmth, given observations of bursts of short trajectories; in all our examples the simulator of ATLAS will take time-steps exactly equal to $\tau$. 

\subsection{ATLAS: learning a reduced effective model} 
Given observations of multiple bursts of short trajectories, of time length $\cO(\tau)$, around each of a collection of initial points $\{\mz\lndmrkidx{l}_0\}_{l=1,\dots,L}\subset\mathbb{R}^D$, we estimate: the local slow variables by estimating a point $\mz^l$ on \IMmth and a local tangent space $T_{\mz^l}\IMmth$ to \IMmth at  $\mz^l$; a subspace $\fsubspace_{\mz^l}$ transversal to \IMmth at $\mz^l$ containing the linearized fast modes $\fvari$; effective drift and diffusion coefficients for $\sovar_t$ in $T_{\mz^l}\IMmth$ as in \cref{e:master-atlas} for the effective dynamics of the slow variables around $\mz^l$ at the timescale $\tau$. These objects are completely local, around each $\mz^l$.  Since a global reduction step bringing the equations to the standard form \cref{e:master} may not be possible, for example because of global topological  obstructions (e.g., a slow manifold consisting of a circle cannot be mapped globally to a linear coordinate), in the spirit of the very definition of manifolds and their atlases, we will ``glue'' together the estimated local charts and equations into a set of charts and smoothly co-ordinated equations, generating a process, called ATLAS, and a corresponding simulator for obtaining global paths on the estimated \IM. ATLAS, given enough data and under suitable assumptions on the dynamics, estimates in a consistent and accurate fashion the local dynamics and its statistics, we also demonstrate in our numerical experiments that important long-time observables, including the stationary distribution and mean residence times in regions, metastable and not, of state space, are accurately estimated by ATLAS.

\begin{algorithm}[t]
\begin{algorithmic}
\Require $\mathcal{S}$: original simulator of the system; $\mu_0$ probability measure for initial conditions; $\tau$: timescale for reduction; $L$: number of initial conditions.
\Ensure $\mathcal{A}$: simulator of the effective process $(\mzAt)_{t\ge0}$.
\State - Estimate local parameters of effective dynamics \hfill [sec.\ref{s:estLocalParams}]
\State - Estimate geometric properties of \IMmth \hfill [sec.\ref{s:IMestimation}]
\State - Construct ATLAS using the above estimators \hfill [sec.\ref{s:ATLASsimul}]
\end{algorithmic}
\caption{High-level pseudo-code for ATLAS construction}
\label{a:ATLAShighlevel}
\end{algorithm}

\section{ATLAS construction}
\label{s:ATLASconstruction}
During construction, ATLAS is assumed to have access to a black-box simulator $\mathcal{S}$, that takes as input an initial condition $\mz_0$ and a time $t_0$, typically $\cO(\tau)$, and returns a path $(\mz_t)_{t\in[0,t_0]}$, driven by the latent equations as in \cref{e:master}. The construction proceeds in multiple steps, see \cref{a:ATLAShighlevel}.

Before describing the details, we present an example.

\begin{figure}[t]
\centering
\includegraphics[width=0.5\textwidth]{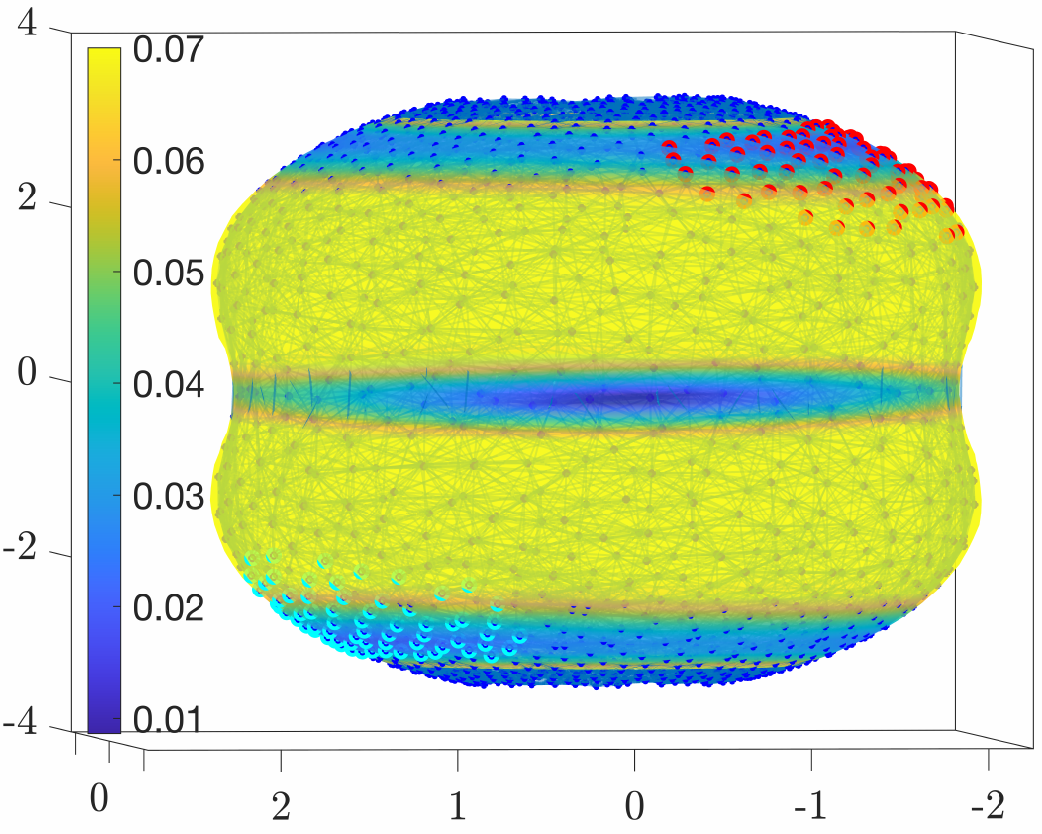}

\caption{Representation of the estimated \peanut \hatIMmth, together with the landmarks $\{\lndmrk{l}\}_l$ (blue dots) and their local connectivity graph, all as constructed by ATLAS during long exploration. The dynamics in this case is not reversible, and the fast modes have large standard deviation (comparable to the reach of \hatIMmth) and are a.e. not orthogonal to \hatIMmth (see \cref{f:peanut-singular}). The surface color is the norm of the effective drift as a function on \hatIMmth. The landmarks are about $10^{-2}$-close to \hatIMmth, see \cref{t:error}. The regions around the poles are very rarely visited, as the drift is unbounded there, creating a repulsion. In this visualization we have truncated those regions as landmarks become denser and denser. The landmarks marked in red and cyan represent the regions have high probability.}
\label{f:peanutM}
\end{figure}

\noindent{\textbf{Example: Fast/slow system around a \peanut}}.
This system is used as a reference throughout our discussion, construction and testing of ATLAS. 
The process is an It\^o diffusion on a ``smoothly pinched'' 2 dimensional sphere centered at the origin (the \IM $\hatIMmth\subseteq\mathbb{R}^3$, see \cref{f:peanutM}), perturbed by very rapid fluctuations in the radial direction. These fast modes are (a) {\it{large}} (equal to a significant fraction of the reach of \IMmth); (b) a.e. {\it{not}} orthogonal to \IMmth, see fig.\ref{f:ATLASmain} (step 3), and fig.\ref{f:peanut-singular}; and (c) may be approximated by a radial Ornstein-Uhlenbeck (O-U) process. For these reasons, a local PCA of an ensemble of short trajectories would fail to estimate the local tangent plane to \IMmth. Given $\tau$, at least as large as the timescale of relaxation of the fast modes, the correct effective equations on \IMmth should be obtained by averaging along an appropriate oblique (radial, in this case) projection onto \IMmth. 

We remark that neither the original system (in $\mathbb{R}^3$) nor the slow system on \IMmth are driven by overdamped Langevin equations: the drift is not the gradient of a potential, the diffusion coefficient is not constant, and the process is not reversible. Therefore, methods based approximations by an overdamped Langevin equations, such as those in \cite{DiffusionPNAS,CKLMN:DiffusionMapsReductionCoordinates,Singer2009,RZMC:ReactionCoordinatesLocalScaling}, would be biased, and likely inaccurate. The effective dynamics on \IMmth has two high probability regions, separated by regions of large volume where  drift is small compared to diffusion (``entropic barriers''), which  could make standard approximations of those inaccurate \cite{ bicout_szabo_2000}.

To give some intuition about local geometric and dynamical quantities that play a fundamental role in this system, and more generally for systems that motivate our constructions, we show in \cref{f:peanut-singular} a portion of the \hatIMmth about a point $\mz_0$, a corresponding trajectory of the system started at $\mz_0$, and several key directions in $\mathbb{R}^3$: the normal to  \hatIMmth at $\mz_0$, the estimated effective direction of the fast modes at timescale $\tau$, which is significantly different from the normal direction. We also depict the direction of the estimated effective (It\^o) drift, which, as expected, is not (and, in fact, far from being) tangent to  \hatIMmth. These depicted objects are exactly those estimated in the ATLAS construction, from local bursts of simulations. 

Finally, the global geometric approximation of \IMmth and effective ATLAS process are assembled. An ATLAS path, used for exploration, is shown in \cref{f:ATLASmain}; the accuracy of ATLAS is demonstrated in various metrics, from the geometric approximation of \IMmth to the approximation of effective drift and diffusion coefficients, to the accuracy of estimation of statistics of the process such as mean residence time in relatively small regions of state space and in metastable states (see sec.\ref{s:expPeanut}).

\subsection{Main steps in the construction}
We are given access to a black-box simulator $\mathcal{S}$ of the process $\mz_t$, a probability measure $\mu_0$ on the state space of the system, a separation timescale $\tau$, and a dimension $d$ for the \IM \IMmth. We will discuss later how to proceed in the very important case when $\mu_0$ is not, or insufficiently, provided (``exploration mode'', sec.\ref{s:explorationmode}), and how to estimate $\tau$ (in sec.\ref{s:estLocalParams}) and $d$ (in \cref{appx:example}). We output ATLAS, consisting of a process $(\mzAt)_{t\ge0}$, and a corresponding simulator, approximating the effective dynamics of $\mz_t$ on \IMmth at the timescale $\tau$ and beyond. 

We sample $L$ initial conditions $\{\mz\lndmrkidx{l}_0\}_{l=1,\dots,L}\sim_{\text{i.i.d.}}\mu_0$, and for each $l$ we use $\mathcal{S}$ to obtain a burst $\mathcal{B}^l$ of $N$ trajectories $\{\mz\lndmrkidx{l,n}_t\}_{n=1,\dots,N}$, each of time-length $\cO(\tau)$, starting at $\mz\lndmrkidx{l}_0$. The time-step $\delta t$ of $\mathcal{S}$ is typically $\lesssim\eps\ll\tau$ and we may think of the output of $\mathcal{S}$ as if it was in continuous time. For each $l$, we focus now on the local construction around $\mz\lndmrkidx{l}_0$, given the single burst $\mathcal{B}^l$: at timescale $\tau$, the \IM{} is locally approximated by an estimated effective tangent space $\tanPlaneldN$ of dimension $d$, at a suitably estimated point $\lndmrk{l}$; the deviation $\fvari_t^0$ from \IMmth reaches equilibrium before time $\tau$, and we approximate the dynamics of the slow variable $(\sovar_t)_{t\ge0}$ on \IMmth around $\lndmrk{l}$ by an It\^o diffusion process on $\tanPlaneldN$ as in \eqref{e:master-atlas}, which requires us to estimate an affine {\it{oblique}} projection $\PNlop$ along the fast modes and onto $\tanPlaneldN$, an effective drift $\blN$ in $\mathbb{R}^D$ (in the It\^o formulation, the drift is in general not tangent to \IMmth, see \cref{f:ATLASmain},\ref{f:peanut-singular}) and an effective diffusion coefficient $\LambdaldN$ in $\tanPlaneldN$.

\subsection{Local low-order moments of the dynamics}
\label{s:preliminaries}

The behavior of the time-dependent mean and covariance of the process started at $\mz\lndmrkidx{l}_0$ reveals crucial local properties of the geometry of the dynamics and of the slow/fast manifolds: at times $t$ comparable to $\tau$ (but, typically, not smaller nor larger), we assume that these approximations hold: 
\begin{equation}  
\begin{aligned}\label{e:meanAndCovBurst}
    &\mm\lndmrkidx{l}_t := \bE \left[\mz\lndmrkidx{l}_t \right|\mz\lndmrkidx{l}_0] = \mzlslow_0 +\mathbf{b}\lndmrkidx{l}t +\cO(\eps), \\ 
    &C(\mz\lndmrkidx{l}_t|\mz\lndmrkidx{l}_0) := \cov(\mz\lndmrkidx{l}_t|\mz\lndmrkidx{l}_0) = \Gamma\lndmrkidx{l} + \Lambda\lndmrkidx{l}t + \cO(\eps) \,,
\end{aligned}
\end{equation}
where $\Gamma\lndmrkidx{l}\succeq0$ has rank $D-d$, and represents an averaging, at time-scale $\tau$, of the covariance of the fast modes $\Xi(\mzlslow_0)$; $\Lambda\lndmrkidx{l}=H\lndmrkidx{l}(H\lndmrkidx{l})^T\succeq0$ has rank $d$ is the diffusivity of the effective reduced slow dynamics at $\mz\lndmrkidx{l}_0$. The span of $\Lambda\lndmrkidx{l}$ and $\Gamma\lndmrkidx{l}$ approximate, respectively, the tangent space $T_{\mzlslow_0}\IMmth$ to \IMmth at $\mzlslow_0$ and, respectively, $\fsubspace_{\mzlslow_0}$, which will be the kernel of a projection (in general {\it{not}} orthogonal) onto $T_{\mzlslow_0}\IMmth$. These expressions result from averaging the fast modes at timescale $\tau$; in particular the memory of the effective reduced slow dynamics is (approximately) forgotten.

The quantities above are unknown, and we estimate them from the observations from the burst $\mathcal{B}^l$:
\begin{equation}
\begin{aligned} 
\mmlNt{t} \!:=\! \frac{1}{N}\sum_{n=1}^N \mz\lndmrkidx{ l,n}_t\,\,,\,\,
\ClNt	\!:= \!\frac{1}{N-1}\sum_{n=1}^N (\mz\lndmrkidx{ l,n}_t\!\!\!\!-\mmlNt{t})(\mz\lndmrkidx{ l,n}_t\!\!\!\!-\mmlNt{t})^T\!.
\end{aligned}
\label{e:mean-cov-est}
\end{equation}
These empirical quantities, estimated from burst data, are consistent estimators of the true local mean and covariance of the process, with an approximation of order $\sqrt{\frac{d+d_f}{N}}$, where $d$ is the dimension of $\IMmth$, and $d_f$ is the number of fast modes of large amplitude - we discuss this further in \cref{s:propertiesATLAS}.

We are now ready to introduce the ATLAS construction, which proceeds in 3 main steps in \cref{a:ATLAShighlevel}, detailed in subsections \ref{s:estLocalParams}, \ref{s:IMestimation}, \ref{s:ATLASsimul} respectively; see \cref{s:SI:algos} for details.

\subsection{Estimation of local parameters of the effective dynamics}
\label{s:estLocalParams}
From each burst $\mathcal{B}_ l$, $ l=1,\dots,L$, of short simulations we compute several key quantities for constructing an approximation to the \IM \IMmth and to the reduced effective stochastic dynamics on it. The relationships in \cref{e:meanAndCovBurst} suggest that the local effective drift and diffusion coefficients for the local slow variables may be estimated from $\mmlNt{t}$ and $\ClNt$ for $t\approx\tau$. The diffusion coefficient should also yield an estimate for local tangent plane to the \IMmth, giving the local slow variables.

\begin{figure}[t]
\centering
\includegraphics[width=0.5\textwidth]{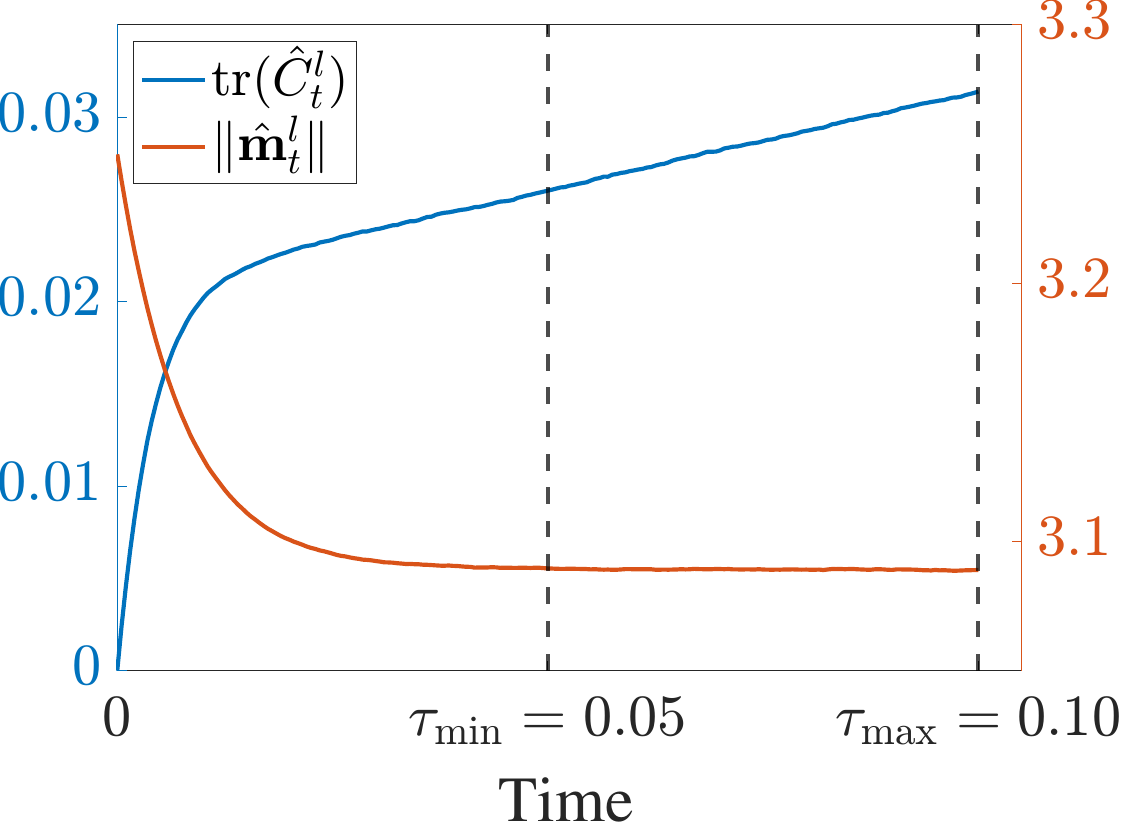}
\caption{Estimating $\tau$ from a burst of trajectories: $||\mmlNt{t} ||$ and $\tr(\ClNt)$ should both behave as linear functions of $t$ at the timescale of interest, as per \cref{e:meanAndCovBurst}.}
\label{f:estTau}
\end{figure}

\subsubsection*{Separation timescale $\tau$} 
When not given, we estimate $\tau$ from the behavior of $||\mmlNt{t}||$ and $\tr(\ClNt)$ as a function of $t$: 
for each burst, we obtain the time interval where these two quantities behave linearly, as per \cref{e:meanAndCovBurst}. We let $[\tau_{\min{}},\tau_{\max{}}]$ be the intersection of such intervals over all $l$'s, which is nonempty since we assume there exists a common relaxation time of the fast modes $\tau$ valid throughout \IM (our techniques do extend to location-dependent $\tau\lndmrkidx{l}$).

\subsubsection*{Drift coefficient of the effective dynamics} 
The estimated drift $\hat{\mb}\lndmrkidx{ l}$ is obtained as the slope (in $t$) of $\mm\lndmrkidx{l}_t$ in \cref{e:meanAndCovBurst}, via a weighted linear regression: with $\{t_m\}_{m=1}^M$ equispaced in $[\tau_{\min},\tau_{\max}]$,
\begin{align} 
     \blN := \frac{\sum_{m=1}^M(\mmlNt{t_m} -\barmmlNM)(t_m-\bar t_M)}{\sum_{m=1}^M(t_m-\bar t_M)^2},  \label{e:drift-est}
\end{align}
where $\barmmlNM:= \frac{1}{M}\sum_{m=1}^M\!\hat{\mm}\lndmrkidx{l}_{t_m}$ and $\overline t_M=\frac1M\sum_{m=1}^M\! t_m$. Fig. \ref{f:peanutM},\ref{f:peanut-singular} show norm and direction of $\blN$ for the \peanut system.

\subsubsection*{Diffusion coefficient of the effective dynamics and local slow variables} 
Similarly, the local diffusivity $\Lambda\lndmrkidx{ l}$ of the slow effective dynamics is estimated as the slope (in $t$) of $C(\mz\lndmrkidx{l}_t|\mz\lndmrkidx{l}_0)$ in \cref{e:meanAndCovBurst}:
\begin{align}
     \LambdalN :=  \frac{\sum_{m=1}^M(\ClN_{t_m} -\ClNbar_M)(t_m-\bar t_M)}{\sum_{m=1}^M(t_m-\bar t_M)^2}\,, 
\label{e:Lambda_Nl}    
\end{align}
where  $\ClNbar_M = \frac{1}{M}\sum_{i=1}^M\ClN_{t_m}$. While $\LambdalN$ is typically not low-rank, for $N$ large enough, with high probability (w.h.p.), its top $d$ singular values may be well-separated from the others, yielding an estimate of the intrinsic dimension of \IM (a dynamics-driven analogue of Multiscale SVD \cite{LMR:MGM1}); this is case in our examples (see \cref{f:peanut-singular},\ref{f:halfmoon20},\ref{f:butane}). We project $\LambdalN $ onto the space of rank $d$ matrices by truncated SVD:
\begin{equation}
\LambdaldN:=\Projrkd\LambdalN=\Uslowl\hat\Sigma^{l}_{d}(\Uslowl)^T\,,
\label{e:LambdadN}
\end{equation}
where $\Projrkd$ denotes the projection onto rank $d$ matrices (positive semidefinite in this case), $\smash{\Uslowl\in\mathbb{R}^{D\times d}}$ orthogonal and $\Sigma^{l}_{d}\in\mathbb{R}^{d\times d}$ diagonal with the first $d$ singular values of $\LambdaldN$.\footnote[2]{We note here that we tried other approaches towards estimating $\LambdaldN$, for example by attempting to solve a least squares problem in the space of positive-definite matrices of rank $d$ directly, ideally with respect to a natural Riemannian metric on that space. While natural, this was significantly more computationally expensive, and it did not lead to significantly different results. 
} Let $\HldN\!\!:=(\LambdaldN)^\frac12$ be the (positive) square root of $\LambdaldN$. 

\subsubsection*{Covariance of the fast dynamics} 
While $\LambdaldN$ suffices to estimate a local tangent plane to \IMmth, the affine projection $P\lndmrkidx{ l}$ of the fast dynamics onto that plane, consistent with the dynamics, requires more information, as it is typically {\it{not}} an orthogonal projection. To estimate the kernel of $P\lndmrkidx{ l}$, i.e. the set of directions ``along which'' the dynamics near $\lndmrk{l}$ should be projected, we first estimate the covariance matrix $\GammalN$ in \cref{e:meanAndCovBurst} as
\begin{align} \label{e:cov-intercept}
     \GammalN :=\ClNbar_M -  \LambdalN \bar t_M\,,
\end{align}
and then let the estimated fast directions to be the span of the $D-d$ eigenvectors of $\GammalN$ with largest eigenvalues, which we group as columns of an orthogonal matrix $\Vfastl$. See \cref{f:ATLASmain},\ref{f:peanut-singular} for the case of the \peanut system. Since $\sigma_{D-d+1}(\Gamma)=0$ (see \cref{e:meanAndCovBurst}), $\sigma_{D-d+1}(\GammalN)\ll\sigma_{D-d}(\GammalN)$ w.h.p., for $N$ large enough. In practice, not all $D-d$ dimension may be fast modes, and we may truncate at the first $d_f\le D-d$ significant eigenvectors (e.g. in the oscillating half-moon system below).
 
\subsection{Construction of a sketch of the \IM\ \IMmth}
\label{s:IMestimation}
We now utilize the quantities estimated above to construct a sketch \hatIMmth of the \IM, consisting of a set of portions of well-distributed affine approximate tangent planes.

\subsubsection*{Landmarks}
The initial conditions $\{\mz\lndmrkidx{l}_0\}_{ l=1}\lndmrkidx{L}$ of the bursts $\{\mathcal{B}_ l\}_{ l=1}\lndmrkidx{L}$ are not assumed to be on the unknown \IMmth, nor well-distributed on it. We construct a set of points, called landmarks, on our estimate of \IMmth. From \cref{e:meanAndCovBurst}, replacing the quantities involved by their empirical counterparts estimated above, for each $ l=1,\dots,L$ we define the landmark $\lndmrk{l}$ as
 \begin{align} \label{e:lndmrk}
        \lndmrk{l} := \barmmlNM-\blN \bar t_M\,.
\end{align}

\subsubsection*{Local tangent plane to \IMmth}
We obtain an approximate tangent space $\tanPlaneldN$ at $\lndmrk{l}$:
\begin{equation}
\tanPlaneldN := \mathrm{span(cols(}\Uslowl))\,.
\label{e:tanPlanldN}
\end{equation}

\subsubsection*{Dynamics-driven oblique projections onto a local tangent plane}
We obtain the oblique affine projection $\PNlop$ onto $\tanPlaneldN$ centered at $\lndmrk{l}$: 
\begin{align} \label{e:projIM}
    \PNlz:=\Uslowl(\Uslowl)^T(\hatEl(\hatEl)^T)^\dag(\mz-\lndmrk{l})+\lndmrk{l}\,,
\end{align}
with kernel $\mathrm{span}(\mathrm{cols}(\Vfastl))$, where $\hatEl:=[\Uslowl,\Vfastl]$ and $\dag$ is the Moore-Penrose inverse. 

\subsubsection*{A dynamics-adapted metric}
We introduce a quasi-metric adapted to the dynamics, then we discard ``redundant'' landmarks that are too close to others in order to create a well-distributed set of landmarks, which, together with the approximate tangent planes estimated above, gives a parsimonious sketch of the estimated \IM \hatIMmth.

Consider the local Mahalanobis similarity based on the quadratic form associated with the effective diffusivity $\LambdaldN$
\begin{equation}
\hattildemetric^2(\mz, \lndmrk{l}) := \frac1{\chi^2_d(p) }{ (\PNlz - \lndmrk{l})^T (\LambdaldN)^\dag (\PNlz - \lndmrk{l})}\,,
\label{e:dtilde}
\end{equation}
for $\mz$ such that $\|\mz-\lndmrk{l}\|\lesssim R\sqrt{\tau}$; otherwise we set $\hattildemetric(\mz, \lndmrk{l})=R\sqrt\tau$. In practice, $R$ is set as 10. $\chi^2_{d}(p)$ is the quantile function at level $p$ of the $\chi^2$ distribution with $d$ degrees of freedom (we set $p=0.95$ throughout). 

Unlike Euclidean distance, $\hattildemetric$ accounts for the anisotropy of the dynamics on \hatIMmth; a similar distance, without the oblique projection, was used, for example, with different objectives, in \cite{DiffusionPNAS} for manifold learning, and in \cite{Singer2009} in the context of dynamical systems. 
We symmetrize $\hattildemetric$ on $\hatIMmth$ by letting $\hatmetric(\lndmrk{l'}, \lndmrk{l}):=\max\{\hattildemetric(\lndmrk{l'}, \lndmrk{l}),\hattildemetric(\lndmrk{l}, \lndmrk{l'})\}$. 
A $\sqrt{t}$-neighbourhood of $\lndmrk{l}$ is defined as $\metricneighborhood(\lndmrk{l}, \sqrt t):= \{\mz\in T_{\lndmrk{l}}\cM_{\epsilon}: \hattildemetric(\mz, \lndmrk{l})< \sqrt{t}\}$: it approximates the set of points reachable from $\lndmrk{l}$ in time $\approx t$ with probability at least $p$. This distance disregards the drift term; this choice reduces the asymmetry in the definition of $\hattildemetric$ and in the quasi-metric property, and is reasonable for diffusion-dominated dynamics see \cref{appx:assumption}). Fig.\ref{f:ATLASmain} (4th inset) visualizes the ellipsoids corresponding to the quadratic form induced by $\LambdaldN$, for the \peanut system.

\subsubsection*{A well-distributed net of landmarks}
We now reduce the number of landmarks to a near-minimal number that still, together with their corresponding neighborhoods of radius $\cO(\sqrt\tau)$, cover the estimated invariant manifold \hatIMmth, by discarding landmarks that are too close to each other. Before this process, we assume here that the collection of $\sqrt{\tau/2}$-neighborhoods, of the landmarks $\{\lndmrk{l}\}_{l=1}^L$ covers the invariant manifold \hatIMmth. While this typically cannot be ascertained in practice, by default ATLAS will be run in ``exploration mode'', which augments \hatIMmth on-the-fly (see \cref{s:explorationmode}): in that case our arguments here apply to the current \hatIMmth during exploration. For $l=1,\ldots$, we remove $\lndmrk{l'}$ if $\hatmetric(\lndmrk{l}, \lndmrk{l'})\le(1-{1}/{\sqrt{2}})\hat\kappa\sqrt{\tau}$ and $l'>l$, where $\hat{\kappa}$ is the scaling constant. In practice, $\hat\kappa$ is set as 1. When this procedure terminates, we are left with $L'\le L$ landmarks (unused landmarks are discarded). We assume, without loss of generality, that $\smash{\{\lndmrk{l}\}_{ l=1}\lndmrkidx{L'}}$ is the reduced collection of landmarks, and to simplify the notation, we will let $L'=L$ in what follows.
These landmarks, under suitable assumptions,
(i) are well-separated: for $ l\neq l'$,  $\hatmetric(\lndmrk{ l'}, \lndmrk{l})\gtrsim\sqrt{\tau}$;
(ii) provide a $\cO(\hat\kappa\sqrt{\tau})$-cover for \IMmth, in the sense that for any $\mz\in\hatIMmth$ there exists $ l$ s.t.  $\hattildemetric(\mz,  \lndmrk{l})\lesssim\sqrt{\tau}$;
(iii) are well-distributed, in the sense that for any $\lndmrk{l}\in \mathcal{T}$, where $\mathcal{T}$ is any connected component of the invariant manifold \IMmth satisfying suitable constraint, there exists $ l'$ such that  $\hatmetric(\lndmrk{ l'}, \lndmrk{l})\lesssim\sqrt{\tau}$. These constants $R, \hat\kappa$ used above could be made explicit, but they depend on quantities typically unknown in practice, such as the curvature of $\IMmth$.

These steps define a collection of charts, each centered at one of the landmarks $\lndmrk{l}$, with an associated oblique projection $\PNlop$ with range $\tanPlaneldN$, an effective drift $\blN\in\mathbb{R}^D$ and effective diffusivity $\LambdaldN\in\tanPlaneldN$. The reader may wish to revisit \cref{f:ATLASmain},\ref{f:peanutM},\ref{f:peanuteigen},\ref{f:peanut-singular}, that visualize these objects.

\subsection{The ATLAS process and simulator for the reduced effective (slow) dynamics}
\label{s:ATLASsimul}
The final step is to smoothly connect both the geometric and dynamic objects estimated so far at the landmarks, in order to obtain a smooth effective invariant manifold \hatIMmth and an It\^o process constrained on it, with the fast modes averaged at the timescale $\tau$, together with a numerical scheme for its simulation.
This smoothing is achieved by a weighted average:
for $\mz\in\mathbb{R}^D$, we let $w\lndmrkidx{ l}(\mz):=\exp(-\hattildemetric(\mz, \lndmrk{l})/\sqrt{\tau})$ and  $\mathcal{N}_\tau^{\mathcal{A}}(\mz):=\{\lndmrk{l}\,:\, \hattildemetric(\mz,\lndmrk{l})\le 2\hat\kappa C\sqrt{\tau}\}$, $Z(\mz):=\sum_{ l\in\mathcal{N}_\tau^{\mathcal{A}}(\mz)}w\lndmrkidx{ l}(\mz)$, and then define:
\begin{eqnarray}
\ProjA(\mz) 	 &:=& \frac{1}{Z(\mz)}\sum\nolimits_{l\in\mathcal{N}_\tau^{\mathcal{A}}(\mz)} \PNlz w\lndmrkidx{l}(\mz) 						\,,	\label{e:zATLAS} \\
\bA(\mz) 				&:=& \frac{1}{Z(\mz)}\sum\nolimits_{l\in\mathcal{N}_\tau^{\mathcal{A}}(\mz)} \blN w\lndmrkidx{l}(\mz)   						\,,	\label{e:meanAndCovBurstATLAS} \\
\LambdaA(\mz) 		&:=& \frac{1}{Z(\mz)}\sum\nolimits_{l\in\mathcal{N}_\tau^{\mathcal{A}}(\mz)} \LambdaldN  w\lndmrkidx{l}(\mz)  					\,,	\label{e:diffDATLAS} \\
\HA_d(\mz) 			&:=& (\Projrkd \LambdaA(\mz))^\frac12 														\,.	\label{e:diffATLAS}
\end{eqnarray} 
This defines the ATLAS stochastic process $\mzAt$, on the estimated \IM $\hatIMmth:=\ProjA(\mathbb{R}^D),$ as the It\^o diffusion with drift $\bA$ and diffusion coefficient $\HA_d$. To simulate this process we use the Euler-Maruyama scheme with re-projection on $\hatIMmth$, with time-step $\lambda\tau$ and $\Delta W_{\lambda\tau}\sim \cN(0, \lambda \tau I_d)$:
\beq 
     \mz^{\mathcal{A}}_{t+\lambda\tau} = \ProjA\big(\mzAt + \bA(\mzAt) \lambda\tau + \HA_d(\mzAt)\Delta W_{\lambda\tau}\big)\,,
     \label{e:ATLAStime-step}
\eeq

In all our experiments, $\lambda=1$, i.e. the time-step of the ATLAS simulator is equal to the timescale $\tau$; in particular, it is independent of, and may be much larger than, the time-step $\delta t$ of the original black-box simulator, which is typically $\lesssim\epsilon$.

\subsection{Refinements to the estimation procedure}
\label{s:refinement}
When the initial conditions for the bursts are far away from the \IM \IMmth (e.g. in the oscillating half-moon example below), or the timescale of separation in the original system is not very large (e.g. in butane example below), it may take a time longer than $\tau$ to relax onto \IMmth. In these cases we perform multiple rounds of the estimation phases, starting each round with initial conditions for the bursts given by the landmarks estimated in the previous round. This refinement process stops when the relative differences of estimated parameters is within 5\% (in our examples, this is achieved in less than $10$ rounds). 

One further step of refinement, after the above ones, may be needed when the linear approximation of the geometry or of the dynamics may not be locally accurate, because \IMmth has large curvature or the effective drift term has large gradient (e.g. in the oscillating half-moons and butane examples below). 
In this case, since the estimated drift term is computed at the timescale $[\tau_{\min{}},\tau_{\max{}}]$, the landmarks in the final stage are refined to be $\lndmrk{l}:= \barmmlNM-\blN (\bar t_M - \tau_{\min{}})$, instead of \cref{e:lndmrk}.

\section{ATLAS in Exploration Mode}
\label{s:explorationmode}
In the construction of ATLAS presented above, initial conditions for the bursts of simulation were sampled from a provided measure $\mu_0$ on the state space. Ideally such measure is well-distributed on \IMmth, e.g. close to the stationary distribution of the effective reduced process, or, at least, such that the set of local means $\barmmlNM$ of bursts started i.i.d. from $\mu_0$ are well-distributed on \IMmth. However, this is too much to hope for in many practical situations, and it is highly desirable to be able handle more general $\mu_0$'s.

There are at least two, not-mutually-exclusive, ways of proceeding. The first one is to use any of many existing techniques  aimed at efficiently sampling the effective state space, to obtain a $\mu_0$. The literature on this subject is vast, including, among many, \cite{ChiavazzoE5494, yannis2009, plumed2,clementiJPC2013, Chen3235}. These techniques often design a bias of the dynamics to ensure rapid exploration, yielding samples with coverage of the effective state space. In some remarkable cases the samples from this biased process allow to recover statistics of the original dynamics, e.g. the stationary distribution in the case of MCMC. However, even when the stationary distribution is recovered, the biased dynamics often does not preserve other dynamic properties such as mean residence times or transition rates, important in applications. Yet other techniques in this broad family require a target statistics to be computed, and then are designed to achieve accuracy in the estimate of that statistics, but in general no other ones. There is typically a strong tension between the attempt to speed up exploration, and the ability to correct the biased sampling to obtain consistent estimates of statistics of interest. We note that, crucially, this tension does {\it{not}} arise in ATLAS: $\mu_0$ needs to have coverage, but has no relationship with the dynamics; it is only used as a starting point for ATLAS, which will then estimates consistently the effective dynamics, and from it the stationary distribution and many other dynamical properties, such as mean residence times and transition rates, as demonstrated in the numerical experiments in sec.\ref{s:numericalExp}.

The second way is to extend ATLAS to run in {\it{exploration mode}}: upon a first round of learning starting from a $\mu_0$ with poor coverage, ATLAS runs trajectories till they exit the current, partial estimate of \IM, and updates itself, accurately but efficiently, by simulating new paths of the original process at those new exit locations. 
In detail, suppose we have constructed ATLAS $\mathcal{A}_1$, starting from a set of bursts $\{\mathcal{B}^l\}_{l=1}\lndmrkidx{L}$, therefore obtaining the process $(\mzAtk1)_{t\ge0}$ on \IMmthepshatk1, perhaps from a small number $L$ (even $L=1$) of initial conditions, which are poorly distributed in the state space (e.g. supported in one or a few metastable states). While simulating $(\mzAtk1)_{t\ge0}$, ATLAS checks if the distance between $\mzAtk1$ and its closest landmark, in the \hattildemetric ``metric'', is larger than some threshold $d_{\text{thr}}=\cO(\sqrt\tau)$: if this is the case, then $\mzAtk1$ is beyond the current ``domain of expertise'' of the ATLAS $\mathcal{A}_1$.
We now stop that process, and run a new burst $\mathcal{B}_*$ of simulations from this ``exit point'' $\mzAtk1$, estimate the local quantities of the dynamics there, and add a new landmark $\lndmrk{l+1}$, with its associated tangent plane, projection, and estimated effective drift and diffusion coefficients. This local and efficient update yields a new ATLAS process $(\mzAtk2)_{t\ge0}$ on an ``enlarged'' \IMmthepshatk2. This procedure is repeated, creating ATLAS processes that capture ever-increasing approximations $\IMmthepshatk1\subset\dots\subset\IMmthepshatk k\subset\dots$ of \IMmth, discovering rarer and rarer events, till a given computational budget (for example expressed in terms of total number of bursts used, which is the most expensive component in the construction) is exhausted, or long-enough trajectories are simulated with $(\mzAtk k)_{t\ge0}$ without leaving \IMmthepshatk k.  See details in \cref{s:SI:algos}. {\it{All our numerical experiments will be performed with ATLAS in exploration mode.}}

We provide a pertinent visualization in step 4 of \cref{f:ATLASmain}: at stage $k$ of exploration, we represent \hatIMmthk k, and a trajectory of $(\mzAtk k)_{t\ge0}$ which at some point leaves \hatIMmthk k (orange dot in the figure). The current ATLAS $\mathcal{A}_k$ stops there, and then will obtain a new burst of paths starting at that location, extract local estimates of the geometric and dynamics quantities needed, obtain \hatIMmthk {k+1}, and then continue. In the figure, the path is linearly interpolated between the points $(\mzA(p\lambda\tau))_p$; note that the ATLAS time-step $\lambda\tau$ leads (here, and in the other examples, $\lambda=1$); as expected, to steps of length comparable to that of the axes of the diffusion ellipsoids representing the level set $\tau$ of the quadratic form $\LambdaldN$.

Note that these additions happen during ATLAS simulations (with their large $\lambda\tau$ time-steps), taking advantage of the computational speed gains achieved by the ATLAS simulator. Therefore, in real clock time, the regions of \IMmth that are rarely visited with the original simulator, are more likely  to be sampled quickly in ATLAS exploration mode. Note how this differs from existing techniques based on ideas of importance sampling: at no point is the dynamics of ATLAS biased, and at any stage the dynamics is consistent with the underlying effective dynamics, by construction. This procedure can be very effective, measured in real clock time, in discovering relatively rare events, such as transitions between metastable states, while updating ATLAS, seamlessly. We also note that this procedure is parallelizable across multiple paths of the current ATLAS, provided one checks that the regions being added are far enough from each other to avoid the simulation of the bursts of simulations at nearby locations: these ``conflicts'' are likely rare in high-dimensional state spaces, at least till ATLAS has explored the vast majority of it.

Finally, it is certainly possible to apply a multitude of techniques and heuristics (e.g. see \cite{ChiavazzoE5494} and references therein) to bias the ATLAS simulator itself during exploration, i.e. combine the two approaches described in this section; once new regions are explored with a biased ATLAS, and charts created (with new local bursts initialized in those regions) and incorporated into ATLAS, then the ATLAS simulator can be run in unbiased mode and will be consistent with the effective dynamics. This decoupling of exploration and consistent estimation of the dynamics is a crucial property of ATLAS, and it is very efficient as the information from the expensive simulations of bursts of trajectories is fully re-used. {This strategy together with parallel learning across multiple paths are particularly helpful when the effective dynamics have a significant amount of meta-stable states.}

\section{Properties of ATLAS}
\label{s:propertiesATLAS}

\subsection*{Avoiding the curse of dimension} The input to ATLAS is random, so are the $L$ initial conditions and the $N$ paths in each burst.  It is natural to ask how many short trajectories in each burst are needed to make sure the random error of all the local quantities estimated by ATLAS is small w.h.p. In particular, it is important to assess how $N$ should scale with the dimensions $D$ of the state space, $d$ of \IMmth, $d_f$ of the fast modes with large magnitude. Using concentration inequalities for high-dimensional vectors and matrices that are concentrated near low-dimensional spaces \cite{vershynin_2018}, it is possible to show that the approximation error between the empirical estimates of $\blN$, $\LambdaldN$ and $\lndmrk{l}$ is smaller than $\eta$, with high probability, as soon as $N\gtrsim {d(d+d_f)}/{\eta^2}$, where $d_f$ is the number of fast modes with large magnitude.

The approximation error of the direction of the fast modes appears to require larger sample size, e.g. 
$N\gtrsim {(d+d_f)^2\ln{D}}/{\eta^2}$ samples appear needed to obtain, w.h.p., $\|\sin\Theta(\Vfastl,\Vfasttruel)\|_{\mathrm{F}}\le\eta$. It is worthwhile to note, though, that this still depends only very weakly on $D$, and only quadratically on the intrinsic dimension $d$, and it is still quite benign as soon as $d_f\ll D$.

To summarize,  there is no curse of dimensionality -- i.e. a requirement of a number of samples exponential in $D$ -- when estimating the local quantities above. 
%

\subsection*{Robustness to model error \& nonlinearities in the fast modes}
The type of stochastic systems for which ATLAS is expected to perform well have been described in sec.\ref{s:FastSlowSDEs}. In particular, locally we assumed the existence of an invariant manifold in the observed space, on which the slow dynamics $(\sovar_t)_{t\ge0}$ takes place, and such that the fast dynamics $(\fovar_t)_{0\le t\le \tau}$ conditioned on $\sovar_t=\mzIC$ approximately is an O-U process on the subspace $\smash{\fsubspace_{\mzIC}}$. In the latent space, the SDEs are those in \cref{e:master}, and one then linearizes locally the observation map $\varphi$ (or, rather, $\varphi_\alpha$, as in sec.\ref{s:FastSlowSDEs}) to obtain approximate local equations in the $\mz$ variables. ATLAS averages the observed process $(\mz_t)_{t\ge0}$ at timescale $\tau$ to obtain the reduced effective dynamics on \hatIMmth. 

ATLAS is quite robust to these assumptions. One reason is that ATLAS uses mainly information at timescale $\tau$: details, such as nonlinearities, or lack of regularity, of the original process $\mz_t$ below that timescale are averaged out, possibly leading to effective processes at the timescale that are amenable to approximation by ATLAS. 
In a forthcoming work, we prove results in this direction, under technical conditions on the coefficients $f,g,F,G$ of the SDEs \cref{e:master}, on the regularity of the map $\varphi$, and on the stationary measure $\nu(\fvari|\sovar_t=\mzIC)$ of the (fast) displacement process $\fvari$ conditioned on $\mzIC$.

This robustness is reflected in the results for both the second and third examples we consider in sec.\ref{s:numericalExp}, which are both significant perturbations of the basic model. In the ``oscillating half-moons'' example, a high-dimensional analogue of an example considered in \cite{Singer2009,Dsilva2016}, the fast displacement process $\fvari$ is nonlinear and not constrained to an affine subspace, but on a curved ``half-moon''-shaped manifold. In the butane model, the fast mode has both large and small nonlinear components; its slow manifold is also highly curved, with effective drift having large gradient. Nevertheless, ATLAS provides accurate estimates of the behavior of these systems, both at timescale $\tau$ and at very large timescales, providing accurate estimates of the stationary distribution, mean residence time and transition rates between metastable states.

\subsection*{Computational complexity and simulation speed-up} 
The input data to ATLAS are the bursts of trajectories $\{\mathcal{B}_ l\}_{l=1}^L$, of time-length $\approx\tau$. The cost of obtaining one time-step from the black box simulator is at least of order $D^2$. The time-step $\delta t$ of the simulator needs to be $\lesssim\epsilon$ due to having to resolve the fast modes. The total number of short paths collected is equal to \#landmarks$\times$\#paths per landmark$=L\times N$. Therefore the total computational cost of obtaining the bursts is at least $\cO(\frac\tau\epsilon D^2 LN/c)$, where $c$ is the number of parallel cores. Constructing ATLAS requires $\cO(D^2dN)$ calculations to estimate local means, covariances, effective drift, effective diffusion coefficients, landmarks and tangent planes; $\cO(C^d D L \log L)$ for constructing and organizing the landmarks using, for example, cover trees \cite{LangfordICML06-CoverTree}. A time-step of the ATLAS simulator as in \cref{e:ATLAStime-step}, which has time-length $\lambda\tau$, has cost $\cO(C^d Dd^2)$ by using iterative SVD combining eq.~(\ref{e:diffDATLAS},\ref{e:diffATLAS}), where $C^d$ corresponds to the number of landmarks in $\mathcal{N}_\tau^{\mathcal{A}}(\mz)$. Therefore, simulating a path of time-length $T$ would have computational cost $\cO(D^2T/\epsilon)$ with the original simulator, and $\cO(C^d Dd^2 T/\tau)$ with ATLAS. This is a dramatic speed-up when $\epsilon\ll\tau$ and $d\ll D$. This is very useful when many long paths are needed to estimate dynamical quantities of interest.

\section{Applications of ATLAS}
\label{s:ATLASuses}

\subsection*{Estimation of large-time dynamical properties}
ATLAS may be used to simulate long paths efficiently, and therefore estimate important properties of the system, such as its stationary distribution, residence times from certain regions of state space (e.g. metastable states), and transition rates between them. Our numerical experiments in sec.\ref{s:numericalExp} show that such large-time quantities may be estimated accurately by ATLAS, even when run in exploration mode. Note that ATLAS is constructed using only local information, at timescale $\tau$, that may be easily collected in parallel; yet the effects of the multiple estimation and numerical simulation errors do not appear to compound in these estimates of large-time quantities \cite{CM:ATLAS}.

\subsection*{ATLAS, approximate generators, eigenfunctions and eigenvalues}
The ATLAS process $(\mzAt)_{t\ge0}$ may be used to approximate the generator of the effective slow dynamics and its spectral components, including eigenvalues and eigenvectors, especially the low-frequency ones. It may serve as a black-box for matrix-vector multiplication in iterative eigensolvers. In general, ATLAS may be used to compute approximations of $\mathbb{E}[h(\sovar_t)]$, for sufficiently regular observables $h$.

\subsection*{Markov State Models (MSMs) from ATLAS}
In MSMs \cite{MSM-review} one constructs (i) a partition of state space $\{C_k\}_{k=1}^K$ and (ii) a Markov transition matrix $P\in\mathbb{R}^{K\times K}$ with $P_{kk'}$ being the probability of transitioning from $C_k$ to $C_{k'}$ in one MSM time-step. 
MSMs may be ``large-timescale MSMs'', where each $C_k$ corresponds to a metastable state and the MSM models the rare transitions between them, and ``small-timescale MSMs'', where $K$ is large and the $C_k$'s are small regions of state space.

Large-timescale MSMs may be constructed if the metastable states are known and a large number of transitions between them are observed. Since these transitions are rare, by definition of metastability, this construction is very expensive in general; however, ATLAS can help identifying metastable states and estimating transition rates efficiently.

Small-timescale MSMs are very flexible tools, and as $K\rightarrow\infty$ the transition matrix $P$ approximates in a suitable sense the generator of the process, and convergence is (under suitable assumptions) strong enough to guarantee convergence of the slow eigenfunctions of $P$ to those of the generator of the process. These eigenfunctions, and the corresponding eigenvalues, yield important information about the process, including metastable states. 
However, the construction of the local clusters $C_k$ is crucial, and many recipes exist \cite{frankJCP2013, DMD-textbook}. This is a challenging task and typically cursed by the ambient dimension $D$. Many existing techniques require, in order to be of any practical value, the a priori knowledge of a suitable small number of slow variables on which the process is projected, and in which the construction of the $C_k$'s is performed \cite{MSM-review,jns-review}. In particular, we are not aware of techniques for efficiently constructing the $C_k$'s in the situation where there are many fast modes, possibly with large amplitude. In this context, ATLAS naturally constructs the small timescale MSMs (at timescale $\tau$), in a principled and well-organized fashion, with soft instead of hard partitions, which may diminish the memory effect. ATLAS also uses dynamics-adapted oblique projections and the corresponding estimated local \IM{} to reduce the dimension, without needing slow variables as inputs. In our experiments, $C_k$'s in the MSM correspond to the Voronoi cells, in the $\hattildemetric$ ``metric'', of the landmarks\footnote[3]{of course, we do not explicitly construct such cells; we only need a function mapping a state $\mz$ to the index of cell it belongs to, and this is achieved by finding the nearest landmark in the $\hattildemetric$ ``metric''.}, and the transition matrix is estimated by running ATLAS trajectories of length $\cO(\tau)$ (see appendix \ref{s:SI:algos}). We may use the small time-scale MSMs to compute approximate slow eigenfunctions and eigenvalues of the system and estimate the number and locations of metastable states, and then construct the large-timescale MSMs.

\section{Numerical experiments}
\label{s:numericalExp}
We construct ATLAS for three model systems: ``\peanut'', ``oscillating half-moons'' and ``butane model''. We evaluate its performance in multiple ways: first of all, {{against analytically-derived reduced models}} with analytical approximations to the \SM \SMmth, effective drift and diffusion coefficients (see \cref{appx:example}). 
For the first two examples, the effective dynamics are calculated in the limit $\eps\rightarrow 0$; for butane, the effective dynamics are chosen to be the dihedral angle dynamics. It is important to remark that these are not the true effective dynamics on the \IM \IMmth at timescale $\tau$, which is what ATLAS approximates, and are also not amenable to analytical calculation for finite $\eps$. Although with this caveat, we regard them as analytical approximations sufficient as a first check on the quality of the ATLAS process for the local statistics, and report in \cref{t:error} the estimator errors for drift, diffusion, invariant manifold and tangent spaces, between ATLAS and these analytically-derived reduced models (details in \cref{appx:error}). 

We also study the accuracy of ATLAS in estimating key medium- and  large-time statistics of the dynamics, in particular the stationary distribution, mean residence times (MRTs) and transition rates for metastable states, and MRTs in regions of state space that are not necessarily metastable. In each example, we repeat the construction of ATLAS $10$ times, to assess the variability over the random observed data. 

We visualize the \IM \hatIMmth for each example, as well as key quantities including the stationary distribution and eigenfunctions of MSMs; in these plots we use suitable parametrizations (that of course were not used nor known to ATLAS). 
Further details and figures for the models are available in \cref{appx:example}.
\begin{figure}
\center
\includegraphics[width =  0.6\textwidth]{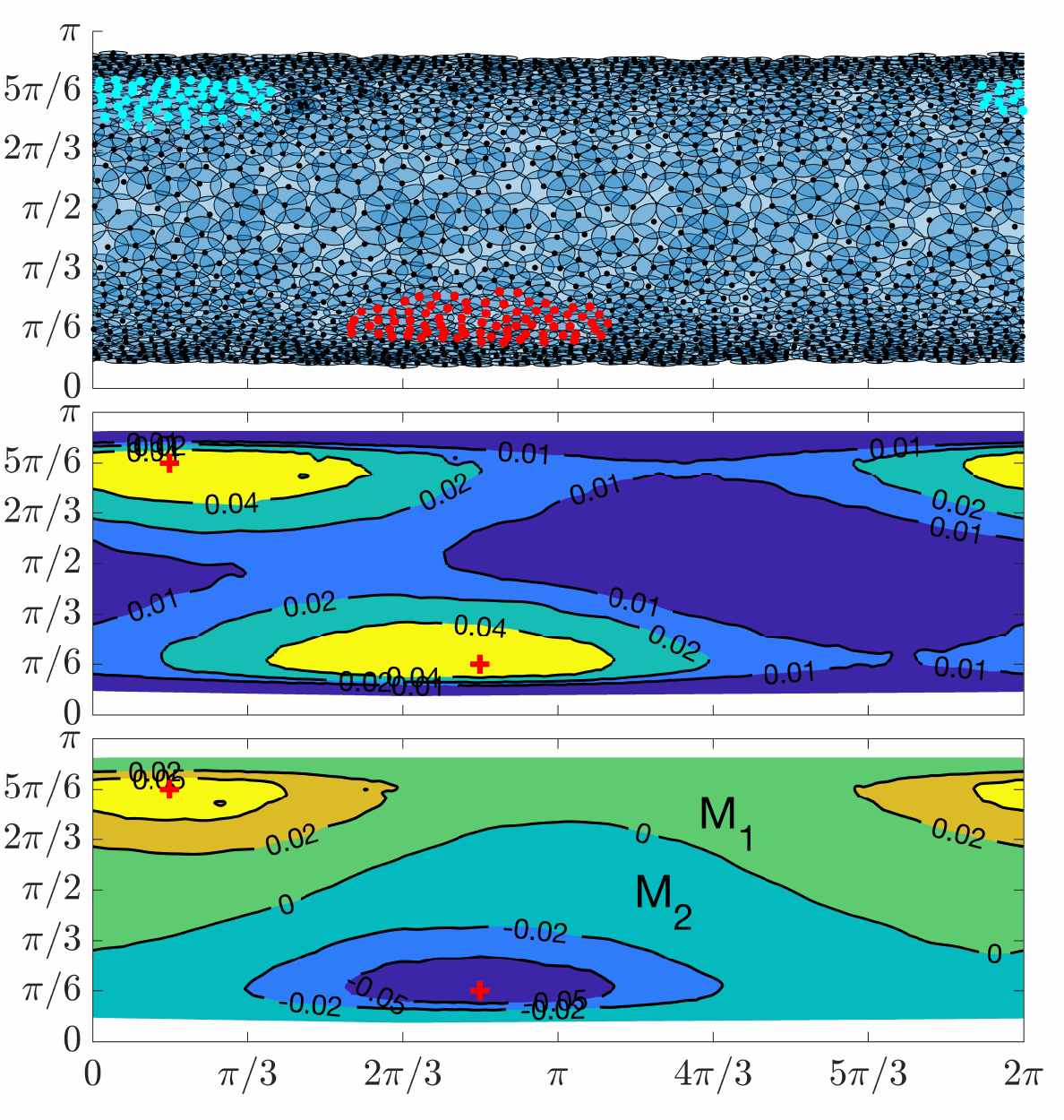}
\vskip-0.5cm
\caption{\Peanut visualized in the plane of the parametrization $(\phi,\theta)$ of \IMmth. {\bf{Top}}: landmarks and their neighborhoods $\metricneighborhood(\lndmrk{l}, \sqrt \tau)$. 
We also show two sets $S_{\text{red}}$ and $S_{\text{cyan}}$ around the fixed points in the reduced deterministic system, marked with red crosses in the other insets. {\bf{Center}}: eigenfunction $\varphi_1$ of the MSM, with eigenvalue $\lambda_1=1$.  {\bf{Bottom}}: eigenfunction $\varphi_2$, with eigenvalue $\lambda_2=0.9995$.}
\label{f:peanuteigen}
\vskip-0.55cm
\end{figure}

\subsection{\Peanut system}
\label{s:expPeanut}
We start with the \peanut system, introduced in sec.\ref{s:ATLASconstruction}. Its  governing equations, expressed in spherical (latent) coordinates, are
\begin{small}
\begin{equation}
\begin{aligned}
&\rd r = -\frac{c_1(r- R(\theta))}{\eps r}\rd t + \frac{c_2}{\sqrt{\eps}r} \rd W_1 \\
&\rd \theta = \frac{c_3\cos(3\theta)}{r\sin(\theta)} \rd t+\frac{c_4\sin(\theta)}{r} \rd W_2 \\
&\rd \phi = \frac{c_5\sin(\phi+\theta)}{r} \rd t + \frac{c_6}{r}\rd W_3\,.
\end{aligned} \label{peanut}
\end{equation}
\end{small}
The fast variable is the radial coordinate $r$; the slow variables are the angles $\phi,\theta$. The \SM{} (in the limit $\eps\rightarrow0$) is $R(\theta)= \sqrt{a_1+a_2\cos^2(\theta)}$, visualized in \cref{f:peanutM}.
The observations $\mz$ are in Cartesian coordinates, each of which contains a mix of nonlinearly coupled slow and fast components.
 Note that the drift diverges near the poles, creating a strong repulsion, and is relatively small in other wide regions of the state space, creating entropic barriers \cite{ bicout_szabo_2000}. 

The dominant local PCA mode only captures the fast direction, due to its large amplitude, and fails to identify the slow variables, which are also not orthogonal to the fast ones. 
ATLAS successfully estimates that the \IM is two-dimensional, and identifies the separation timescale $\tau$ (see \cref{appx:peanut} and \cref{f:peanut-singular}). ATLAS yields an accurate estimation of the effective drift and diffusion terms, as well as of \IMmth (see \Cref{t:error}). 
We visualize in fig.\ref{f:peanuteigen} the $\sqrt{\tau}$-neighborhoods over \IMmth (unwrapped in the $(\phi,\theta)$ coordinates for clarity),
reflecting the ellipsoids associated with the diffusion coefficient.
At the time we terminate exploration, as expected the only regions that are not covered are those around the south and north poles, which are very rarely visited.  
\begin{table}[] 
\centering
\caption{Mean residence times (MRTs) for \Peanut}
\begingroup
\setlength{\tabcolsep}{2pt} 
\begin{tabular}{c|c|c|c|c|c|c|}
\cline{2-7}
 & \multicolumn{3}{c|}{First exit from $C_1$ or $C_2$} & \multicolumn{3}{c|}{First exit from $M_1$ or $M_2$} \\ \cline{1-7}
\multicolumn{1}{|c|}{ \diagbox[dir=NW]{IC}{Sim.}} & \begin{tabular}[c]{@{}c@{}}Original\\ $\mathcal{S}$\end{tabular} & \begin{tabular}[c]{@{}c@{}}ATLAS\end{tabular} & \begin{tabular}[c]{@{}c@{}}ATLAS\\ orth. proj.\end{tabular} & \begin{tabular}[c]{@{}c@{}}Original\\ $\mathcal{S}$\end{tabular} & \begin{tabular}[c]{@{}c@{}}ATLAS\end{tabular} & \begin{tabular}[c]{@{}c@{}}ATLAS\\ orth. proj.\end{tabular} \\ \hline
\multicolumn{1}{|c|}{$S_{\text{Cyan}}$} & \begin{tabular}[c]{@{}c@{}}$30.6$\\ $\pm0.4$   \end{tabular}  & \begin{tabular}[c]{@{}c@{}}$29.7$\\ $\pm0.4$   \end{tabular}   &  \begin{tabular}[c]{@{}c@{}}$28.9$\\ $\pm0.4$   \end{tabular}  & \begin{tabular}[c]{@{}c@{}}$905.6$\\ $\pm14.0$   \end{tabular}   &  \begin{tabular}[c]{@{}c@{}}$903.1$\\ $\pm 14.1$   \end{tabular}  &  \begin{tabular}[c]{@{}c@{}}$712.2$\\ $\pm10.9$   \end{tabular}   \\ \hline
\multicolumn{1}{|c|}{$S_{\text{Red}}$} &  \begin{tabular}[c]{@{}c@{}}$30.6$\\ $\pm0.5$   \end{tabular} &   \begin{tabular}[c]{@{}c@{}}$30.2$\\ $\pm0.5$   \end{tabular}   &  \begin{tabular}[c]{@{}c@{}}$29.4$\\ $\pm0.4$   \end{tabular} &   \begin{tabular}[c]{@{}c@{}}$1009.3$\\ $\pm15.6$   \end{tabular}   &  \begin{tabular}[c]{@{}c@{}}$1016.5$\\ $\pm15.8$   \end{tabular}   &  \begin{tabular}[c]{@{}c@{}}$791.2$\\ $\pm12.2$   \end{tabular}   \\ \hline
\multicolumn{1}{|c|}{Runtime} &0.39h  &0.02h  &0.02h  &    4.15h & 0.66h & 0.52h \\ \hline
\hline
\multicolumn{1}{|c|}{$S_{\text{Cyan}}$} & N/A & \begin{tabular}[c]{@{}c@{}}$0.002$\\ $\pm0.010$   \end{tabular}       &    \begin{tabular}[c]{@{}c@{}}$-0.026$\\ $\pm0.008$   \end{tabular} & N/A &   \begin{tabular}[c]{@{}c@{}}$0.010$\\ $\pm0.008$   \end{tabular}  & \begin{tabular}[c]{@{}c@{}}$-0.212$\\ $\pm0.008$   \end{tabular}  \\ \cline{1-7} 
\multicolumn{1}{|c|}{$S_{\text{Red}}$} & N/A &  \begin{tabular}[c]{@{}c@{}}$0.003$\\ $\pm0.013$   \end{tabular}  &   \begin{tabular}[c]{@{}c@{}}$-0.019$\\ $\pm0.010$   \end{tabular}  &  N/A&    \begin{tabular}[c]{@{}c@{}}$0.011$\\ $\pm0.009$   \end{tabular} &   \begin{tabular}[c]{@{}c@{}}$-0.211$\\ $\pm0.007$   \end{tabular}   \\ \hline
\end{tabular}
\endgroup
\caption*{{\textmd{\footnotesize{Top: MRTs from initial conditions in IC set till exit from specified sets; bottom: corresponding relative errors of MRTs.}}}}
\label{t:peanut}
\end{table}

The TICA method \cite{prl1994, frankJCP2013} is global and indicates that all observed coordinates are important; in particular the common approach of constructing MSMs in the TICA coordinates would be cursed by the ambient dimension. Here we construct MSMs using ATLAS. 
The top two eigenfunctions of the transition matrix of an MSM constructed from ATLAS on \hatIMmth are visualized in fig.\ref{f:peanuteigen}. 
The first eigenfunction $\varphi_1$ is (up to rescaling) the invariant distribution;
 the level set $\varphi_2=0$ partitions the state space into two metastable states $M_1$ and $M_2$. We also let $C_1:=\{\varphi_2>+0.02\}$ and $C_2:=\{\varphi_2<-0.02\}$; initial conditions for paths used in the computation of mean residence times (MRTs) will be from $S_{\text{cyan}}:=\{\varphi_2>0.05\}$ and $S_{\text{red}}:=\{\varphi_2<-0.05\}$ (see \cref{f:peanutM},\ref{f:peanuteigen}), where $\varphi_1$ is large.
 
 In \cref{t:peanut} we report the accuracy of ATLAS in estimating the MRTs in $M_1, C_1$ (resp. $M_2, C_2$) starting from set $S_{\text{cyan}}$ (resp., $S_{\text{red}}$). ATLAS yields $\le2\%$ relative error for these quantities, with runtime at least $6$ times smaller than original simulator $\mathcal{S}$; training time is about 21hrs. Of course, the transition rates between metastable states, which are determined by the mean residence times for double-well systems, are also very accurate.  
  Using orthogonal projections, instead of the ATLAS oblique projections, leads to a significant loss of accuracy in long-time observables (e.g. exit times from $M_1, M_2$).
The estimated $L^1$-norm of the difference of the density of the invariant distribution between original and ATLAS simulators is $ 0.107\pm0.009$.
\begin{figure}[t]
\centering
\includegraphics[width =  0.6\textwidth]{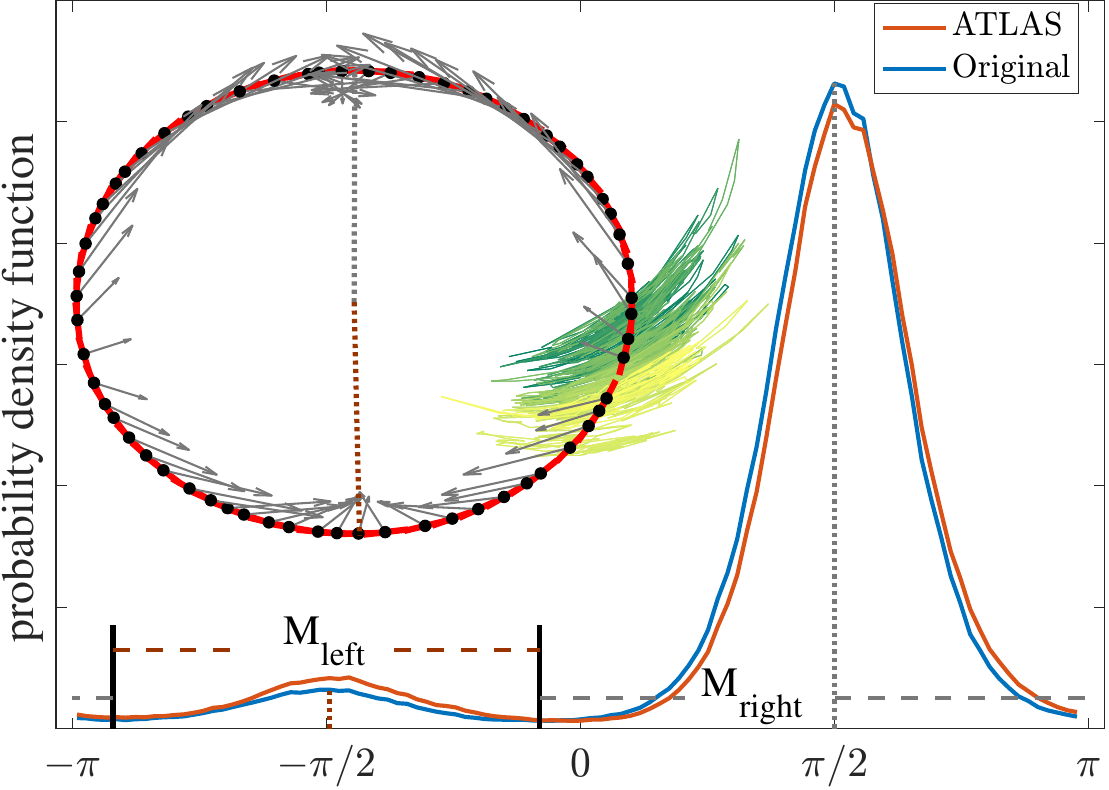}
\vskip-0.25cm
\caption{Oscillating half-moons. The short illustrative trajectory of time $\num{1e2}$ is plotted  in $(z_1,z_2)$ and colored according to the time $t$.  The landmarks (black dots) and their neighborhood (red lines) and effective drift direction (gray arrows) in $(z_1,z_2)$ are plotted in the left.  In the right, the smoothed histograms from the trajectories of time length $\num{8e6}$ generated by $\mathcal{S}$ and ATLAS simulator, projected with ATLAS's projection, are plotted in the coordinate of angle $\theta$.}
\label{f:halfmoon}
\begin{table}[H]
\centering
\caption{Mean residence time (MRT) for oscillating half-moons}
\begin{tabular}{c|c|c|c|}
\cline{1-4}
\multicolumn{1}{|c|}{ \diagbox[dir=NW]{IC}{Sim.}}  & Original $\mathcal{S}$ &ATLAS & ATLAS orth. proj. \\ \hline
\multicolumn{1}{|c|}{$M_{\text{Left}}$} &$918\pm  11$ & $902\pm11$  & $897\pm 11 $  \\ \hline
\multicolumn{1}{|c|}{$M_{\text{Right}}$} & $12552\pm 142$ & $11221\pm 128$ &  $10855\pm123$ \\ \hline
\multicolumn{1}{|c|}{Runtime} & 2.49h& 1.05h& 1.05h \\ \hline
\hline
\multicolumn{1}{|c|}{$M_{\text{Left}}$}  & N/A & $-0.02\pm 0.01$  & $-0.02\pm 0.01$ \\ \cline{1-4}
\multicolumn{1}{|c|}{$M_{\text{Right}}$} & N/A &$-0.13\pm 0.01$  &$-0.13\pm 0.01$ \\ \hline
\end{tabular}
\caption*{\textmd{\footnotesize{Top: MRTs from initial conditions in IC set till exit from specified sets; bottom: corresponding relative errors of MRTs. }}}
\label{t:halfmoon}
\end{table}
\vskip-1.25cm
\end{figure}

\subsection{Oscillating half-moons}
This is a multiscale stochastic system in $\smash{\mathbb{R}^2\times\mathbb{R}^{18}}$ that generalizes the one in \cite{Singer2009} to high dimensions. Its governing equations in latent coordinates are:
\begin{small}
 \begin{equation}
 \begin{aligned}
&\rd \theta = \left( a_1 + a_2\sin(2\theta)   +a_3\cos(\theta)\right) \rd t + a_4 \rd W_1, \\
&\rd r_1 = \frac{b_1}{\eps}\left(1-r_1\right)\rd t + \frac{b_2}{\sqrt{\eps}} \rd W_2, \\
&\rd u_i =  \frac{b_3}{\eps}(-u_i)\rd t + \frac{b_4}{\sqrt{\eps}} \rd W_i, \ \ \text{i=3,\dots, 20}.
\end{aligned}
\end{equation}
\end{small}
The observables in Cartesian $\mathbb{R}^{20}$ by
\begin{small}
\beq
\begin{aligned} 
z_1 = r_1\cos(\theta+r_1-1), z_2= r_1\sin(\theta+r_1-1),z_i= r_1+u_i\,,
\end{aligned}
\eeq
\end{small}
for $i=3,\ldots, 20$. The dynamics of the angle $\theta$ is that of an uneven double-well system with metastable states $M_{\text{Left}}$ and $M_{\text{Right}}$ around $\theta=\pm\pi/2$. The radial variable $r$ and other $u_i$'s evolve as O–U processes. The fast variables $r,u_2,\dots,u_{19}$ are nonlinearly coupled in the observed Cartesian coordinates.

A typical trajectory exhibits fast oscillations with a half-moon shape, far from a radial direction, while evolving slowly along the circular slow manifold driven by the double-well potential and diffusion along it (see in \cref{appx:halfmoon}). 

Local PCA again fails to detect the slow manifold (see \cref{f:halfmoon20}).
Notwithstanding the lack of linearity of the fast modes, ATLAS accurately identifies the invariant manifold and the effective dynamics on it, see \cref{t:error}. While the relative error of estimated drift and covariance matrix seems large ($32\%$ and $11\%$, resp.), if the error is measured only in the first two important coordinates -- since the error in the other $18$ dimensions does not contribute to effective observations -- then these relative errors drop to $19\%$ and $6\%$ (resp.). 
The invariant distribution estimated by ATLAS is very close to the one by original simulator $\mathcal{S}$ (see \cref{f:halfmoon}), with the estimated $L^1$-norm of the difference of their densities is $0.098\pm0.006$. The main reason for the small translational bias in the estimate of the stationary distribution is that the fast modes do not fully relax at the time scale $\tau$, and increasing $\tau$ is not an option in this case due to the high curvature of the \IMmth. As reported in \cref{t:halfmoon}, the estimated MRTs in the metastable states are quite accurate, and so are the transition rates. The training time for ATLAS is about $17$hrs; runtime for estimating the large-time quantities above is less than half that of original simulator $\mathcal{S}$. 

\begin{figure}[t]
\centering
\includegraphics[width = 0.6\textwidth]{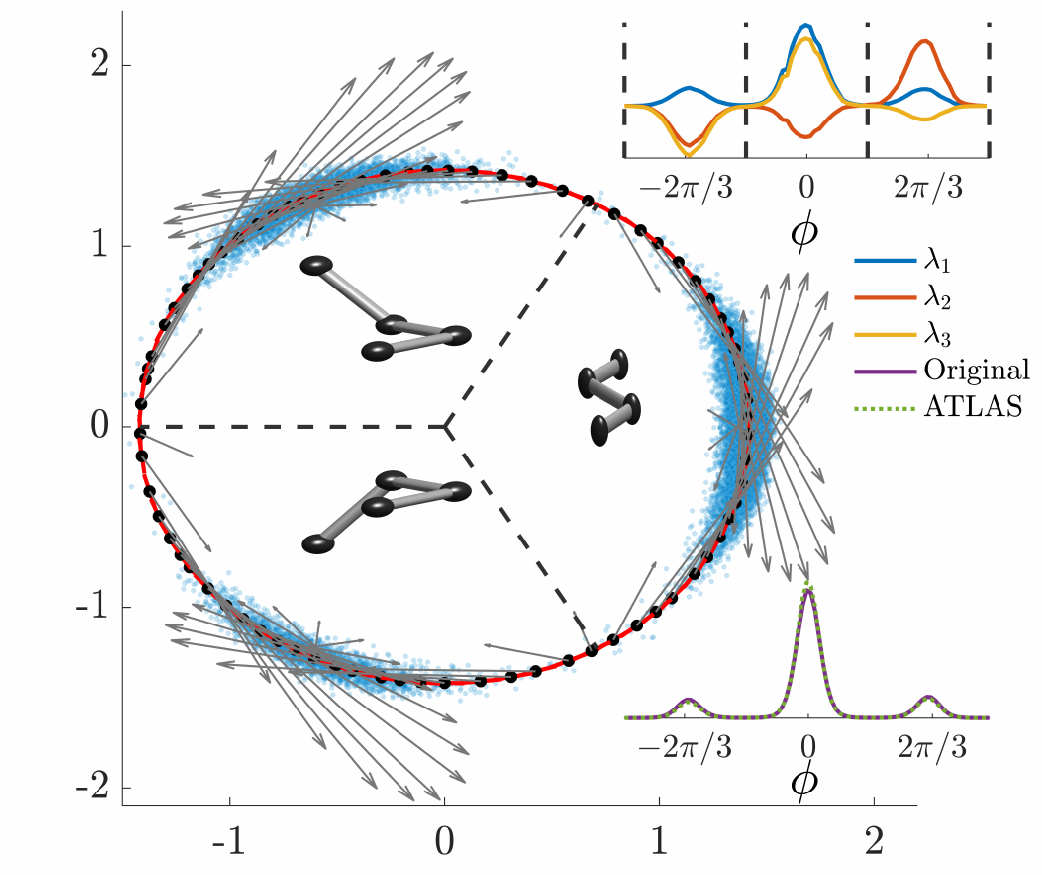}
\vskip-0.35cm
\caption{Butane. The points in the sample trajectory with time length 20 simulated by the original simulator are scattered in blue dots in $(x_4, z_4)$. 
Three data clusters corresponds to $M_{\text{top-cis}}$ (upper left), $M_{\text{bot-cis}}$ (lower left) and $M_{\text{trans}}$ (right) metastable states. The landmarks (black dots) and their neighborhood (red lines) and effective drift direction (gray arrows) in $(x_4, z_4)$ are plotted. In the upper right, the top three eigenfunctions of the transition matrices are plotted in the coordinate of $\phi$ with $\lambda_1=1, \lambda_2=0.9999$ and $\lambda_3=0.9999$. In the bottom right, the kernel-fitted invariant distribution from the trajectory of time length 500 generated by the original simulator and ATLAS simulator are plotted in the coordinate of $\phi$.}
\label{f:butanemanifold}
\end{figure}

\begin{figure}
  \centering
     \captionof{table}{Mean residence time (MRT) and reaction rates for butane}
\begingroup
\setlength{\tabcolsep}{2pt} 
\renewcommand{\arraystretch}{1.3}
\small{
\begin{tabular}[b]{c|c|c|c|}
\cline{1-4}
 \multicolumn{1}{|c|}{ \diagbox[dir=NW]{IC}{Sim.}} &  \begin{tabular}[c]{@{}c@{}}Original\\ $\mathcal{S}$\end{tabular}  & ATLAS & \begin{tabular}[c]{@{}c@{}}Ave. rel. \\ error\end{tabular} \\ \hline
\multicolumn{1}{|c|}{$M_{\text{trans}}$} &\begin{tabular}[c]{@{}c@{}}$0.412$\\ $\pm0.005$   \end{tabular}  &\begin{tabular}[c]{@{}c@{}}$0.414$\\ $\pm0.005$   \end{tabular}   & \begin{tabular}[c]{@{}c@{}}$0.02$\\ $\pm0.01$   \end{tabular}   \\ \hline
\multicolumn{1}{|c|}{$M_{\text{top-cis}}$} & \begin{tabular}[c]{@{}c@{}}$0.168$\\ $\pm0.002$   \end{tabular}    &\begin{tabular}[c]{@{}c@{}}$0.166$\\ $\pm0.002$   \end{tabular}    & \begin{tabular}[c]{@{}c@{}}$-0.03$\\ $\pm0.02$   \end{tabular} \\ \hline
\multicolumn{1}{|c|}{$M_{\text{bot-cis}}$}&  \begin{tabular}[c]{@{}c@{}}$0.168$\\ $\pm0.002$   \end{tabular}  & \begin{tabular}[c]{@{}c@{}}$0.158$\\ $\pm0.002$   \end{tabular}    & \begin{tabular}[c]{@{}c@{}}$-0.03$\\ $\pm 0.02$   \end{tabular}   \\ \hline
\multicolumn{1}{|c|}{Runtime} & 5.61h &3.84h  \\ 
\cline{1-3}
\end{tabular}}
    \centering
    \includegraphics[width =3.4cm]{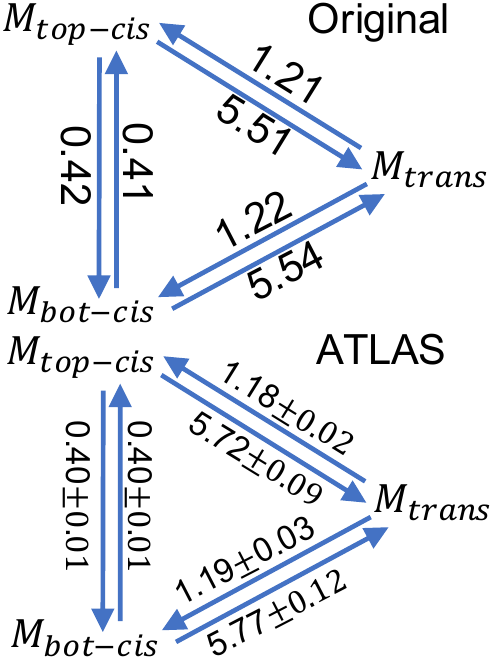}
    \qquad
\endgroup
\vskip0.1cm
    \caption*{\textmd{\footnotesize{{\bf Left:} the metastable states are given in $\phi$ variable: $M_{\text{trans}}=(-\frac{\pi}{3}, \frac{\pi}{3}), M_{\text{top-cis}}= (\frac{\pi}{3},\pi)$ and cis $M_{\text{bot-cis}}=(-\pi, -\frac{\pi}{3})$. {\bf Right:} mean transition rate constants between metastable states, with corresponding confidence intervals (estimated over multiple ATLASes). }}}
    \label{t:butane}
    \vskip-0.7cm
\end{figure}

\subsection{Butane model}
This is a model for the butane molecule, inspired by molecular dynamics \cite{Legoll2, TraPPE}, in the form of overdamped Langevin equations in $\mathbb{R}^6$ (see \cref{appx:butane}).
The dihedral angle $\phi$, which determines the distance of two outer carbons groups, is usually considered to be the slow variable. 
TICA however flags two coordinates, $x_4$ and $z_4$, as important coordinates; in the plane that they span three metastable states $M_{\text{trans}}$, $M_{\text{bot-cis}}$ and $M_{\text{top-cis}}$, concentrated around a circular \IMmth, are apparent (see fig.\ref{f:butanemanifold}).
ATLAS identifies that the slow variable is one-dimensional, accurately estimates the tangent line direction and \IMmth. 
The relative error of the estimated drift in the $(x_4, z_4)$ plane are on average $9\%$, vs. $20\%$ in all $6$ dimensions reported in \cref{t:error}. 
The $5$ fast variables are almost orthogonal to the slow variable (as suggested in \cite{Legoll2}): we therefore expect the local orthogonal projections to work as well as the oblique ones.
The top three eigenfunctions of an MSM estimated by ATLAS simulator identify these three metastable regions on the slow manifold, see \cref{f:butanemanifold} and \cref{appx:butane}. The invariant distribution of the ATLAS process has density very close, on \hatIMmth, to the one generated by the original simulator, with the estimated $L^1$-norm of the difference of the density $0.060\pm0.013$. 
The results reported in Table \ref{t:butane} show that the mean residence times in the three metastable states, estimated with ATLAS, are within 4\% relative error, with a runtime is about $68\%$ of the original simulator. All estimated reaction rate constants are within $5\%$ relative error. The training time of ATLAS is about 13hrs.

  \begin{table}[b!]
  \centering
\caption{Summary of error analysis.}
\begingroup
\setlength{\tabcolsep}{2pt} 
\renewcommand{\arraystretch}{1.3}
\begin{tabular}{c|c|c|c|c|}
\cline{1-5}
\multicolumn{1}{|c|}{ \diagbox[dir=NW]{Models}{Errors}}         & \multicolumn{1}{c|}{ $\abserrIMmth$} & \multicolumn{1}{c|}{ $\abserrTangle$} & \multicolumn{1}{c|}{ $\relerrb$} & \multicolumn{1}{c|}{ $\relerrLambda$} \\ \hline
\multicolumn{1}{|c|}{\multirow{2}{*}{\begin{tabular}[c]{@{}c@{}}Pinched\\ sphere\end{tabular}}} & $7(3)\times 10^{-3}$ & $0.02(0.03)$  & $0.11(0.10)$  & $0.04(0.03)$ \\ \cline{2-5}
\multicolumn{1}{|c|}{} &\cellcolor{Gray} $3(1)\times 10^{-2}$  &\cellcolor{Gray} $0.03(0.01)$  &  \cellcolor{Gray} $0.10(0.08)$ &\cellcolor{Gray} $0.05(0.03)$  \\ \hline
\multicolumn{1}{|c|}{\multirow{2}{*}{\begin{tabular}[c]{@{}c@{}}Oscillating\\ half-moons\end{tabular}}} & $ 9.63(0.04)\times 10^{-3}$ & $0.10(0.06)$  &  $0.33(0.23)$  & $0.12(0.07)$ \\ \cline{2-3}
\multicolumn{1}{|c|}{} & \cellcolor{Gray} $5(2)\times 10^{-3}$  & \cellcolor{Gray} $0.09(0.05)$  &  \cellcolor{Gray} $0.32(0.18)$&\cellcolor{Gray} $0.11(0.05)$ \\ \hline
\multicolumn{1}{|c|}{\multirow{2}{*}{Butane}} & $6(2)\times 10^{-3}$  & $0.014(0.007)$  & $0.17(0.12)$  & $0.06(0.02)$ \\ \cline{2-5}
\multicolumn{1}{|c|}{} &\cellcolor{Gray} $5(2)\times 10^{-3}$ &\cellcolor{Gray} $0.011(0.007)$  &\cellcolor{Gray} $0.21(0.13)$  &\cellcolor{Gray} $0.08(0.01)$  \\ \hline
\end{tabular}
\endgroup
\caption*{\textmd{\footnotesize{In each example, we report the mean and standard derivation (in parentheses) of error terms for the quantities evaluated at landmarks (white rows), and at points along ATLAS trajectories (gray rows); see \cref{appx:error}. }}}
\label{t:error}
\vskip-0.5cm
\end{table}

\section{Conclusion}
\label{s:comparisonExistingTechniques}
We have introduced a nonlinear nonparametric technique for reduction of fast-slow stochastic systems, that given a timescale $\tau$ and access to short trajectories from a black-box simulator, estimates an invariant manifold and an effective stochastic process, called ATLAS, on it, that averages the original system below the timescale $\tau$. The simulator for ATLAS has time-step of order $\tau$, typically much larger than the time-step of the original simulator $\delta t$ (which depends on the fastest timescale), and is intrinsically low-dimensional, making it possible to compute efficiently many long paths of the effective dynamics, and compute approximations to important quantities, such as stationary distributions, mean residence times, and transition rates. We have shown that, under suitable conditions, the estimation of ATLAS is not cursed by the dimension of the state space, and that ATLAS is robust to certain model errors.

This technique significantly extends the one introduced in \cite{CM:ATLAS} by correctly handling (i) large fast modes, instead of only very small fast oscillations around a slow manifold, which could be estimated by local PCA, (ii) fast modes that are not orthogonal to the slow manifolds, (iii) smoothly interpolating all estimated geometric and dynamics quantities, increasing the accuracy of the estimation. Last but not least, it is designed to efficiently run in exploration mode, without loss of accuracy.

The literature on model reduction, averaging and homogenization is vast, see e.g. \cite{pavliotis2008multiscale,Hartmann2020,MSM-review, Bruna2014,Givon-review}. Unlike existing techniques, here we do not require: previous knowledge of reaction coordinates or of the slow variables, which we estimate directly; linearity of the slow variables (as in PCA/PODs \cite{holmes_lumley_berkooz_rowley_2012}); that the fast modes are small (as in local PCA/PODs \cite{holmes_lumley_berkooz_rowley_2012} or DMD \cite{rowley_mezic_bagheri_schlatter_henningson_2009, DMD-textbook} or TICA \cite{prl1994, frankJCP2013}), nor that they are orthogonal to the slow manifold, nor that they can be globally defined (as in manifold learning techniques such as \cite{CKLMN:DiffusionMapsReductionCoordinates,RZMC:ReactionCoordinatesLocalScaling,Singer2009} and many others), which either requires the absence of even simple topological obstructions (loops) or require a possibly arbitrarily large number of additional coordinates. We also do not require to sample long trajectories, and in exploration mode we do not require a set of sufficiently well-behaved initial conditions; unlike exploration techniques such as \cite{ChiavazzoE5494} (and references therein). These techniques can fail (and they do in our examples) to correctly parametrize the invariant manifold, or (not exclusive) the effective dynamics, or would be cursed by the dimension of the state space.
 Our ATLAS algorithm estimates consistently and accurately the effective dynamics and its invariant manifold in an exploration scheme, which by itself is useful in many cases. Our reduction onto the estimated invariant manifold is nonlinear, and the estimation of both the invariant manifold and of the It\^o diffusion are locally parametric, in order to reduce the local sample size required for a given accuracy, but globally nonparametric.

The setting of our work, where a latent slow-fast system in a natural linear coordinate system is observed through a nonlinear observation map, is inspired by the works \cite{Singer2009,Dsilva2016}. These works start with a latent model significantly simpler than that in \cref{e:master}, and their objective is to learn the map back to the latent space, or at least to the slow variables in the latent space, from bursts of trajectories in observed space. That problem is tackled under significantly stronger assumptions on the latent system, and the approach is typically cursed by the ambient dimension, mainly because it seeks the reduction to slow variables {\it{after}} having constructed an approximation to the full system in the state space. In our work we first locally estimate a reduced system, and do so parsimoniously, by using a rather minimal set of parametric tools, and avoiding the curse of dimensionality.

Extensions to higher order equations, such as Langevin equations, more general local models {and nonlinear open neighborhood}, incorporating symmetries and conserved quantities, {considering non-Gausian noise} and combination with rare sampling techniques are currently being explored.

\bigskip

\noindent \textbf{Data Availability }

Data deposition: The software package implementing the proposed algorithms can be found on \url{https://github.com/yexf308/ATLAS}.

\bigskip
\noindent \textbf{Acknowledgments }
We thank Y. Kevrekidis and F. Lu for helpful discussions related to this work. MM is grateful for partial support from DOE-255223, FA9550-20-1-0288, NSF-1837991, NSF-1913243, and the Simons Fellowship. FY is grateful for partial support from AMS-Simons travel grant, Travel Support for Mathematicians
from Simons Foundation. Prisma Analytics, Inc. provided computing equipment and support.

\newpage
\let\normalsize\small
\appendix
\small

\section{Assumption, linear approximation, and averaging}
\label{s:AssumptionsAverageMLE}

We briefly review here the definitions of slow and invariant manifolds, some of the very basic expansions in geometric singular perturbation theory that motivate our key linearized model in \cref{e:meanAndCovBurst}, and the assumptions underlying them.

As a matter of notation, $E_{i\cdot}$ denotes the $i$-th row of a matrix $E$, and $E_{\cdot j}$ denotes the $j$-th column of a matrix $E$. 

\subsection{Assumption} \label{appx:assumption}
\label{s:assumption}
The following assumptions 1-4 ensure that for the original latent stochastic system \cref{e:master} there exists a uniformly asymptotically stable \IM{} \IMxmth \cite{NB-book, NB-jde, Kuehn-book}. 
\begin{myassum} \label{assump-diff}
Domain and differentiability: $f\in \cC^2(\cD,\bR^{D-d}), g\in \cC^2(\cD, \bR^d)$ and $F\in \cC^1(\cD, \bR^{(D-d)\times (D-d)}), G\in \cC^1(\cD, \bR^{d \times d })$, where  $\cD$ is an open subset of $\bR^d \times \bR^{D-d}$. We further assume that $f, g, F, G$ are bounded in sup-norm by a constant $M$ within $\cD$. 
\end{myassum}

\begin{myassum}\label{assump-slow}
Slow manifold: there is a connected open subset $\cD_0\subset \bR^d$ and a continuous function $\fvar^\star: \cD_0\rightarrow \bR^{D-d}$ such that 
\begin{equation*}\label{det-slow-manifold}
\SMxmth=\{(\svar,\fvar) \in \cD: \fvar=\fvar^\star (\svar), \svar\in  \cD_0\}
\end{equation*}
is a slow manifold of the deterministic system, that is, $(\svar,\fvar^\star(\svar))\in \cD$ and $f(\svar, \fvar^\star(\svar))=0$ for all $\svar\in{\cD}_0$.
\end{myassum}

\begin{myassum}\label{assump-stab}
Stability: the slow manifold is uniformly asymptotically stable, that is, all eigenvalues of the Jacobian matrix 
\begin{equation*}	
A^{\star}(\svar)=\partial_\fvar f( \svar,\fvar^\star(\svar))
\end{equation*}
have negative real parts, uniformly bounded away from 0 for all $\svar\in \cD_0$. 
\end{myassum}

\begin{myassum}\label{assump-nonde}
Non-degeneracy: the diffusivity matrix $F(\svar,\fvar )F(\svar, \fvar)^T$ is positive definite. 
\end{myassum}

Under these assumptions, Fenichel's theorem guarantees the existence of an invariant manifold (also called adiabatic manifold) \cite{NB-book,NB-jde,gst-review}
\begin{equation*}
\IMxmth=\{(\svar,\fvar)\in \cD: \fvar=\bar{\fvar}(\svar, \epsilon), \svar\in \cD_0\},
\end{equation*}
in a neighborhood of which trajectories concentrate for an extended time w.h.p. Also, \IMxmth is close to the \SM \SMxmth in the sense that $\bar{\fvar}(\svar, \epsilon)=\fvar^\star(\svar)+\cO(\epsilon)$.

The next assumption \ref{assump-diff-domin} imposes that the effect of the drift term is small relative to the effect of the diffusion term. 
\begin{myassum} \label{assump-diff-domin}
Diffusion-dominated dynamics:  for any $1\leq l\leq L$, $\sqrt{\sigma_d(\Lambda\lndmrkidx{l})}  \gg \|\mb\lndmrkidx{l}\|\sqrt{\tau}$.
\end{myassum}
This assumption allows us to simplify the construction of the diffusion-adapted, Mahalonobis-like metric. Indeed, the $\sqrt{\tau}$-neighbourhood of the landmark $\metricneighborhood(\mz^l, \sqrt{\tau})$ that we use is not exactly the same as the estimated $p\%$ confidence region of finding the effective reduced stochastic system at time $\sqrt{\tau}$, started at $\mz^l$: that would be better approximated by $$\tilde{\metricneighborhood}(\mz^l, \sqrt{\tau}):=\{\bv:\frac1{\cX^2_d(p)}(\bv-\mzlslow_0 -\mb\lndmrkidx{l}\tau)^T(\Lambda\lndmrkidx{l})^\dag(\bv-\mzlslow_0 -\mb\lndmrkidx{l}\tau)<\tau \}\,.$$
However, with the assumption of diffusion-dominated dynamics, the approximation we use is satisfactory. 
Indeed, the boundary of $\metricneighborhood(\mz^l, \sqrt{\tau})$ is a hyperellipsoid of dimension $d$ embedded in $\mR^D$. Columns of $\Uslowtruel$, which are the eigenvectors of $\Lambda\lndmrkidx{l}$, define the principle axes of the hyperellipsoid. $\sigma_{1}(\Lambda\lndmrkidx{l}), \sigma_{2}(\Lambda\lndmrkidx{l}), \cdots, \sigma_{d}(\Lambda\lndmrkidx{l})$ are proportional to squares of the lengths of the semi-axes. The vertices of the hyperellipsoid at time $t$ are $\mzlslow_0\pm \sqrt{\chi_d^2(p)\sigma_i(\Lambda\lndmrkidx{l}) t} \left(\Uslowtruel\right)_{\cdot i}$ for $i=1,2,\cdots, d$, so the minimum length of semi-axes at $t=\tau$ for the hyperellipsoid is $\smash{\sqrt{\chi_d^2(p)\sigma_d(\Lambda\lndmrkidx{l}) \tau}}$. In the mean time, the center of the hyperellipsoid is moved by $\|\mb\lndmrkidx{l}\|\tau$ if we use $\tilde{\metricneighborhood}(\mz^l, \sqrt{\tau})$ instead of $\metricneighborhood(\mz^l, \sqrt{\tau})$. Assumption \ref{assump-diff-domin} guarantees that the movement of the center is negligible relative to the length of the semi-axes of the hyperellipsoid.

\subsection{Linear approximation} \label{appx:linear}
In \cref{e:meanAndCovBurst} we assumed linear approximations of the time-dependent expectation $\trueexp$ and covariance $\truecov$. The slopes of these quantities, as a function of time, are $\smash{\trueexpslp}$ and $\smash{\truecovslp}$, and the intercepts of these quantities are $\smash{\mz_0\lndmrkidx{l}}$ and $\smash{\Gamma\lndmrkidx{l}}$. In this section, we provide some mathematical intuitions for this assumption, following the exposition of \cite{NB-book}, to which we refer the reader for further details. 
We start by considering the system in the latent space, and its linear approximation (\ref{e:master}) near the the invariant manifold. First, we define the deviation of sample paths from the invariant manifold: $\mzeta_t:=\fvar_t-\bar{\fvar}(\svar_t, \epsilon)$. An application of It\^o's formula implies that the fast dynamics part $\mzeta_t$ satisfies
\begin{equation} 
{\begin{aligned}  \label{xi-diff}
\rd \mzeta_t 
&=\frac{1}{\epsilon}\big( f(\svar_t,\bar{\fvar}(\svar_t, \epsilon)+\mzeta_t)- \epsilon  \partial_\svar\bar{\fvar}(\svar_t, \epsilon)  g(\svar_t,\bar{\fvar}(\svar_t, \epsilon)+\mzeta_t)-\eps c^{\Ito}(\svar_t, \mzeta_t)\big)\rd t \\ 
&\qquad+\frac{1}{\sqrt{\epsilon}} \begin{bmatrix}
 -\sqrt{\eps}\partial_\svar\bar{\fvar}(\svar_t, \epsilon)  G(\svar_t,\bar{\fvar}(\svar_t, \epsilon)+\mzeta_t) & F(\svar_t,\bar{\fvar}(\svar_t, \epsilon)+\mzeta_t)
\end{bmatrix}\rd W_t\,,
\end{aligned}}
\end{equation}
where  
$c^{\mathrm{\Ito}}(\svar,\mzeta):=\frac{1}{2}\sum_{j,k=1}^d \frac{\partial^2 \bar{\fvar}(\svar, \epsilon)}{\partial x_j \partial x_k} G_{j\cdot}(\svar,\bar{\fvar}(\svar, \epsilon)+\mzeta)G_{k\cdot}^T(\svar,\bar{\fvar}(\svar, \epsilon)+\mzeta)$, and 
$W_t = [U_t ; V_t]$, (Here ``$;$'' denotes concatenation of column vectors).
Always following \cite{NB-book}, {\it{if}} we ignore the \Ito-term and use the fact that the drift term vanishes when $\mzeta=0$, we derive that the invariant manifold $\bar{\fvar}(\svar_t, \epsilon)$ should satisfy the following PDE:
\beq \label{slow-manifold-eq}
f(\svar_t,\bar{\fvar}(\svar_t, \epsilon))= \epsilon  \partial_\svar\bar{\fvar}(\svar_t, \epsilon)  g(\svar_t,\bar{\fvar}(\svar_t, \epsilon))\,.
\eeq
Therefore, by Taylor expansion and \cref{slow-manifold-eq}, we have the following linear approximation $\tilde{\mzeta}_t$ of the fast dynamics $\mzeta_t$ in \cref{xi-diff}: 
\begin{align} \label{xi-OU}
\rd \tilde{\mzeta}_t= \frac{1}{\epsilon} A(\svar_t, \epsilon)\tilde{\mzeta}_t\rd t + \frac{1}{\sqrt{\epsilon}} F_0(\svar_t, \epsilon)\rd W_t\,,
 \end{align}
where 
\begin{align*}
 &A(\svar,\epsilon):= \partial_\fvar f(\svar,\bar{\fvar}(\svar, \epsilon) )-\epsilon  \partial_\svar\bar{\fvar}(\svar, \epsilon)  \partial_\fvar g(\svar,\bar{\fvar}(\svar, \epsilon))\quad ,\quad F_0(\svar,\epsilon):=\frac{1}{\sqrt{\epsilon}} \begin{bmatrix}
 -\sqrt{\eps}\partial_\svar\bar{\fvar}(\svar, \epsilon)  G(\svar,\bar{\fvar}(\svar, \epsilon)) & F(\svar,\bar{\fvar}(\svar, \epsilon))
\end{bmatrix}\,.
 \end{align*}

As for the slow dynamics $[\svar_t ; \bar{\fvar}(\svar_t, \epsilon)]$, we have 
\begin{align} \label{slow-dynamics}
\rd \begin{bmatrix}
\svar_t \\ 
 \bar{\fvar}(\svar_t, \epsilon)
\end{bmatrix} = g^{\text{slw}}(\svar_t)\rd t   +   G^{\text{slw}}(\svar_t)\rd U_t, 
\end{align}
where $g^{\text{slw}}(\svar):=B(\svar)g(\svar,\bar\fvar(\svar,\eps))+\tilde{c}^{\mathrm{\Ito}}(\svar)$, $G^{\text{slw}}(\svar):=
 B(\svar)G(\svar,\bar\fvar(\svar,\eps))$, and here $B(\svar):=\begin{bmatrix}I_{d\times d} \\ \partial_\svar \bar{\fvar}(\svar,\epsilon)\end{bmatrix} \in \bR^{D\times d}$, the $\Ito$ term
  $\tilde{c}^{\mathrm{\Ito}}(\svar)= \begin{bmatrix}{\displaystyle O}_{d\times (D-d)} \\ I_{(D-d)\times (D-d)}\end{bmatrix}c^{\mathrm{\Ito}}(\svar, 0) $. Here of course $\bar{\fvar}(\svar_t, \epsilon)$ is slaved to $\svar_t$, and we call the dynamics of $\svar_t$ only the slow reduced dynamics happening in $\mathbb{R}^d$.

The dynamics of $\tilde{\mzeta}_t$ shown in \cref{xi-OU} is a high dimensional Ornstein-Uhlenbeck process, and thus its expectation $\mathbb{E}\left[\tilde{\mzeta}^l_t\right|\mz_0^l]$ and covariance $\mathrm{cov}(\tilde{\mzeta}^l_t|\mz_0^l)$ at landmark $l$ stabilize exponentially fast to $\mathbf{0}$ and some matrix $\Theta(\svar_0^l)$. In particular, the error is of order $O(\eps)$ once $t$ reaches the timescale of separation  $\tau\gg\epsilon$, where here we have utilized assumption \ref{assump-stab}.
At the same time, at the timescale of separation $\tau$, the slow dynamics does not change significantly. In particular, in a neighborhood of a landmark $\mz^{l}$, the drift and diffusion coefficients $g^{\text{slw}}(\svar^l_t), G^{\text{slw}}(\svar^l_t)$ in \cref{slow-dynamics} do not change too much, and we may treat the slow dynamics as having nearly constant drift and diffusion coefficients. These equations and reasoning justify the form of reduced equations obtain by averaging, in the limit $\eps\rightarrow0$, of the form of \cref{e:reduced-par}.
 
 Now, notice that we have the following decomposition of the latent coordinate $\mw_t^l$ into coordinates for fast dynamics $\mzeta^l_t$ and coordinates for slow dynamics $[\svar^l_t\,;\bar{\fvar}(\svar^l_t, \epsilon)]$: 
\begin{align} \label{decomp-fast-slow}
\mw_t^l = \begin{bmatrix}
{\displaystyle O}_{d\times (D-d)}\\
I_{(D-d)\times (D-d)} 
\end{bmatrix} \mzeta^l_t + \begin{bmatrix}
\svar^l_t\\ 
\bar{\fvar}(\svar^l_t, \epsilon)
\end{bmatrix}.
\end{align}
Then, according to previous analysis regarding timescale of separation as well as \cref{decomp-fast-slow}, we have the following linear approximations of the latent coordinate at times $t$ comparable to $\tau$:
\begin{equation}  
\begin{aligned}\label{e:meanAndCovBurstlatent}
     \bE \left[\mw\lndmrkidx{l}_t \right|\mw\lndmrkidx{l}_0] &= \begin{bmatrix}
\svar^l_0\\ 
\bar{\fvar}(\svar^l_0, \epsilon)
\end{bmatrix} +g^{\text{slw}}(\svar^l_0)t +\cO(\eps)\,, \\ 
     \cov(\mw\lndmrkidx{l}_t|\mw\lndmrkidx{l}_0) &=  \begin{bmatrix}
     {\displaystyle O}_{d\times (D-d)}\\
I_{(D-d)\times (D-d)} 
\end{bmatrix} \Theta(\svar_0^l)\begin{bmatrix}
{\displaystyle O}_{d\times (D-d)}\\
I_{(D-d)\times (D-d)} 
\end{bmatrix}^T + G^{\text{slw}}(\svar^l_0)\left[G^{\text{slw}}(\svar^l_0)\right]^Tt + \cO(\eps) \,.
\end{aligned}
\end{equation}
These equations motivate \cref{e:meanAndCovBurst}, but in the latent space.

We now proceed to consider the situation in the observed variables, locally around a fixed point $\mz_t^l$; in particular we consider the behavior of the time-dependent expectation $\trueexp$ and covariance $\truecov$. Assume the local dynamics around landmark $\mz^l$ is within the chart $(\mathcal{U}_{\alpha(l)},\varphi_{\alpha(l)})$, and let $\varphi_{\alpha(l)}(\mz_t^l)=\mw_t^l=[\svar_t^l ; \fvar_t^l]$. 
We assume that the $1$-st order Taylor approximation of $\varphi^{-1}_{\alpha(l)}$ is accurate enough, which corresponds to the map $\varphi$ having sufficiently small Hessian, or restricting the size of the neighborhood under consideration to be small enough. We can then assume that for $t$ comparable to $\tau$, we have:
\begin{align*}
     \varphi^{-1}_{\alpha(l)}(\svar^l_t,\fvar^l_t)=\varphi^{-1}_{\alpha(l)}(\svar^l_0,\bar{\fvar}(\svar^l_0, \epsilon))+J^T(\varphi^{-1}_{\alpha(l)})\mid_{(\svar^l_0 , \bar{\fvar}(\svar^l_0, \epsilon))} \left(\begin{bmatrix}\svar^l_t \\  \fvar^l_t \end{bmatrix}-\begin{bmatrix}\svar^l_0 \\  \bar{\fvar}(\svar^l_0, \epsilon) \end{bmatrix}\right)+\cO(\eps)\,.
\end{align*}
Then, according to \cref{e:meanAndCovBurstlatent}, we have the linear approximations in the observation space of the form
\begin{align*}
 \mzlslow_0 &= \varphi^{-1}_{\alpha(l)}(\svar^l_0,\bar{\fvar}(\svar^l_0, \epsilon))\,,\\ 
 \mathbf{b}\lndmrkidx{l} & = J^T(\varphi^{-1}_{\alpha(l)})\mid_{(\svar^l_0 , \bar{\fvar}(\svar^l_0, \epsilon))}g^{\text{slw}}(\svar^l_0)\,, \\ 
 \Gamma\lndmrkidx{l} &= J^T(\varphi^{-1}_{\alpha(l)})\mid_{(\svar^l_0 , \bar{\fvar}(\svar^l_0, \epsilon))}\begin{bmatrix}
I_{(D-d)\times (D-d)} \\
{\displaystyle O}_{d\times (D-d)}
\end{bmatrix} \Theta(\svar_0^l)\begin{bmatrix}
I_{(D-d)\times (D-d)} \\
{\displaystyle O}_{d\times (D-d)}
\end{bmatrix}^T J(\varphi^{-1}_{\alpha(l)})\mid_{(\svar^l_0 , \bar{\fvar}(\svar^l_0, \epsilon))}\,, \\ 
\Lambda\lndmrkidx{l} & = J^T(\varphi^{-1}_{\alpha(l)})\mid_{(\svar^l_0 , \bar{\fvar}(\svar^l_0, \epsilon))}G^{\text{slw}}(\svar^l_0)\left[G^{\text{slw}}(\svar^l_0)\right]^T J(\varphi^{-1}_{\alpha(l)})\mid_{(\svar^l_0 , \bar{\fvar}(\svar^l_0, \epsilon))}\,,
\end{align*}
which justify the crucial approximation in \cref{e:meanAndCovBurst}, which motivates all our local estimators.
\subsection{Averaging}\label{appx:averaging}
In this section, we briefly review the idea of stochastic averaging, which is a classical method to analyze fast/slow systems, see for example \cite{freidlin2012random,pavliotis2008multiscale} for a comprehensive review of this subject. See in particular, theorem 2.1 in chapter 7 of \cite{freidlin2012random}, known as the averaging principle, \cite{Givon-review} for a survey of several approaches to the problem of extracting effective dynamics including averaging, where similarities and differences between these approaches are highlighted.

From the theory perspective, there exists huge body of literature on stochastic averaging, where strong and weak convergence results are provided under different regularity assumptions on slow-fast SDE coefficients, see e.g. \cite{Givon_2006,Khas_minskii_1966,Bakhtin_2004,Kifer_2005,Li_2008,Veretennikov_1991}. Besides these, many works also study the rate of convergence of the original process to the reduced one (e.g. as a function of $\eps$); see for example \cite{pavliotis2008multiscale,Givon_2007,Khasminskii_2004,cms/1288725269,Vanden_2003,Zhang_2018,rockner2019,Abourashchi_2010,Has_minskii_1966,Weinan_2005}. In particular, the recent work \cite{rockner2019} provides a very general, robust and unified method for establishing the averaging principle, involving both strong and weak convergence, for slow-fast SDEs with irregular coefficients and under the fully coupled case (i.e., the diffusion coefficient in the slow equation can depend on the fast term) for weak convergence (but not for strong convergence). This leads to simplifications and extensions of previously established results. It also shows that the strong and weak convergence rates depend only on the regularity of all the coefficients with respect to the slow variable.

In addition, a very recent work \cite{Hartmann2020} provides quantitative results on the connection of averaging to coarse-graining and effective dynamics in multiscale studies. It also presents a detailed comparison of the averaging and the conditional expectation approach in the case of (non-reversible) Ornstein-Uhlenbeck (O-U) processes and isolate sufficient conditions under which the two approaches agree. 

Now, we briefly review the formulation of averaging, and we start from considering in the latent space. From \cref{e:master}, we define the averaged approximation around the invariant manifold, as in \cref{e:reduced-par}, with the averaged coefficients given by \cite{rockner2019}:
\begin{align}  \label{s:average-coeff-latent}
\bar{g}(\svar) := \int_{\mathbb{R}^{D-d}} g(\svar, \fvar)\nu(\rd\fvar|\svar) \quad,\quad \bar{G}(\svar) := \sqrt{\int_{\mathbb{R}^{D-d}} G(\svar,\fvar)G^{T}(\svar,\fvar)\nu(\rd\fvar|\svar)}\,.  
\end{align}
Here the conditional invariant measure $\nu(\fvar|\svar)$, as mentioned in \cref{e:reduced-par}, is the unique invariant measure of the process $\mY_t^\svar$ \cite{rockner2019}, governed by the equations ``frozen'' at $\svar$:
\begin{equation}
\left\{
\begin{aligned}   \nonumber
&\rd {\mY_t^\svar}= f(\svar,\mY_t^\svar)\rd t+F(\svar,\mY_t^\svar)\rd V_t \\
&\mY_0^\svar=\fvar
\end{aligned}
\right. .
\end{equation}

Now we move to the observation space. In the local coordinates discussed in \cref{s:FastSlowSDEs}, split between the local tangent plane to the invariant manifold and the affine subspace containing the fast variables, we can decouple slow and fast variables. We then have the following local SDEs written in the variables $\sovar$ and $\fvari$, derived from \cref{e:master} by applying It\^o's formula:
\begin{equation*}
\left\{
\begin{aligned}   
&\rd \sovar_t = \tilde{g}(\sovar_t, \fvari_t)   \rd t +  \tilde{G}(\sovar_t, \fvari_t) \rd U_t \\
&\rd \fvari_t = \frac{1}{\eps} \tilde{f}(\sovar_t, \fvari_t) \rd t + \frac{1}{\sqrt{\eps}}\tilde{F}(\sovar_t, \fvari_t) \rd V_t
\end{aligned}
\right. .
\end{equation*}
From these SDEs in local coordinates, we can define the reduced SDEs, as in \cref{e:master-atlas}, with the averaged coefficients given by 
\begin{align}  \label{s:average-coeff-obs}
b(\sovar) = \int_{\mathbb{R}^{D}} \tilde{g}(\sovar, \fvari)\nu(\rd\fvari|\sovar) \,\,,\,\, H(\sovar) = \sqrt{\int_{\mathbb{R}^{D}} \tilde{G}(\sovar,\fvari)\tilde{G}^{T}(\sovar, \fvari)\nu(\rd\fvari|\sovar)}.  
\end{align}
Here again, the conditioned invariant measure $\nu(\fvari|\sovar)$, as mentioned in \cref{e:master-atlas}, is the unique invariant measure of the process $\fvari_t^{\sovar}$, governed by the equations ``frozen'' at $\sovar$:
\begin{equation}
\left\{
\begin{aligned}   \nonumber
&\rd {\fvari_t^{\sovar}}= \tilde{f}(\sovar,\fvari_t^{\sovar})\rd t+\tilde{F}(\sovar,\fvari_t^{\sovar})\rd V_t \\
&\fvari_0^{\sovar}=\fvari
\end{aligned}
\right. .
\end{equation}
As mentioned in \cref{s:FastSlowSDEs}, these reduced SDEs as in \cref{e:master-atlas} may be viewed in intrinsic coordinates, or in Cartesian coordinates in the ambient space $\mathbb{R}^D$, with $\sovar_t\in\mathbb{R}^D$ but on \IMmth, $b\in\mathbb{R}^D$ a vector field on \IMmth, and $H\in\mathbb{R}^{D\times d}$ acting on a Wiener process $U_t$ in $\mathbb{R}^d$.

\section{Algorithms}
\label{s:SI:algos}
We provide here detailed pseudo-code for the construction of ATLAS, see \cref{alg:general,alg:EstLocal,alg:EstGeometric,alg:ConSke,alg:ConATLAS,alg:ConATLASExplore}.
The algorithm follows closely the theory, with minor caveats, having to do with constants that are assumed known, while in practice they would either need to be estimated, or set by the user using external information. 

We also discuss several minor modifications to the algorithms. 

\begin{enumerate}

\item In \cref{alg:ConATLAS},  the neighbor landmarks of the current point $\mz_t$, $\mathcal{N}_\tau^{\mathcal{A}}(\mz_t)$ are approximated by the nearest landmark and its neighbors, $\{k_t, \mathcal{N}(k_t)\}$. Then we need an efficient method to search the nearest landmark $k_{t+1}$ for the next point $\mz_{t+1}$ in the ``metric'' $\hattildemetric$. we update the nearest landmark by only calculating the distance of $\mz_{t+1}$ to the current nearest landmark and its neighbors, and repeat this procedure until the nearest landmark remains the same. This procedure avoids the global search which could be very expensive. Then when simulating ATLAS process, there is no need to check whether $\|\mz-\lndmrk{l}\|\le  \hat{R}_{\max}$ when calculating $\hattildemetric(\mz_t, \lndmrk{l})$, since the current point $\mz_t$ is always close enough to the neighbor landmarks.

 \item In \cref{alg:ConSke} and \cref{alg:ConATLASExplore} we say that when $\hatmetric(\lndmrk{l}, \lndmrk{k})<d_{\text{con}} $, we add $k$ to $\mathcal{N}(l)$ and add $l$ to $\mathcal{N}(k)$: in practice, we relax this condition to $\min(\hattildemetric(\lndmrk{l}, \lndmrk{k}), \hattildemetric(\lndmrk{k}, \lndmrk{l}))<d_{\text{con}} $, in order to include more landmarks. This slight modification is particularly useful when the condition that the process is diffusion-dominated (see assumption \ref{assump-diff-domin} in \cref{s:AssumptionsAverageMLE}) does not hold. 
 
\item In  \cref{alg:EstLocal} and \cref{alg:ConATLAS}, there is no need to explicitly calculate $\LambdaA(\mz)$ since we can use the iterative algorithm to output top $d$ singular values and corresponding singular vectors to estimate the diffusion coefficient $\HA_d(\mz)$ with the cost of $O(C^dDd^2)$ where $C^d$ is the number of landmarks in $\mathcal{N}_\tau^{\mathcal{A}}(\mz)$.  {In the 
$d=1$ scenarios (i.e., Oscillating half-moons and Butane), the average number of landmarks in $\mathcal{N}_\tau^{\mathcal{A}}(\mz)$ is approximated 4 and for the $d=2$ example (Pinched Spheres), this average increases to about 8.} {However, if $d$ is not relatively small,  the number of landmarks in $\mathcal{N}_\tau^{\mathcal{A}}(\mz)$ may grow exponentially as $d$, so it might be challenging to simulate trajectories with ATLAS simulator.}
When the ambient dimension $D$ is very large, one could use randomized SVD in eq.(\ref{e:LambdadN}, \ref{e:diffDATLAS}, \ref{e:diffATLAS}) to further significantly lower the computational complexity in projection to rank $d$ \cite{rsvd-pnas}. In all three models, we didn't use this approach since the biggest $D$ that we tested is 20 and the most time consuming part is not this projection step. 

\item When refinement of the landmark is necessary in \cref{alg:EstGeometric}, we perform this only in the last round of refinement. The reason that we don't use this correction at each round of refinement is it will consistently move the landmark position towards the direction of the effective drift and the region like around the saddle point will be not covered by the neighborhoods of landmarks. 

\end{enumerate}

\begin{algorithm}
\small{
\begin{algorithmic}[1]
\Require $\mathcal{S}(\mz_0, N, t_0)$: original simulator generates $N$ trajectories starting from $\mz_0$ of time length $t_0$; $\mu_0$: probability measure for initial conditions; $L$: number of initial conditions; $[\tau_{\min}, \tau_{\max}]$: interval for regression; $\tau$: timescale for reduction; $d$: intrinsic dimension; $d_{\text{con}}$: connection threshold.
\Ensure ATLAS $\mathcal{A}$: structure containing data to evaluate $\ProjA,\bA,\LambdaA,\HA_d$ at any $\mz$, and simulate $\mzAt$ on $\hatIMmth$.
\State Sample $L$  initial conditions $\{\mz\lndmrkidx{l}_0\}_{l=1,\dots, L}\sim_{\text{i.i.d.}}\mu_0$  
\For{$l=1,\dots, L$}
\State $\mathcal{B}^l \leftarrow\mathcal{S}(\mz\lndmrkidx{l}_0,N,\tau_{\max})$.
 \State $[\{\blN,\LambdaldN,\HldN, \barmmlNM, \bar t_M,\Vfastl, \Uslowl\}_l] \leftarrow \text{LearningDynamics}(\mathcal{B}^l , \tau_{\min}, \tau_{\max}, d)$: estimation of local parameters of effective dynamics.
 \State $[\{\lndmrklN,\tanPlaneldN,\PNlz\}_l] \leftarrow \text{LearningGeometry}( \blN, \barmmlNM, \bar t_M, \Vfastl, \Uslowl)$: estimation of local parameters of \IMmth.
\EndFor
\State $\{\lndmrk{l}\}_{l=1}^{L'} \leftarrow \text{ConstructIM}(\{\lndmrk{l}\}_{l=1}^{L}, \tau, d_{\text{con}} )$: construct sketch of the \IM\ \IMmth.
\State $\ProjA,\bA,\LambdaA,\HA_d \leftarrow$ assemble interpolated quantities using the estimated quantities above and eq.(\ref{e:zATLAS}--\ref{e:diffATLAS}).
\end{algorithmic}
}
\caption{Pseudo-code for ATLAS construction}
\label{alg:general}
\end{algorithm}

\begin{algorithm}
\small{
$[\{\blN,\LambdaldN,\HldN, \barmmlNM, \bar t_M,\Vfastl, \Uslowl\}_l] \leftarrow \text{LearningDynamics}(\mathcal{B}^l , \tau_{\min}, \tau_{\max}, d)$
\begin{algorithmic}[1]
\Require $\mathcal{B}^l$, the local burst;  $[\tau_{\min}, \tau_{\max}]$, interval for regression; $d$, intrinsic dimension.
\Ensure  
          Estimated drift, diffusivity matrix and diffusion coefficient $\blN,\LambdaldN, \HldN$ of landmarks.
\State $t_1, \dots, t_M$:  $M$ equispace points in  $[\tau_{\min},\tau_{\max}]$.
\State $\mmlNt{t} \leftarrow \frac{1}{N}\sum_{n=1}^N \mz\lndmrkidx{ l,n}_t $, $\ClNt\leftarrow \frac{1}{N-1}\sum_{n=1}^N (\mz\lndmrkidx{ l,n}_t-\mmlNt{t})\otimes (\mz\lndmrkidx{ l,n}_t-\mmlNt{t})$: empirical means and covariances, eq.(\ref{e:mean-cov-est}).
\State  $\barmmlNM \leftarrow \frac{1}{M}\sum_{m=1}^M\hat{\mm}\lndmrkidx{l,N}_{t_m}$, $\ClNbar_M \leftarrow \frac{1}{M}\sum_{i=1}^M\ClN_{t_m}$, $\overline t_M\leftarrow\frac1M\sum_{m=1}^M t_m$.
\State $\blN \leftarrow {\sum_{m=1}^M(\hat{\mm}\lndmrkidx{l,N}_{t_m} -\barmmlNM)(t_m-\bar t_M)}/{\sum_{m=1}^M(t_m-\bar t_M)^2} $: estimated drift term, \cref{e:drift-est}.
\State $\LambdalN \leftarrow {\sum_{m=1}^M(\ClN_{t_m} -\ClNbar_M)(t_m-\bar t_M)}/{\sum_{m=1}^M(t_m-\bar t_M)^2}$
\State $  \LambdaldN \leftarrow \Projrkd(\LambdalN )=\Uslowl\hat\Sigma_d(\Uslowl)^T$: estimated diffusivity matrix, eq.(\ref{e:Lambda_Nl}).
\State  $\HldN \leftarrow (\LambdaldN)^\frac12= \Uslowl\sqrt{\hat\Sigma_d}$: estimated diffusion coefficient, eq.(\ref{e:LambdadN})
\State $ \hat\Gamma\lndmrkidx{l}_{D-d} \leftarrow \text{Proj}_{\text{rk}(D-d)} ( \ClNbar_M -  \LambdalN \bar t_M )=\Vfastl \tilde\Sigma_{D-d} (\Vfastl)^T $:  estimated covariance matrix of the fast modes, eq.(\ref{e:cov-intercept}). 
\end{algorithmic}
}
\caption{Estimation of local parameters of effective dynamics \hfill  [sec.\ref{s:estLocalParams}] }
\label{alg:EstLocal}
\end{algorithm}

\begin{algorithm}
\small{
$[\{\lndmrklN,\tanPlaneldN,\PNlz\}_l] \leftarrow \text{LearningGeometry}( \blN, \barmmlNM, \bar t_M, \Vfastl, \Uslowl)$
\begin{algorithmic}[1]
\Require  Estimated quantities in \cref{alg:EstLocal}; 
\Ensure  $\lndmrklN$: landmarks; $\tanPlaneldN$: estimated tangent planes; $\PNlz$: oblique projection along the fast mode.
\State $ \lndmrklN \leftarrow \barmmlNM-\blN \bar t_M$: estimated landmarks,  eq.(\ref{e:lndmrk}).
\State  $\PNlz \leftarrow \Uslowl(\Uslowl)^T(EE^T)^\dag(\mz-\lndmrk{l})+\lndmrk{l}$, where $E=[\Uslowl,\Vfastl]$: estimated oblique affine projections along the fast modes, \cref{e:projIM}.
\State  $\tanPlaneldN \leftarrow \mathrm{span(cols(}\Uslowl))$: estimated local tangent plane, eq.(\ref{e:tanPlanldN}).
\end{algorithmic}
}
\caption{Estimation of local geometric parameters of \IM{} \IMmth \hfill  [sec.\ref{s:IMestimation}] }
\label{alg:EstGeometric}
\end{algorithm}

\begin{algorithm}
\small{
\begin{algorithmic}[1]
\Require landmarks $\{\lndmrk{l}\}_{l=1}^L$ and their estimated quantities; $\tau$, timescale for reduction; $d_{\text{con}}$: connection threshold.
\Ensure updated landmarks $\{\lndmrk{l}\}_{l=1}^{L'}$ and their neighbors $\mathcal{N}(l)$.
\For{$l=1, \dots, L$}
\ \ \ \For{$k=l+1,\dots, L$}
  \If{$\hatmetric(\lndmrk{l}, \lndmrk{k})<\left(1-\sqrt{\frac12}\right)\sqrt{\tau}$}
  \State Remove $\lndmrk{k}$.
  \ElsIf{$ \hatmetric(\lndmrk{l}, \lndmrk{k})<d_{\text{con}} $}
  \State Add $k$ to $\mathcal{N}(l)$, add $l$ to $\mathcal{N}(k)$.
  \EndIf
\ \ \ \EndFor
\EndFor
\end{algorithmic}
}
\caption{Construct sketch of the \IM\ \IMmth, $\{\lndmrk{l}\}_{l=1}^{L'} \leftarrow \text{ConstructIM}(\{\lndmrk{l}\}_{l=1}^{L}, \tau, d_{\text{con}})$ \hfill [sec.\ref{s:IMestimation}]}
\label{alg:ConSke}
\end{algorithm}

\begin{algorithm}
\small{
\begin{algorithmic}[1]
\Require landmarks $\{\lndmrk{l}\}_{l=1}^{L'}$ and their estimated quantities; $\lambda$: time-step in ATLAS simulator in the unit of $\tau$; $\mz_t$: current point; $k_{t}$: the nearest landmark of current point
\Ensure $\mz_{t+1}$, next point; $k_{t+1}$, the nearest landmark of current point in the ``metric'' $\hattildemetric$.
\State $\omega^l(\mz_t)\leftarrow\exp(-\hattildemetric(\mz_t, \lndmrk{l})/\sqrt{\tau}) \text{ for }l\in\{k_{t}, \mathcal{N}(k_{t})\} , Z(\mz_t) \leftarrow\sum_{l\in\{k_t, \mathcal{N}(k_t)\}} \omega^l(\mz_t) $. 
\State $\mz_t^{\mathcal{A}} \leftarrow \frac{1}{Z(\mz_t)}\sum\nolimits_{l\in\{k_{t}, \mathcal{N}(k_{t})\}} \Plz w\lndmrkidx{l}(\mz_t) $, \  $\bA(\mz_t) \leftarrow\frac{1}{Z(\mz_t)}\sum\nolimits_{l\in\{k_t, \mathcal{N}(k_t)\}} \bl w\lndmrkidx{l}(\mz_t) $ in eq.(\ref{e:zATLAS}, \ref{e:meanAndCovBurstATLAS}). 
\State $\LambdaA(\mz_t) \leftarrow \frac{1}{Z(\mz_t)}\sum\nolimits_{l\in\{k_t, \mathcal{N}(k_t)\}} \Lambdal  w\lndmrkidx{l}(\mz_t), \HA_d(\mz_t)\leftarrow(\Projrkd \LambdaA(\mz_t))^\frac12 $ in eq.(\ref{e:diffDATLAS}, \ref{e:diffATLAS}).
\State $\mz_{t+1}\leftarrow \mzA + \bA(\mz_t) \lambda\tau+ \HA_d(\mz_t) \Delta W_{\lambda\tau}$, where $\Delta W_{\lambda\tau}\sim \mathcal{N}(0, \lambda\tau I_d)$.
\State Update the nearest landmark $k_{t+1}$ for the next point $\mz_{t+1}$.
\end{algorithmic}
}
\caption{Simulate one time-step with ATLAS simulator,   $[\mz_{t+1}, k_{t+1}] \leftarrow \text{ATLAStime-step}(\mz_t, k_t, \{\lndmrk{l}\}_{l=1}^{L'}, \lambda )$   \hfill [sec.\ref{s:ATLASsimul}]}
\label{alg:ConATLAS}
\end{algorithm}

\begin{algorithm}
\small{
\begin{algorithmic}[1]
\Require  $\mathcal{S}(\mz_0, N, t_0)$: original simulator generates $N$ trajectories starting from $\mz_0$ of time length $t_0$; $\mu_0$, probability measure for initial conditions; $T_{\max}$: the maximum time-steps; $L_{\text{int}}$: number of initial conditions; $[\tau_{\min}, \tau_{\max}]$: interval for regression; $\tau$: timescale for reduction; $d$: intrinsic dimension; $d_{\text{con}}$: connection threshold;  $d_{\text{thr}}$: exploration threshold;  $\lambda$: time-step in ATLAS simulator in the unit of $\tau$; $\mz$:  current point; $k$: the nearest landmark of the current point.
\Ensure $\mz$: next point; $k$: the nearest landmark of next point; updated landmarks $\{\lndmrk{l}\}$ and their estimated quantities; ATLAS $\mathcal{A}$. 
\State Construct ATLAS $\mathcal{A}$ with $L_{\text{int}}$ initial conditions by performing \cref{alg:general} and construct $L$ landmarks and their estimators.
\State Start the initial condition from the first landmark: $k_0\leftarrow 1$, $\mz_0\leftarrow \lndmrk{1} $.
\While{$t\le T_{\max}$}
\If{$\min_{j\in\{k_{t}, \mathcal{N}(k_{t})\}}\hattildemetric(\mz_t, \lndmrk{j})>d_{\text{thr}} $}
	\State  $l\leftarrow L+1$, $\mz_0^{l}\leftarrow \mz_t$,	$\mathcal{B}^l \leftarrow\mathcal{S}(\mz\lndmrkidx{l}_0,N,\tau_{\max})$,
 \State $[\{\blN,\LambdaldN,\HldN, \barmmlNM, \bar t_M,\Vfastl, \Uslowl\}_l] \leftarrow \text{LearningDynamics}(\mathcal{B}^l , \tau_{\min}, \tau_{\max}, d)$,
 \State $[\{\lndmrklN,\tanPlaneldN,\PNlz\}_l] \leftarrow \text{LearningGeometry}( \blN, \barmmlNM, \bar t_M, \Vfastl, \Uslowl)$.
 	\For{$k=1,\dots, L$}
             \If{$ \hatmetric(\lndmrk{l}, \lndmrk{k})<d_{\text{con}}$}
                  \State Add $k$ to $\mathcal{N}(l)$, add $l$ to $\mathcal{N}(k)$.
             \EndIf
      \EndFor
      \State Update ATLAS $\mathcal{A}$ and $L\leftarrow L+1$.
  \State     $\mz_t \leftarrow \lndmrk{l}$, $k_t\leftarrow l$.
\EndIf
\State Simulate one time-step with ATLAS simulator:  $[\mz_{t+1}, k_{t+1}] = \text{ATLAStime-step}(\mz_t, k_t, \{\lndmrk{l}\}_{l=1}^{L}, \lambda )$.
\State $t\leftarrow t+1$.
\EndWhile
\end{algorithmic}
}
\caption{Construct ATLAS simulator in exploration mode \hfill [sec.\ref{s:explorationmode}]}
\label{alg:ConATLASExplore}
\end{algorithm}

\begin{algorithm}
\small{
\begin{algorithmic}[1]
\Require  landmarks $\{\lndmrk{l}\}_{l=1}^{L}$ and their estimated quantities; $N_{\text{msm}}$: number of initial points to construct MSM, $\rd t_{\text{msm}}$: time of the short paths to construct MSM, $\tau$: timescale for reduction. 
\Ensure M: transition matrix for the Markov State Model constructed from ATLAS
 \For{$i=1,\dots, L$}
      \For{$j = 1, \dots, N_{\text{msm}}$}
      \State  $[\mz, k(j)] \leftarrow \text{ATLAStime-step}(\lndmrk{i}, i, \{\lndmrk{l}\}_{l=1}^{L}, \rd t_{\text{msm}}/\tau )$
      \EndFor
      \State $M_{il} \leftarrow \frac{\#(k==l)}{N_{\text{msm}}}$
 \EndFor 
\end{algorithmic}
}
\caption{Construct Markov state models from ATLAS \hfill [sec.\ref{s:ATLASuses}]}
\label{alg:ConMSM}
\end{algorithm}

\section{Examples}  \label{appx:example}
In this section we discuss in depth details and results of ATLAS in the numerical examples in \cref{s:numericalExp}. In \cref{t:parameter} we report the parameters of the models and for the construction of ATLAS. 

In our numerical tests on the accuracy of ATLAS for the approximation of invariant distribution, we proceed as follows. We will generate two sufficiently long trajectories with both the original and ATLAS simulator, and we take samples from each of the two trajectories.  Then, we apply the projection $\PNlop$ to the samples obtained from the original simulator. Here we only use the oblique projection which is consistent with the ATLAS simulation method, and does not require any knowledge of the latent space. For plotting purposes only, we visualize the smoothed histograms by binning the projected samples according to latent slow variables, or other specified coordinates, for both the original and ATLAS simulator. The $L^1$- and $L^2$-norm of the difference of their approximate probability densities are calculated directly from the histograms. The bin widths should be of order $\sqrt{\tau}$, consistently with the spirit of ATLAS simulator, which averages information below timescale $\tau$, with a constant coefficient depending on the scaling of the diffusion coefficient for the slow variable. If the latent slow variable is unknown, we could automatically cluster the samples by its nearest landmark. 

In our numerical tests on the accuracy of ATLAS for the local statistics, we proceed as follows. We will generate a long trajectory with the ATLAS simulator and compare the estimated invariant manifold, estimated tangent space, estimated effective drift and diffusion terms at each point with the analytical-derived reduced dynamics on the slow manifold.

In our numerical tests on the accuracy of ATLAS for the approximation of medium- and large-time observables, we proceed as follows. We will sample initial conditions with respect to the invariant measure restricted to the specified initial regions of state space, and obtain an estimate of the residence time in the target regions, with both simulators. To be specific, first, we generate a single sufficiently long trajectory with either the original or ATLAS simulator and uniformly sample $N_{\text{IC}}$ initial conditions that are restricted to the corresponding initial regions (e.g. specified metastable states). Second, we run in parallel the dynamics with both simulators giving to each the same initial conditions, sampled above.  We check whether a trajectory reaches the boundary of region at each ATLAS time-step for both simulators and record the residence time once they leave. This is to ensure the consistency of time-steps for both simulators in recording residence times.  At last, we compute the mean residence time and its confidence interval for both simulators. To ensure the robustness of the ATLAS algorithm, we repeat the construction of ATLAS (which is random with its input, the observed bursts) and sampling of residence time ten times, and calculate the confidence interval of the relative error of the mean resident time.

It is nontrivial to identify the metastable regions in fast-slow stochastic systems when $D$ is very large. With ATLAS one can easily build up Markov state models(MSMs) and construct the transition matrix for the effective reduced process. First, the number of landmarks naturally become associated to the states of MSMs; more precisely state $l$ of the MSMs is the set of points on the invariant manifold whose nearest landmark, in the $\hattildemetric$, is $\lndmrk{l}$. Starting from the node of the same $i$-th landmark  we use ATLAS simulator to simulate in parallel $N_{\text{msm}}$ short trajectories of time $\tau$, which is one ATLAS time step.  Let $N_{ij}$ be the number of trajectories whose end positions land in the same state of the $j$-th landmark. The probability transition from $i$-th landmark to $j$-th landmark is estimated as $M_{ij}:=N_{ij}/N$, see \cref{alg:ConMSM}. The eigenvalues and eigenvectors are approximations to the spectrum and eigenfunctions of the transfer operator of the reduced stochastic system, $\exp(\tau \mathcal{L})$, where $\mathcal{L}$ is the generator of the reduced effective process. Since we don't assume the reversibility, the spectrum and eigenvectors could be imaginary.  The top left eigenvector of the transition matrix is proportional to the invariant distribution of reduced stochastic process and is always real.  The spectral gap, which is the distance between the next dominant eigenvalue and the eigenvalue 1, indicates the decay rate of correlations and the reciprocal of the spectrum gap shows the order of ATLAS time-steps to reach the equilibrium. The number of the dominant eigenvalues are the number of metastable states. Performing some clustering or connectivity-detection algorithms on landmarks with high equilibrium density can yield metastable regions. At least in the setting of real eigenvectors, the positive or negative regions of successive dominant eigenvectors can be used to identify metastable regions.
 
\begin{table*}[]
\caption{Parameters of three examples}
\footnotesize
\centering
\begin{tabular}{c|c|c|c|c|c|c|c|}
\cline{2-8}
\multicolumn{1}{l}{} &  \multicolumn{1}{|c|}{Model parameters} & $\delta t$ & $\tau$ & $[\tau_{\min{}}, \tau_{\max{}}]$ &$N$  & Refine & $N_{\text{refine}}$\\ \hline
\multicolumn{1}{|c}{\begin{tabular}[c]{@{}c@{}}Pinched \\ sphere\end{tabular}} & \multicolumn{1}{|c|}{\begin{tabular}[c]{@{}c@{}}$ \epsilon=\num{5e-3},a_1=4, a_2=8$ \\ $c_1=2, c_2=\num{5e-1}, c_3=\num{5e-2}$\\ $c_4=\num{4e-1}, c_5=\num{5e-2}, c_6=\num{4e-1}$\end{tabular}} &$\num{5e-4}$  & $200\delta t$  & $[100\delta t, 200\delta t]$ &$\num{4e5}$ &No & NA\\ \hline
\multicolumn{1}{|c}{\begin{tabular}[c]{@{}c@{}}Oscillating\\  half-moons\end{tabular}} & \multicolumn{1}{|c|}{\begin{tabular}[c]{@{}c@{}}$ \eps=\num{1e-2}, a_1 =0, a_2 = \num{5e-3}$\\ $a_3=\num{2.5e-3}, a_4 =\num{6e-2}, b_1=\num{4e-2}$ \\$  b_2=\num{3.5e-2},b_3=\num{5e-2}, b_4=\num{2e-2}$\end{tabular}} & $\num{5e-2}$ &  $20\delta t$&$[20\delta t, 25\delta t]$ & $\num{8e6}$  & Yes &\num{1.6e7}\\ \hline
\multicolumn{1}{|c}{butane} &  \multicolumn{1}{|c|}{\begin{tabular}[c]{@{}c@{}}$ l=1.53, k_2=\num{3.19225e5}, k_3 = \num{6.25e4}$\\ $\theta_{\text{eq}}=1.9548, c_1 =\num{2.03782e3}, \beta = \num{4e-3}$ \\$  c_2=\num{1.5852e2},c_3=\num{-3.22770e3}$\end{tabular}}  &\num{1e-6}  & $10\delta t$ &$[10\delta t, 15\delta t]$  & $\num{1.6e6}$& Yes & $\num{1.6e7}$\\ \hline
\end{tabular}
\vskip 2em
\begin{tabular}{|c|c|c|c|c|c|c|c|c|c|c|c|}
\cline{2-11}
\multicolumn{1}{l|}{}                                                                                                          &  $L_{\text{int}}$     & $\chi_d^2(0.95)$ & $\hat\kappa$ & $\hat{R}_{\max}$ &$\lambda$ & $d_{\text{con}}$ & $d_{\text{thr}}$& $N_{\text{msm}}$ & $\rd t_{\text{msm}}$ & $N_{\text{IC}}$\\ \hline
\multicolumn{1}{|c|}{\begin{tabular}[c]{@{}c@{}}Pinched \\ sphere\end{tabular}}       &100      &5.991  &1 & 1   &  1& $ 3\sqrt{\tau}$ &$1.05\sqrt{\tau}$ & $10^6$ & $\tau$  & 15000\\ \hline
\multicolumn{1}{|c|}{\begin{tabular}[c]{@{}c@{}}Oscillating\\  half-moons\end{tabular}}  & 10    &  1.96& 1&0.4 & 1& $4\sqrt{\tau}$  & $0.95\sqrt{\tau}$  & $10^6$ & $\tau$  & 30000\\ \hline
\multicolumn{1}{|c|}{butane}                                                                                            & 2      & 1.96&  1&1.5  &1& $4\sqrt{\tau}$ & $0.95\sqrt{\tau}$ & $10^6$  &$\tau$ & 30000\\ \hline
\end{tabular}
\label{t:parameter}
\end{table*}

 \begin{figure}
\centering
{\includegraphics[width=0.32\textwidth, height=0.203\textwidth]{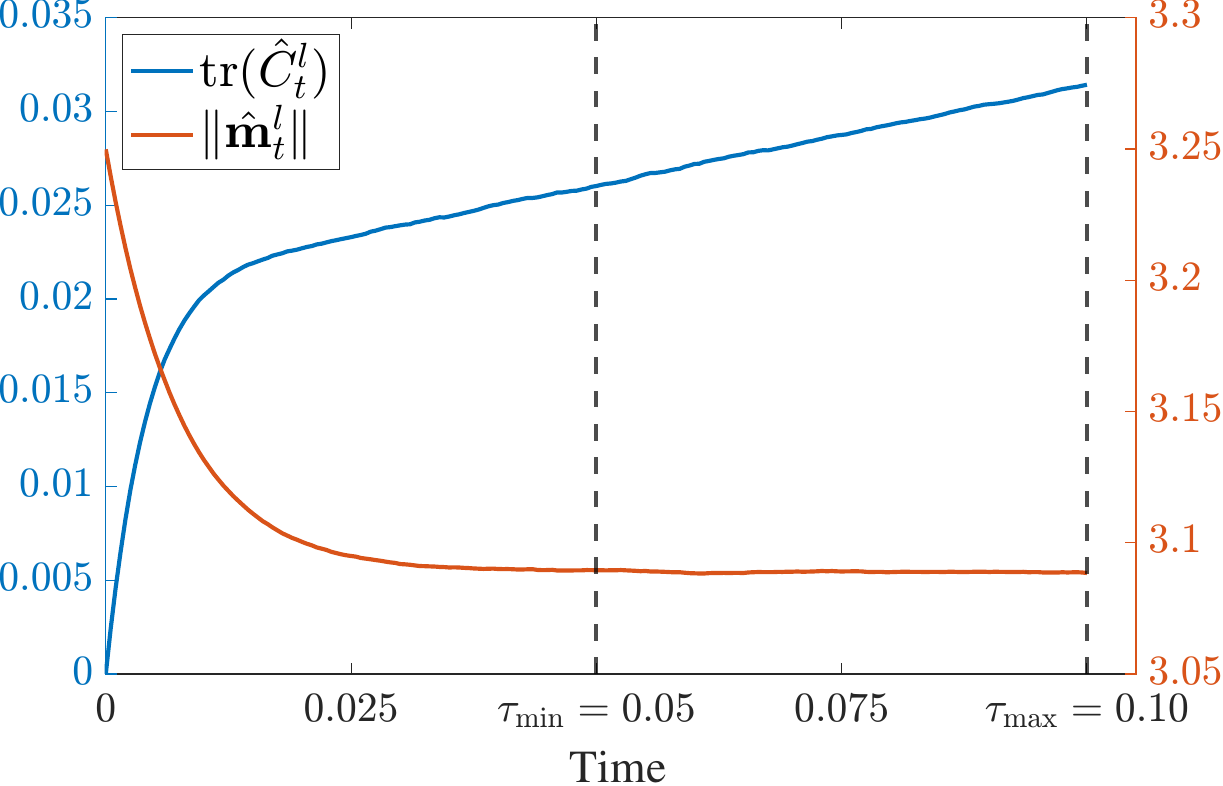}}
{\includegraphics[width=0.32\textwidth, height=0.203\textwidth]{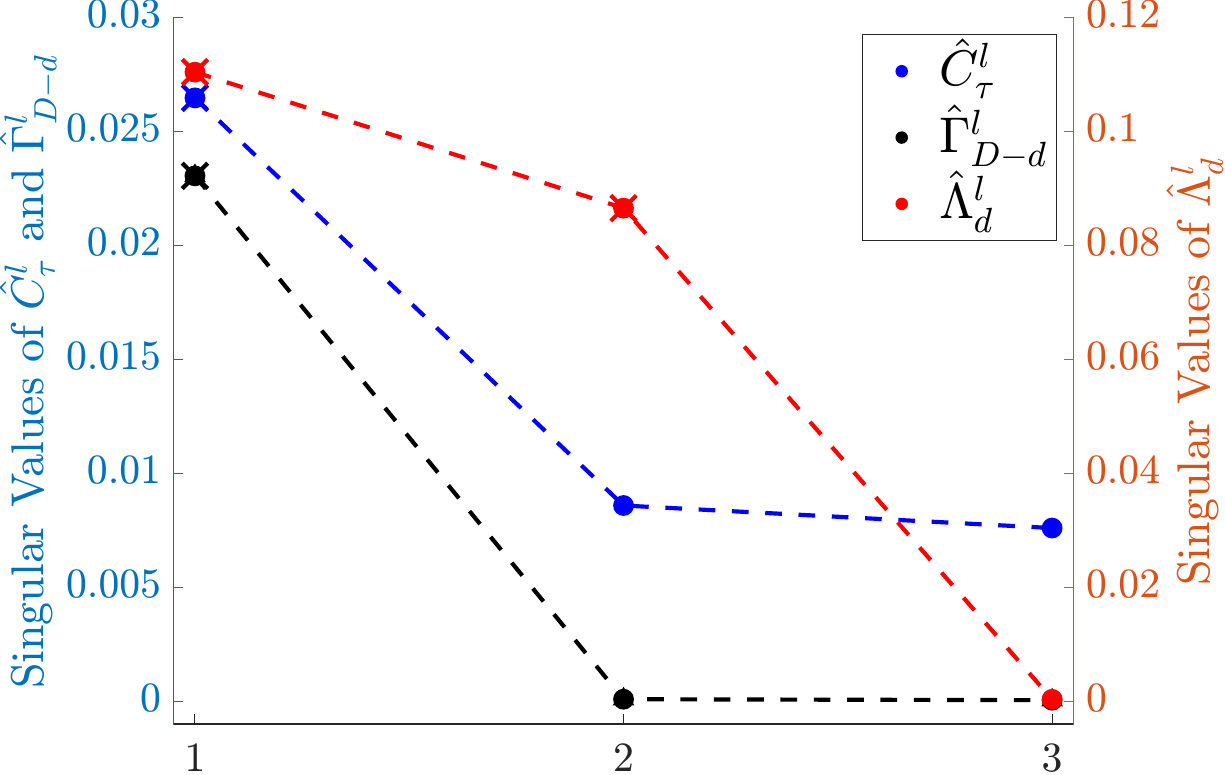}}
{\includegraphics[width=0.32\textwidth, height=0.203\textwidth]{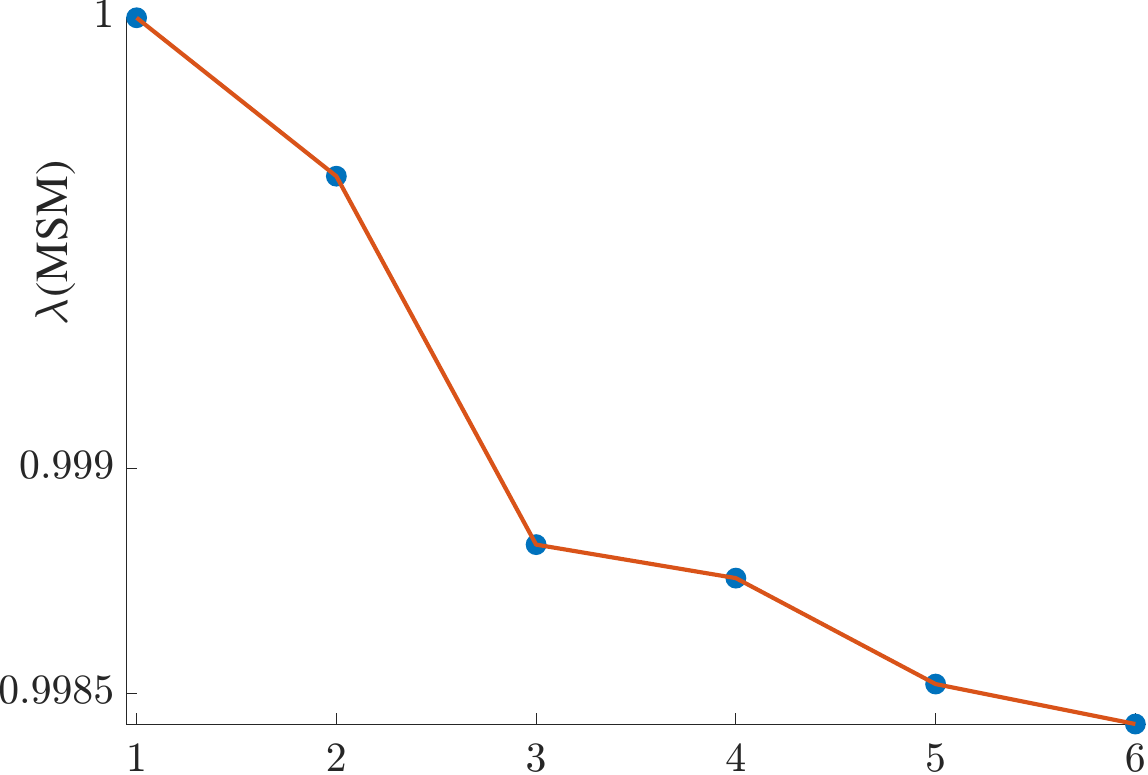}}
{\includegraphics[width=0.66\textwidth]{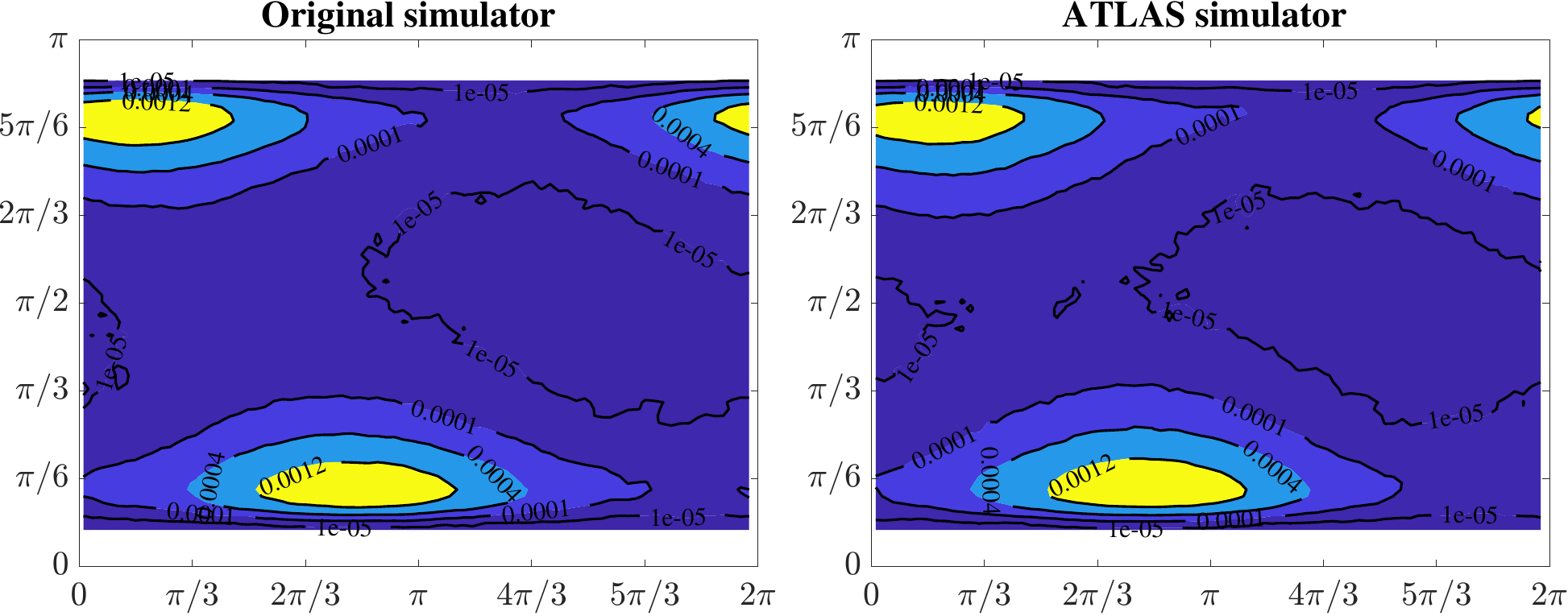}}
\includegraphics[width = 0.33\textwidth]{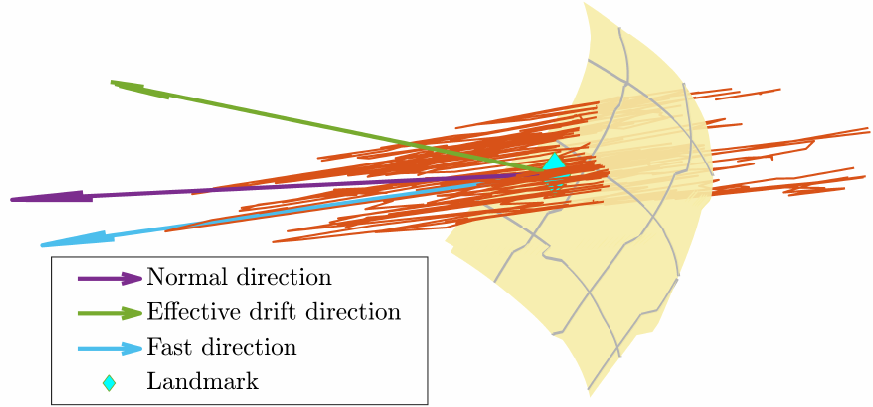}
\caption{  \Peanut: {\bf Top right:}  plot of trace of empirical covariance and norm of empirical mean, $\tr(\hat{C}_t^l)$ and $\|\hat{\mm}_t^l\|$ verse time at one landmark. {\bf Top center:} singular values of $\hat{C}_\tau^l, \hat{\Gamma}^{l}_{D-d}$ and $\LambdaldN$ at one landmark, the dominant singular values are marked as cross labels. {\bf Top left:}   top six eigenvalues of the Markov transition matrix. {\bf Bottom left:} the contour plot of kernel-fitted invariant distribution in the coordinate $(\phi, \theta)$ from the trajectory of time length $10^6$ generated by the original simulator.  {\bf Bottom center:} the contour plot of kernel-fitted invariant distribution in the coordinate $(\phi, \theta)$ from the trajectory of time length $10^6$ generated by the ATLAS simulator. {\bf Bottom right:} A trajectory of time-length 2.5$\tau$ starting from the marked landmark is simulated with the original simulator is shown. We depict a portion of the key estimated objects: the \IM \IMmth, drift at $\mzIC$, fast direction, and normal direction. Note that the fast direction is far from the normal one, and the (It\^o) drift is far from being tangent to \IMmth, and also that \IMmth is locally a graph over the tangent plane at $\mzIC$.}
\label{f:peanut-singular}
\end{figure}

\subsection{\Peanut model} \label{appx:peanut}
The governing slow-fast SDE in Cartesian coordinates becomes

\begin{equation}
\begin{aligned}
\rd\mz& =\left(J(\mz) \cdot \mb_1(\mz) + \frac{1}{2}A(\mz)\cdot \mb_2(\mz)\right)\rd t
                                                +     J(\mz)\cdot 
                                                  \sigma(\mz) \bcm \rd W_1 \\ \rd W_2 \\ \rd W_3 \ecm
\end{aligned}
\end{equation}
where 
\begin{small}
\begin{equation}
\begin{aligned}
&J= \bcm                                               \frac{z_1}{\|\mz\|}  & \frac{z_3z_1}{\sqrt{z_1^2 +z_2^2}}  & -z_2  \\ 
                                                             \frac{z_2}{\|\mz\|}  & \frac{ z_3z_2}{\sqrt{z_1^2 +z_2^2}} & z_1  \\ 
                                                             \frac{z_3}{\|\mz\|}  & -\sqrt{z_1^2 +z_2^2 }                       &   0     \ecm, 
\mb_1 =      \bcm  -\frac{c_1}{\eps} \frac{1-\sqrt{a_1 + a_2z_3^2/\|\mz\|^2}}{\|\mz\|}    \\
                                             c_3\frac{4z_3^3/\|\mz\|^3 -3z_3/\|\mz\|}{\sqrt{z_1^2+z_2^2}}          \\
                                             c_5 \left( \frac{z_2z_3}{\sqrt{z_1^2+z_2^2}\|\mz\|^2}+\frac{z_1}{\|\mz\|^2} \right)
                                    \ecm, 
      A = \bcm -z_1 & -z_1  \\ -z_2 & -z_2 \\ -z_3 & 0 \ecm,     \\                         
 &\mb_2  =      \bcm \frac{c_4^2 (z_1^2+z_2^2)}{\|\mz\|^4} \\ c_6^2/\|\mz\|^2 \ecm,
 \sigma = \text{diag}\bcm c_2/(\sqrt{\eps}\|\mz\|) \\ \frac{c_4\sqrt{z_1^2+z_2^2}}{\|\mz\|^2}  \\  c_6/\|\mz\| \ecm         
 \end{aligned}        
\end{equation}
\end{small}

Without the noise term, the deterministic counterpart of this system has two stable fixed points, $(\theta^*, \phi^*)=(\pi/6, 5\pi/6), (5\pi/6, \pi/6)$, which are marked in \cref{f:peanuteigen}.
 Note under the current parameter settings,  $c_6\gg c_5$ and $c_4\gg c_3$, the assumption of diffusion-dominated dynamics is satisfied.
We claimed in the main text that this system is not reversible, and we demonstrate it here.  The corresponding Fokker Planck equation of the governing equation in Cartesian coordinates is

\begin{align}
&\partial_t p(t, \mz) = -\nabla\cdot \left[ (J(\mz) \mb_1(\mz) +\frac{1}{2}A(\mz)\mb_2(\mz)) p(t, \mz)\right] +\nabla^2 \left[ D(\mz) p(t, \mz)\right]  \,.
\end{align}
where $D(\mz)= \frac{1}{2}(J(\mz)\sigma^2(\mz)J^T(\mz))$.
We transform the equation into the symmetric form
\begin{align}
&\partial_t p(t, \mz) =  -\nabla\cdot \left[ (J(\mz) \mb_1(\mz) +\frac{1}{2}A(\mz)\mb_2(\mz)-\nabla D(\mz)) p(t, \mz)\right] +\nabla\cdot \left[ D(\mz) \nabla p(t, \mz)\right]  \,.
\end{align}
where $\nabla D(\mz) = \left[ \sum_{j=1}^3\frac{\partial}{\partial z_j}D_{1j}(\mz),  \sum_{j=1}^3\frac{\partial}{\partial z_j}D_{2j}(\mz) ,  \sum_{j=1}^3\frac{\partial}{\partial z_j}D_{3j}(\mz) \right]^T$.

The theorem 3.3.7 in \cite{jqq} indicates that the system is reversible if and only if the force $\mb_3(\mz)=D^{-1}(\mz)(J(\mz) b_1(\mz) +\frac{1}{2}A(\mz)b_2(\mz)-\nabla D(\mz))$ 
is a conservative vector field. We calculate the curl of $\mb_3(\mz)$ is 

\begin{equation}
\nabla \times \mb_3(\mz) = \bcm    z_1f_1(z_1,z_2)+z_2f_2(z_1,z_2,z_3),   & z_2f_1(z_1,z_2)-z_1f_2(z_1,z_2,z_3), &\frac{2c_5z_1}{c_6^2 (z_1^2+z_2^2)}     \ecm^T 
\end{equation}
where 

\begin{equation}
f_1(z_1,z_2) =  -\frac{2c_5z_2}{c_6^2(z_1^2+z_2^2)^{3/2}}, \qquad f_2(z_1, z_2, z_3)= \frac{2c_3z_3(3(z_1^2+z_2^2)-z_3^2)}{c_4^2(z_1^2+z_2^2)^2 \|\mz\|}- \frac{2a_2c_1z_3}{c_2^2\|\mz\|^2 \sqrt{a_1+\frac{a_2z_3^2}{\|\mz\|^2}}}
\end{equation}
The curl is not zero if all coefficients are positive. So this system is not reversible. 

\noindent{\textbf{Identifying the slow manifold and the effective stochastic dynamics}}.  The \SM{} $\SMxmth$ in spherical coordinate is $r^\star(\theta)=R(\theta)$ in the limit of $\eps\rightarrow 0$. Since Cartesian coordinates $z_1, z_2, z_3$ are linear with the radius $r$, the \SM{} $\SMmth$ in Cartesian coordinates is $[R(\theta)\sin(\theta)\cos(\phi), R(\theta)\sin(\theta)\sin(\phi), R(\theta)\cos(\theta)]^T$ in the limit of $\eps\rightarrow 0$. In this example, $\SMmth$ is the image of $\SMxmth$ under the coordinate transformation.  It is uniformly asymptotically stable and it has negative and positive curvature at different points (with the chosen set of parameter $a_1, a_2$). 
The reduced dynamics (in the limit $\eps\rightarrow0$) on the \SM{} \SMmth, in Cartesian coordinates, is given as 

\begin{equation}
\rd \mz = \mb^{\text{eff}}\rd t + H^{\text{eff}}\bcm \rd W_2 \\ \rd W_3 \ecm   
\end{equation}
where 

\begin{equation}
\begin{aligned}
 \mb^{\text{eff}}&= \bcm \frac{\rd }{\rd \theta}(R(\theta)\sin(\theta)) \cos(\phi) & -R(\theta)\sin(\theta)\sin(\phi)    \\ 
                                                              \frac{\rd }{\rd \theta}(R(\theta)\sin(\theta)) \sin(\phi) &  R(\theta)\sin(\theta)\cos(\phi)   \\ 
                                                                \frac{\rd }{\rd \theta}(R(\theta)\cos(\theta))  &             0   \ecm\cdot  \bcm  \frac{c_3\cos(3\theta)}{R(\theta)\sin(\theta)} \\ \frac{c_5\sin(\phi+\theta)}{R(\theta)} \ecm 
                                                                +\frac{1}{2} \bcm  \frac{\rd^2 }{\rd \theta^2}(R(\theta)\sin(\theta)) \frac{c_4^2\sin^2(\theta)\cos(\phi)}{R^2(\theta)} - \frac{c^2_6 \sin(\theta)\cos(\phi) }{R(\theta)}  \\
                                                                                             \frac{\rd^2 }{\rd \theta^2}(R(\theta)\sin(\theta)) \frac{c_4^2\sin^2(\theta)\sin(\phi)}{R^2(\theta)} -  \frac{c^2_6 \sin(\theta)\sin(\phi) }{R(\theta)}    \\
                                                                                             \frac{\rd^2 }{\rd \theta^2}(R(\theta)\cos(\theta)) \frac{c_4^2\sin^2(\theta)}{R^2(\theta)} \ecm,  \\
H^{\text{eff}}&=   \bcm \frac{\rd }{\rd \theta}(R(\theta)\sin(\theta)) \cos(\phi) & -R(\theta)\sin(\theta)\sin(\phi)    \\ 
                                                              \frac{\rd }{\rd \theta}(R(\theta)\sin(\theta)) \sin(\phi) &  R(\theta)\sin(\theta)\cos(\phi)   \\ 
                                                                \frac{\rd }{\rd \theta}(R(\theta)\cos(\theta))  &             0   \ecm \cdot       \bcm     \frac{c_4\sin(\theta)}{R(\theta)} & 0 \\  0 &    \frac{c_6}{R(\theta)}  \ecm                                                              
\end{aligned}
\end{equation}

and $\theta = \arctantwo(\sqrt{z_1^2 +z_2^2}, z_3), \phi =\text{mod}(\arctantwo(z_2, z_1),2\pi)$. This result is not exactly the effective dynamics (averaged at the finite time-scale $\tau$) on the \IM{} \IMmth, but it provides us some good reference. Here, and in the other numerical experiments, we will in fact consider this as ground truth, for measuring the accuracy of the ATLAS estimators of geometric objects (e.g. \IMmth) and dynamics quantities (e.g. $\mathbf{b}$ and $\Lambda$). Note that when we measure the accuracy of other quantities such as mean residence times and accuracy of the stationary distributions, these approximations are not used, and the ATLAS estimates are compared directly with those obtained from the original simulator.

\noindent{\textbf{Estimating the relevant time scale $\tau$}}. In \cref{f:peanut-singular} we visualize the behavior of $\tr(\hat{C}_t^l)$ and $\|\hat{\mm}_t^l\|$ as a function of time: we observe that they initially behave nonlinearly, but then transit to a linear regime, at time about 0.025, consistent with the approximations in \cref{e:meanAndCovBurst}. In this example, we choose $(\tau_{\min}, \tau_{\max})=(0.05, 0.10)$ and the timescale of separation $\tau=0.10$. 

\noindent{\textbf{Estimating dimension and tangent spaces of \SMmth, and direction of the fast modes}}. We all report in \cref{f:peanut-singular} the behavior of the singular values of $\hat C_\tau^l$, $\hat\Gamma^l_{D-d}$ and $\LambdaldN$ at one landmark in descending order.  
The dominant singular values are visualized with cross markers. 
Local Principal Component Analysis (PCA) corresponds to the analysis of $\hat{C}_N(\tau)$: it exhibits 1 dominant singular value, with the corresponding singular vector close to the direction of fast variables, because the large fluctuations of the fast modes dominates. The other two singular vectors of $\hat{C}_N(\tau)$, which are necessarily, in PCA, orthogonal to the leading mode, are not the directions of slow variables at most regions.
The number of the dominant singular values of $\hat{\Gamma}^l_{D-d}$ 
 is 1 and its corresponding singular vector correctly estimates the direction of the fast variable. 
Finally, the number of the dominant singular values of $\LambdaldN$ 
  is 2, equal to the correct dimension of the \SM, and the span of their corresponding singular vectors correctly estimates the tangent space of the \SM. {On average, it requires 1585 charts to fully describe the invariant manifold. }

\noindent{\textbf{Identifying metastable states, and estimation of large-time properties of the process}}. We assume we do not know the metastable states, nor the number of such states.  The top six eigenvalues of the Markov transition matrix (see \cref{f:peanut-singular}) are real and they exhibit a clear spectral gap after the top two eigenvalues, which indicates, correctly, that this system has 2 metastable states. We simulate trajectories of time length $10^6$ with the original simulator and with the ATLAS simulator, and from those we estimate the invariant distribution by using smoothed histograms with bins constructed in the latent coordinates $(\phi, \theta)$, which parametrize the invariant manifold. We visualize them in \cref{f:peanut-singular}: from the contour plot of both distributions they appear very close, and indeed in the $L^1$-norm of the difference of two densities (i.e. the total variation distance between the distributions) is $ 0.107\pm0.009$, and the $L^2$-norm of the difference of densities (a more robust but less stringent distance) is $0.0034\pm0.0004$.

\noindent{\textbf{Estimation of residence times}}. In the stage of estimating residence times, we generate a single long trajectory of time length $\num{5e5}$ with the ATLAS simulator (this time length is much larger than the residence times to be estimated), from which we uniformly sample $N_{\text{IC}}$ initial conditions that are in $S_{\text{cyan}}=\{\varphi_2>0.05\}$ and $S_{\text{red}}=\{\varphi_2<-0.05\}$. Here one can use either original simulator or ATLAS simulator, however, the ATLAS simulator is much faster. In this example, we test the medium and large residence time by defining the boundary of the residence set as $\{\varphi_2>0.02\}$ and $\{\varphi_2<-0.02\}$ in the first experiments, and, the metastable states $M_1$ and $M_2$ in the second experiment.  We say that a point on \IM is in a set if the closest landmark of the point is in the corresponding set (with this definition, the set is slightly different from the corresponding sublevel set or suplevel set of an eigenfunction).
\begin{figure}
\centering
{\includegraphics[width=0.32\textwidth]{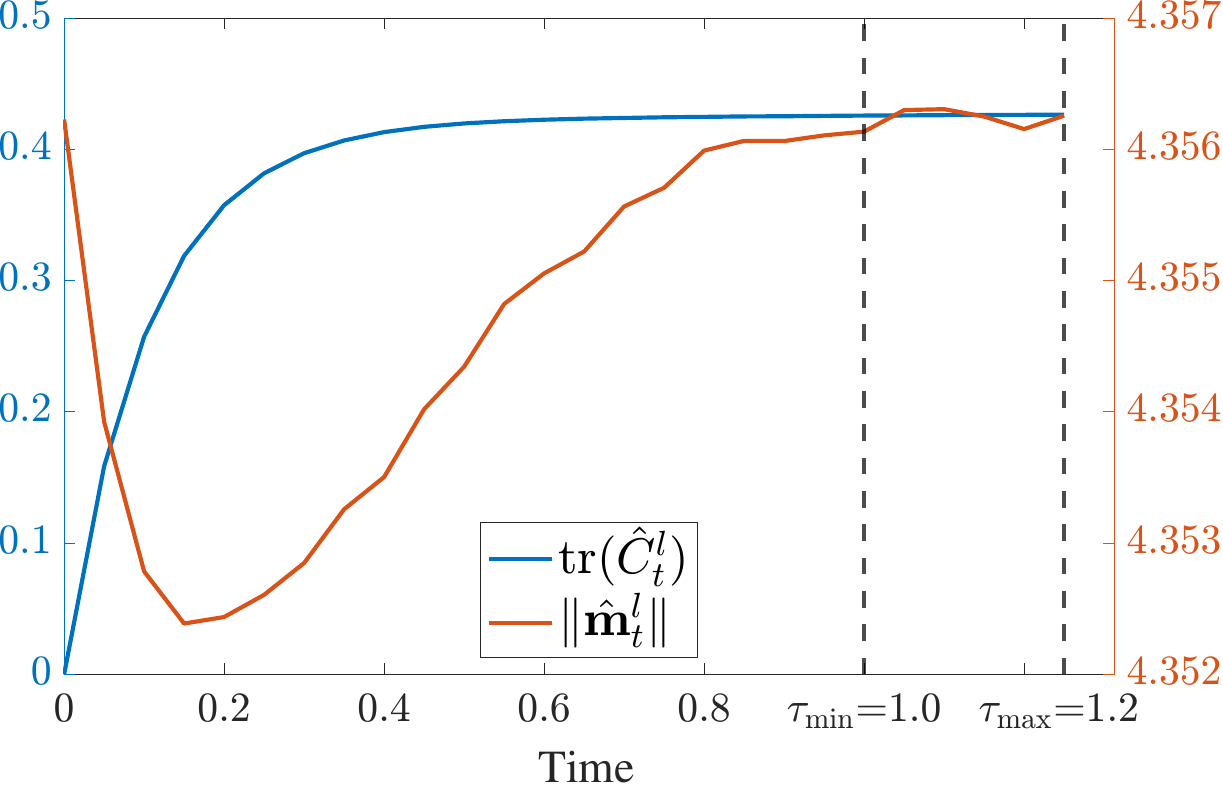}}
{\includegraphics[width=0.32\textwidth]{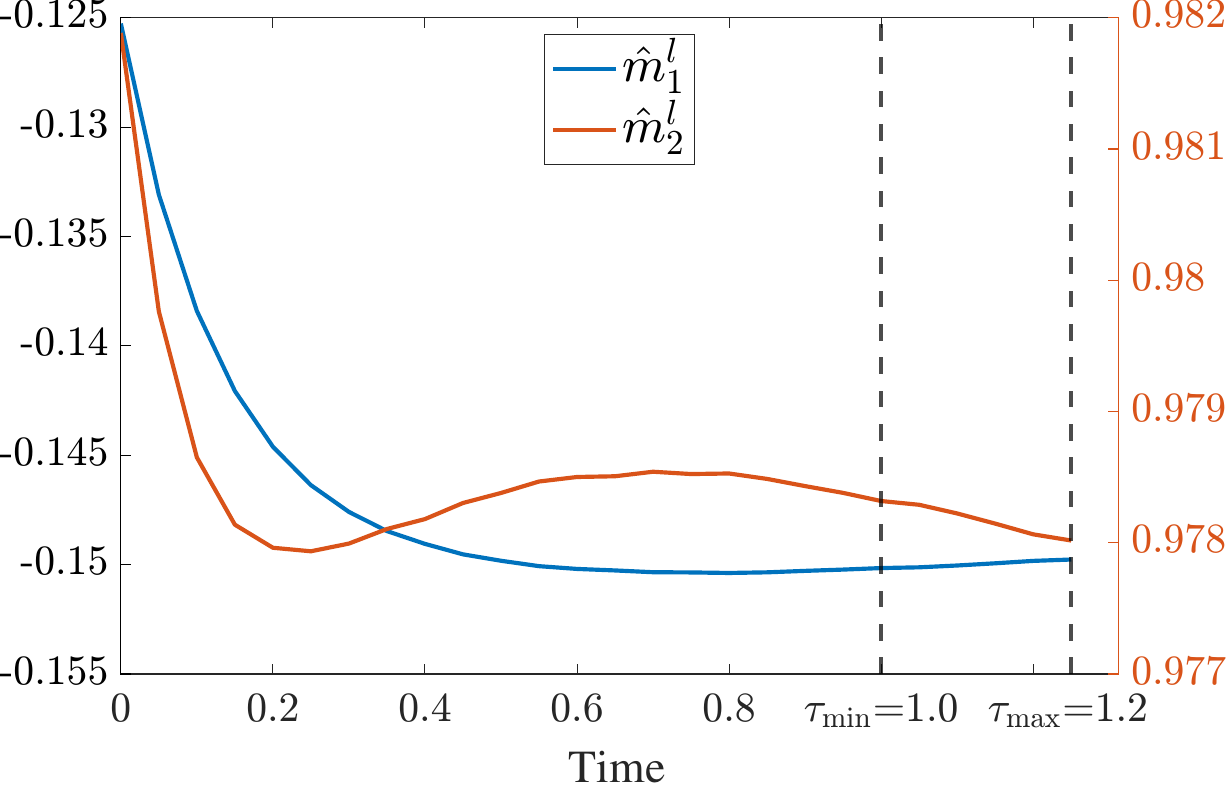}}
{\includegraphics[width=0.32\textwidth]{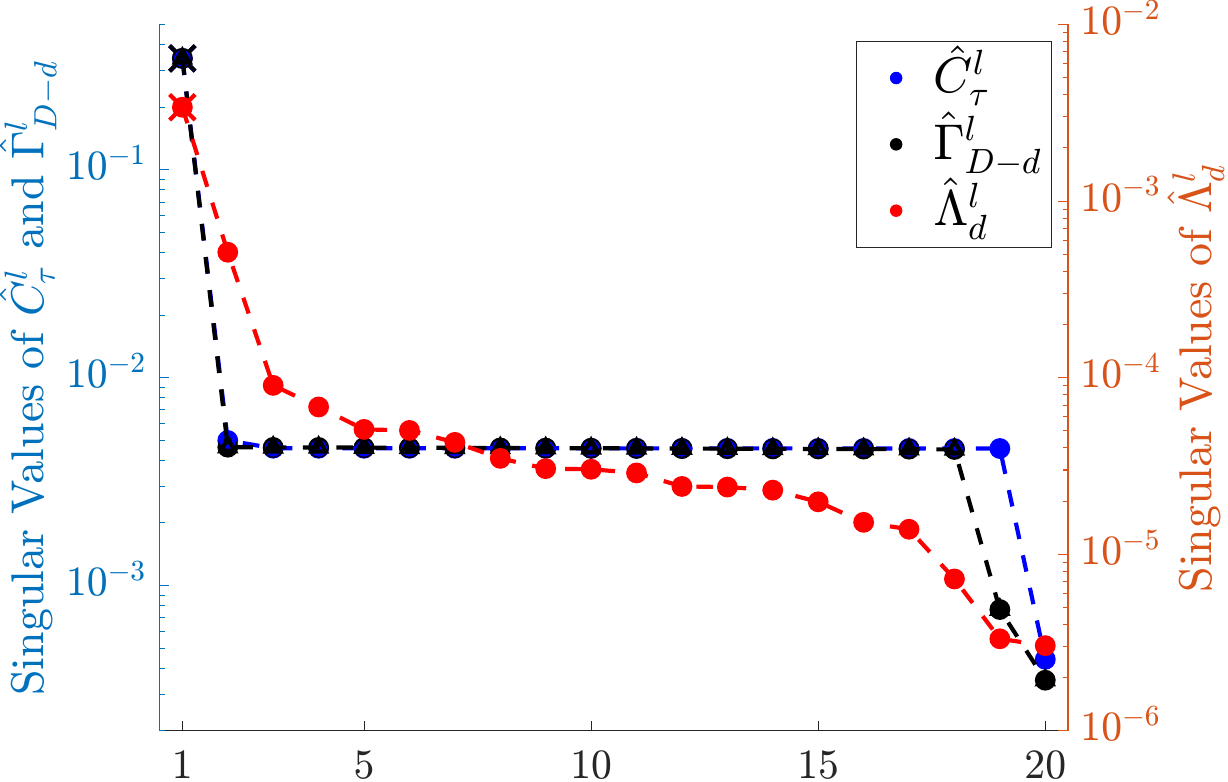}}
{\includegraphics[width=0.33\textwidth, height=0.203\textwidth]{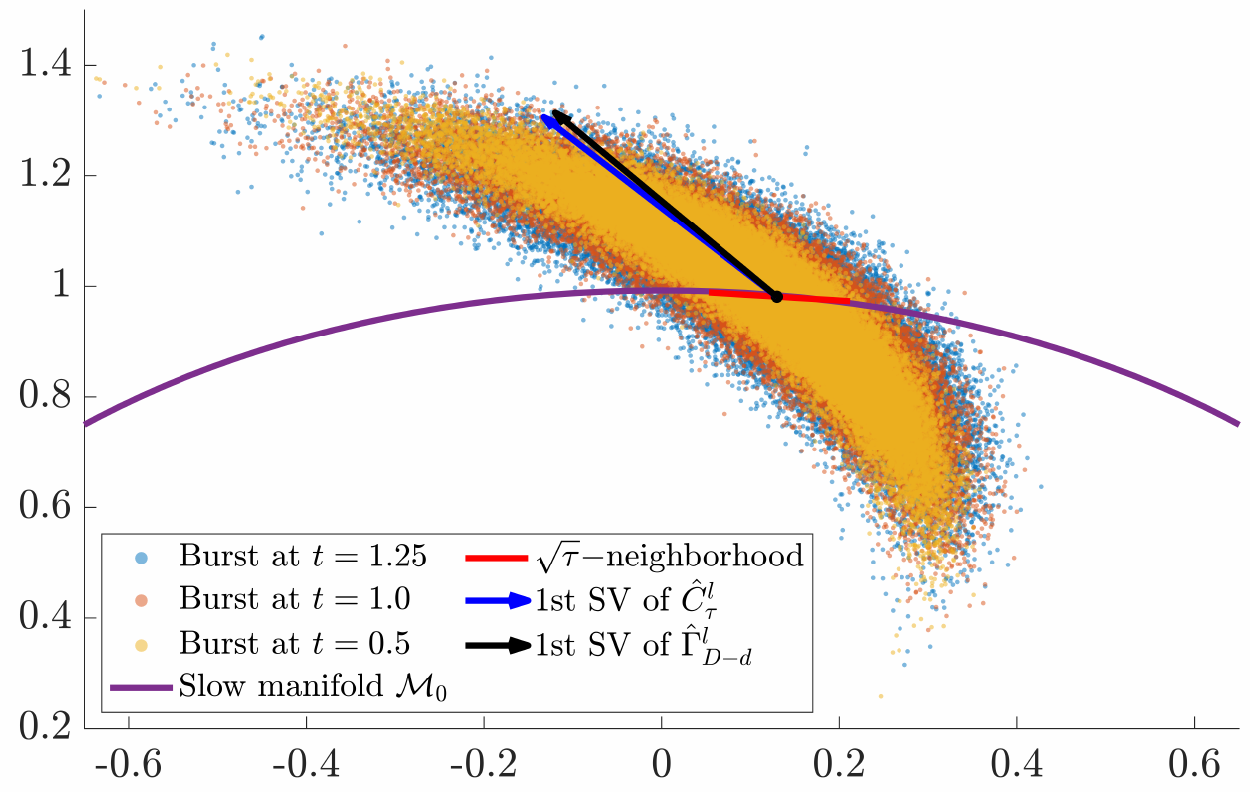}}
{\includegraphics[width=0.33\textwidth, height=0.203\textwidth]{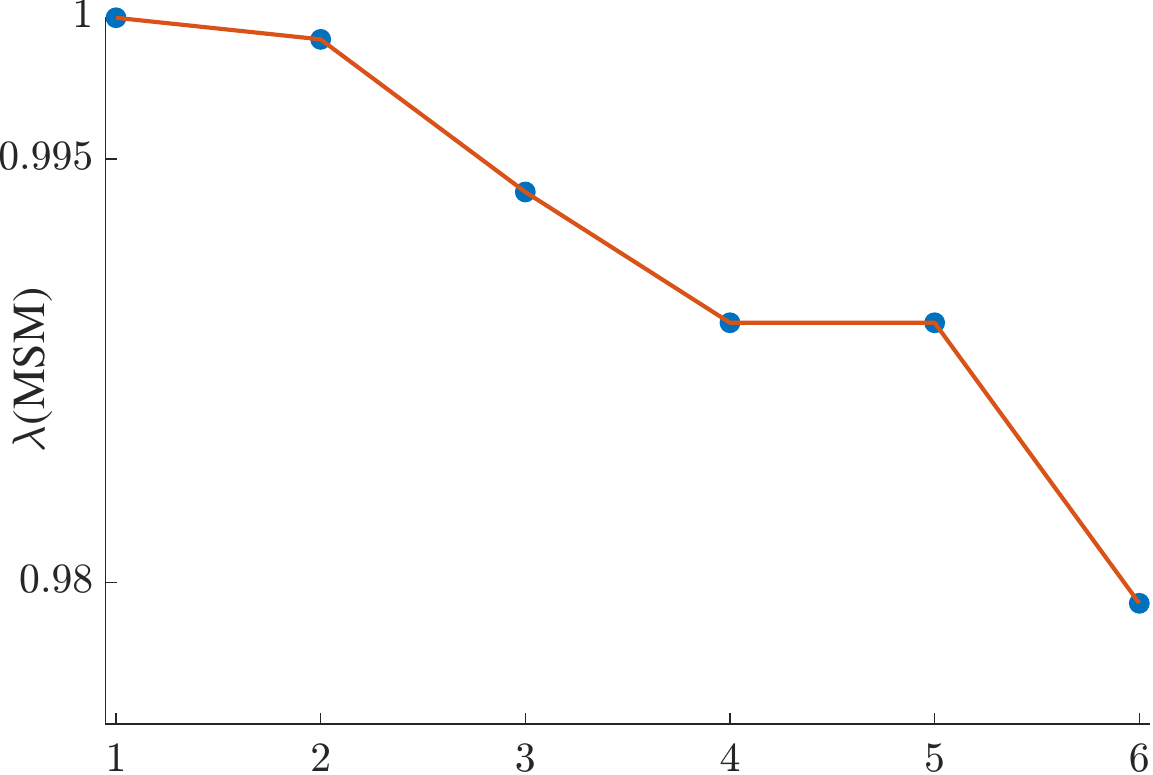}}
\caption{ Oscillating half-moons: {\bf Top left:}  plot of trace of empirical covariance and norm of empirical mean, $\tr(\hat{C}_t^l), \|\hat{\mm}_t^l\|$ verse time at one landmark. {\bf Top center:} plot the first and second coordinates of the empirical mean $\hat{\mm}_t^l$ verse time at one landmark.
{\bf Top right:} singular values in log scale of $\hat{C}_\tau^l, \hat{\Gamma}^{l}_{D-d}$ and $\LambdaldN$ at one landmark, the dominant singular values are marked as cross labels. {\bf Bottom left:} scatter of the burst in the $(z_1, z_2)$ coordinate at $t=0.5, 1.0, 1.25$, together with  the \SM, $\sqrt{\tau}$-neighborhood and dominant singular vectors of $\hat{C}_\tau^l, \hat{\Gamma}^{l}_{D-d}$.
 {\bf Bottom right:}   top six eigenvalues of the Markov transition matrix.}
\label{f:halfmoon20}
\end{figure}

\subsection{Oscillating Half-moons} \label{appx:halfmoon}
 
\noindent{\textbf{Identifying the slow manifold and the effective stochastic dynamics}}. With the setting of parameter, our model is reversible since $a_1=0$\cite{gq-chaos}.  In the latent variable space, the fast variables are $r_1$ and $u_i$ and the slow variable is $\theta$. In the limit $\eps\rightarrow 0$,  the fast variables relax to the equilibrium at  $r_1=1$ and $u_i=0$ so the \SM{} $\SMxmth$ in the latent variable is $r_1=1$, which is the unit circle. The "local" invariant distribution of the fast variable $r_1$ and $u_i$ are 

\begin{align}
&\nu(r_1|\theta)=\sqrt{\frac{b_1}{b_2^2\pi }}\exp\left[ -\frac{b_1}{b_2^2}\left(r_1-1\right)^2\right],\qquad \nu(u_i|\theta)=\sqrt{\frac{b_3}{b_4^2\pi }}\exp\left[ -\frac{b_3}{b_4^2}\left(u_i\right)^2\right].
\end{align}
With the current parameter setup, there is very small probability that $r_1$ can go to the negative side but we will ignore the possibility in our analytical calculations.

In the example, the fast variables are nonlinearly coupled in the observed Cartesian coordinates, so the slow manifold $\SMmth$ in Cartesian coordinates is not the image of $\SMxmth$ under the coordinate transformation. 
In the limit $\eps\rightarrow 0$, the landmark position in Cartesian coordinate for given $\theta$  is 
\begin{align}
&\bar{z}_1(\theta):= \mathbb{E}(z_1|\theta)=\int_{-\infty}^{+\infty} r_1\cos(r_1+\theta-1)\nu(r_1|\theta) \rd r_1  = \exp\left(-\frac{b_2^2}{4b_1}\right)\sqrt{1+\left(\frac{b_2^2}{2b_1}\right)^2} \cos(\theta+ \theta_s), \\
&\bar{z}_2(\theta):= \mathbb{E}(z_2|\theta)=\int_{-\infty}^{+\infty} r_1\sin(r_1+\theta-1) \nu(r_1|\theta) \rd r_1 =  \exp\left(-\frac{b_2^2}{4b_1}\right)\sqrt{1+\left(\frac{b_2^2}{2b_1}\right)^2}  \sin(\theta+\theta_s), \\
&\bar{z}_i(\theta) := \mathbb{E}(z_i|\theta) = \int_{-\infty}^{+\infty}\left( \int_{-\infty}^{+\infty}  \left( r_1 + u_i\right)\nu(r_1|\theta) \rd r_1 \right) \nu(u_i|\theta)\rd u_i =1.
\end{align}
where the angle shift $\theta_s$ is $\theta_s= \arctan \left(\frac{b_2^2}{2b_1}\right)$. Then the slow manifold $\SMmth$ is the circle embedded in $(\bar{z}_1, \bar{z}_2, 1, \dots, 1)$ with the radius $ \bar{r}=\exp\left(-\frac{b_2^2}{4b_1}\right)\sqrt{1+\left(\frac{b_2^2}{2b_1}\right)^2}$ and the angle is shifted by $\theta_s$ compared to the standard angle in Cartesian coordinates. This shift is due to the nonlinearity of the fast modes, in particularly their curvature. 
 With current parameter setup, the radius is approximately $\bar{r}=0.9925$ and $\theta_s$ is approximately 0.0153. 

The distance of the point from the slow manifold $\SMmth$, 
$\text{dist}(\mz, \SMmth) = \sqrt{(\sqrt{z_1^2+z_2^2}-\bar{r})^2+\sum_{i=3}^{20}(z_i-1)^2}$.
In Cartesian coordinate, the effective dynamics of the first and second coordinate, $\bar{z}_1, \bar{z}_2$ are 
\begin{align}
    \rd \bcm \bar{z}_1(\theta) \\   \bar{z}_2(\theta) \ecm = \left(\left( a_1 +a_2 \sin(2\theta)+a_3\cos(\theta) \right)    \bcm -\bar{z}_2(\theta)   \\ \bar{z}_1(\theta) \ecm   -\frac{a_4^2   }{2}       \bcm   \bar{z}_1(\theta) \\  \bar{z}_2(\theta)  \ecm  \right) \rd t +  a_4 \bcm -\bar{z}_2(\theta)   \\ \bar{z}_1(\theta) \ecm  \rd W_t
    \end{align}
The effective dynamics on the other 18 coordinates has zero drift term and zero diffusion term. 

\noindent{\textbf{Estimating the relevant time scale $\tau$}}. In \cref{f:halfmoon20}, $\tr(\hat{C}_t^l)$ reaches the linear regime at $t=0.5$, however, the norm of the empirical mean $\|\hat{\mm}_t^l\|$ behaves linearly only after $t=1.0$ (this is true also of the first and the second coordinates which are also TICA coordinates, $\hat{m}_1^l, \hat{m}_2^l$). This is consistent with 
the large fast modes, and high curvature of the slow manifold; the training interval that we choose is $[\tau_{\min}, \tau_{\max}]= [1.0, 1.25]$, and $\tau=1.0$. 

\noindent{\textbf{Estimating dimension and tangent spaces of \SMmth, and direction of the fast modes}}.  We report in \cref{f:halfmoon20} the behavior of the singular values of $\hat C_\tau^l$, $\hat\Gamma^l_{D-d}$ and $\LambdaldN$ at one landmark in descending order.  The dominant singular values are visualized with cross markers. 
Although the number of the dominant singular value of covariance matrix $\hat{C}_\tau^l$ and the diffusivity matrix $\hat{\Lambda}^l$ are 1, the corresponding singular vector of $\hat{C}_\tau^l$ is not the slow direction, but close to fast direction, dooming a na\"ive approach based on local PCA. On the other hand, the dominant singular vector of $\LambdaldN$ correctly estimates the tangent direction of the \SM.
The number of the dominant singular values of covariance matrix $\hat\Gamma^l_{D-d}$ is 1, which is the correct number of fast variables.
The scatter plot of the samples from paths of a burst, at $t=0.5, 1.0, 1.25$ in the global TICA coordinates $(z_1, z_2)$, visualizes how the data cloud is stretched out nonlinearly away from the \SM.  Due to the relatively big curvature, the part of \SM{} that these data clouds covers at different time clearly show the nonlinear effects. A local linear approximation on dynamics and geometry might not be accurate at these scales, making the refinement of the landmark positions necessary, for one round of refinement as discussed in  \cref{s:ATLASconstruction}.  This procedure significantly reduces the bias introduced by the local linear approximation.  To further reduce the effect of the nonlinearity, we choose the effective timescale $\tau$ as lower bound of the training interval. Here the scale of separation is large enough, multiple rounds of refinement are not necessary. {On average, it requires 65 charts to fully describe the invariant manifold. }

\noindent{\textbf{Identifying metastable states, and estimation of large-time properties of the process}}.
From the top six eigenvalues of the transition matrix of a Markov state model constructed from ATLAS, we note the significant gap between the second and third eigenvalues, which correctly indicates the system has two metastable states. In this example, we explicitly provide the regions of the two metastable states, $M_{\text{Left}}=\{\theta\in\left(-2\tan^{-1}(4+\sqrt{15}),-2\tan^{-1}(4-\sqrt{15}) \right)\}$ and $M_{\text{Right}}=\{\theta\in\left(-2\tan^{-1}(4-\sqrt{15}),-2\tan^{-1}(4+\sqrt{15}) +2\pi\right)\}$ in the latent variable $\theta$, that parametrizes the slow manifold.

We simulate trajectories of time $\num{8e6}$ with the original simulator and with the ATLAS simulator,  and from those we plot the invariant distributions in the latent variable $\theta$. The standard bin width is $a_4\sqrt{\tau}=0.06$ and we use it for the plot and calculation of the difference of the two distributions, which are very close (see \cref{f:halfmoon}).
The $L^1$-norm of the difference of two approximated densities is $0.098\pm0.006$ and $L^2$-norm of the difference is $0.015\pm0.001$. Another observation is that albeit ATLAS is only expected to be accurate for the standard bin width, we also attempted other smaller bin widths, all the way down to $0.0019$, which is much smaller than $a_4\sqrt\tau$, and observed no increase in the estimation error, thanks to the regularity of such distributions. 

\noindent{\textbf{Estimation of residence times}}.  In the stage of simulating residence time, we generate a single long trajectory of time $\num{2.4e6}$ by the original simulator and uniformly sample $N_{\text{IC}}$ initial conditions in each metastable state.  The boundary of the residence set is the same as the region of the metastable state. In this example, we say the point is in the residence set if its latent variable $\theta$ is in the corresponding metastable state. 
\subsection{Butane} \label{appx:butane}

\noindent{\textbf{Governing equations for butane dynamics}}. 
We consider the overdamped Langevin dynamics of the butane molecule. The positions are denoted as  $q^{i}\in \mathbb{R}^3$ for $1\le i \le 4$. To remove the rigid body motion invariant, we set 

\begin{align*}
q^1=\bcm x_1 \\ y_1 \\0 \ecm, q^2 =\bcm 0 \\ 0\\0\ecm, q^3 = \bcm 0 \\y_3 \\0 \ecm, q^4 = \bcm x_4\\y_4\\ z_4 \ecm.  
\end{align*}
The potential energy is given as 

\begin{align*}
V = \sum_{i=1}^3 V_{\text{bond}}\left(\|q^{i+1}-q^i\|\right) + \sum_{i=1}^2 V_{\text{angle}}\left(\theta_i\right) + V_{\text{torsion}}(\phi).
\end{align*}
where $\theta_1, \theta_2$ are the angles formed by the three first atoms and the three last atoms respectively.  $\phi$ is the dihedral angle, i.e, the angle between the plane on which the first three atoms lay and the plane on which the three last atoms lay. These potential functions are 
$V_{\text{bond}}( l)=\frac{k_2}{2}\left(  l - l_{eq}\right)^2, V_{\text{angle}}(\theta) =\frac{k_3}{2}\left( \theta-\theta_{eq}\right)^2$
 and 
$V_{\text{torsion}}(\phi) = c_1\cos\phi+ c_2\cos^2\phi+c_3\cos^3\phi$.
The numerical values for the constants are those in \cite{TraPPE}. The overdamped Langevin dynamics on the state space $\bR^6$ is 

\begin{align}
\rd \mz = -\nabla  V(\mz) \rd t + \sigma\rd W_t
\end{align}
where $\mz = \bcm x_1 & y_1 & y_3& x_4 & y_4& z_4 \ecm^T$ and the diffusion coefficient is $\sigma=\sqrt{2\beta^{-1}}$. 
The dihedral angle $\phi$ has the explicit form when $x_1<0$,

\begin{equation}
\cos(\phi) = \frac{(\vec{v}_{43}\times \vec{v}_{32}) \cdot (\vec{v}_{12}\times \vec{v}_{23})}{|\vec{v}_{43}\times \vec{v}_{32}| \cdot |\vec{v}_{12}\times \vec{v}_{23}|}=\frac{x_4}{\sqrt{x_4^2+z_4^2}},
\end{equation}
where $\vec{v}_{ij}= q^j-q^i$.
If we define the counterclockwise rotation as positive, then  $x_4= l\sin(\theta_{eq})\cos(\phi)$ and $z_4= l\sin(\theta_{eq})\sin(\phi)$.
If $x_1<0, y_3>0$,  the explicit form of the potential is 

\begin{align}\nonumber
&V(x_1,y_1,y_3, x_4, y_4, z_4) = c_1\frac{x_4}{(x_4^2+z_4^2)^{1/2}} + c_2\frac{x_4^2}{x_4^2+z_4^2} +c_3\frac{x_4^3}{\left(x_4^2+z_4^2\right)^{3/2}}  \\ \nonumber
&+ \frac{1}{2}k_2 \left( \left(\sqrt{x_1^2 +y_1^2}- l\right)^2 + \left(\sqrt{y_3^2}- l\right)^2 +\left(\sqrt{x_4^2+z_4^2 + \left(y_3-y_4\right)^2}- l\right)^2\right) + \\
& \frac{1}{2}k_3\left(\left(\theta_{eq} - \arccos\left( \frac{y_1}{\sqrt{x_1^2+y_1^2}}\right) \right)^2 + \left(\theta_{eq}-\arccos\left(\frac{(y_3-y_4)}{\sqrt{x_4^2+(y_3-y_4)^2+z_4^2}}\right)\right)^2\right).
\end{align} 

\noindent{\textbf{Identifying the slow manifold and the effective stochastic dynamics}}.
The dihedral angle $\phi$ is usually chosen as the slow variable of the butane dynamics and we will have the slow manifold $\SMmth$, which is a circle embedded in $\mathbb{R}^6$, if $x_1>0, y_3<0$,

\begin{align*}
&\SMmth=\left\{\bcm - l\sin(\theta_{eq}), &  l\cos(\theta_{eq}), &  l, & x_4,&  l- l\cos(\theta_{eq}),&z_4\ecm^T \ \text{with}\ x_4^2 + z_4^2 =  l^2\sin^2(\theta_{eq}) \right\}
\end{align*}
The potential energy for the dihedral angle has three local minima at $\phi=-2\pi/3$ (bot-cis), $\phi=0$(trans) and $\phi=2\pi/3$ (top-cis).  Butane dynamics is well known for its conformation isomerism and can be treated as the unimolecular reaction of three states. The trans conformer has lower energy than the cis conformer, so the trans state is more stable than the cis state. 
The distance of the point from the \SM{} $\cM_0$ is calculated as follows, 

\begin{align}
\text{dist}(\mz, \SMmth)=\sqrt{ \left(x_1+ l\sin(\theta_{eq})\right)^2 + \left(y_1 - l\cos(\theta_{eq})\right)^2 +\left(y_3- l \right)^2 +\left(y_4+  l\cos(\theta_{eq})- l \right)^2 +\left(\sqrt{x_4^2 + z_4^2 } -  l\sin(\theta_{eq})\right)^2}
\end{align}
The proposed effective stochastic dynamics of the dihedral angle $\phi$ is 

\begin{align}
\rd \phi_t = -\nabla V_{\text{torsion}}(\phi_t) \rd t + \sqrt{2\beta^{-1}}\sigma(\phi_t) \rd W_t
\end{align} 
 As suggested in \cite{Legoll2}, the $\sigma(\phi_t)$ is  
$\sigma^2(\phi_t) = \mathbb{E}\left(|\nabla \phi|^2(\fvar) |\phi(\fvar)=\phi_t\right)$.
In this case, $\sigma(\phi_t)$ can be explicitly calculated here, $\sigma(\phi_t) =  \frac{1}{ l \sin(\theta_{eq})}$.  
Then in $\mathbb{R}^6$, the explicit form of the effective stochastic dynamics of the fourth and sixth coordinates, $x_4, z_4$ are 

\begin{align}
\partial_t x_4 & = \left(-z_4^2 \frac{c_1\left(x_4^2+z_4^2\right)+2c_2x_4\sqrt{x_4^2+z_4^2}  +3c_3x_4^2 }{(x_4^2+z_4^2)^{5/2}}  - \frac{x_4}{\beta l^2 \sin^2(\theta_{eq})}\right)\rd t -\frac{z_4 \sqrt{2\beta^{-1}}}{l \sin(\theta_{eq})} \rd W_t, \\
\partial_t z_4 &= \left(x_4z_4 \frac{c_1\left(x_4^2+z_4^2\right)+2c_2x_4\sqrt{x_4^2+z_4^2} +3c_3x_4^2  }{(x_4^2+z_4^2)^{5/2}}  - \frac{z_4}{\beta l^2 \sin^2(\theta_{eq})}\right)\rd t +\frac{x_4 \sqrt{2\beta^{-1}}}{l \sin(\theta_{eq})}\rd W_t.
\end{align}
Other variables, $x_1, y_1, y_3, y_4$ have no drift and no diffusion in the effective stochastic dynamics.  

\begin{figure}
\centering
{\includegraphics[width=0.32\textwidth]{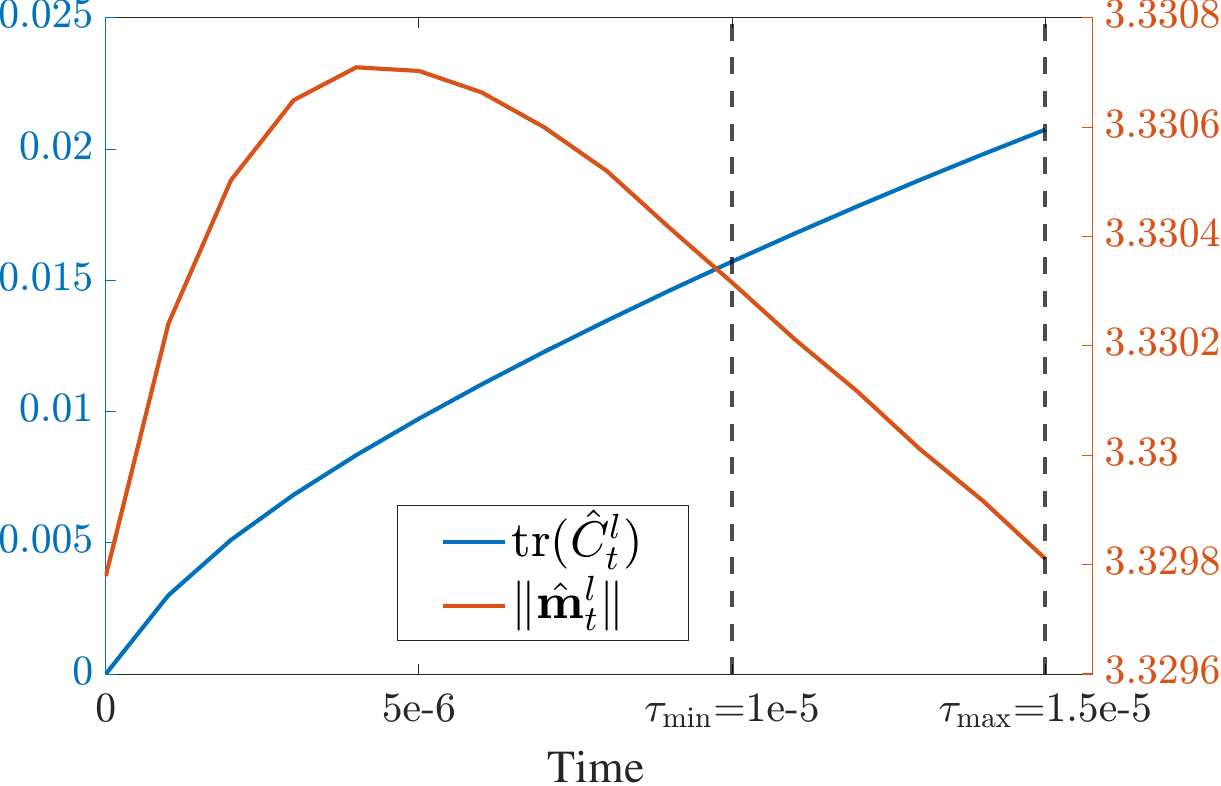}}
{\includegraphics[width=0.32\textwidth]{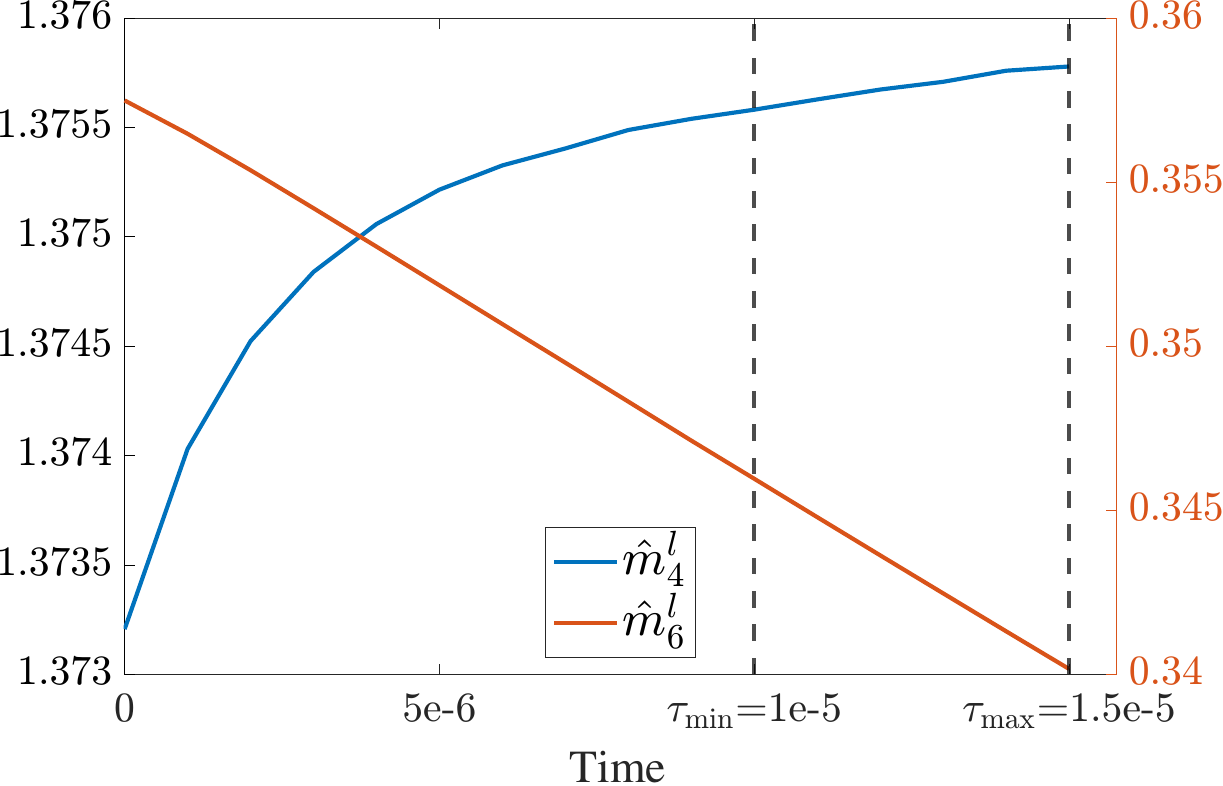}}
{\includegraphics[width=0.32\textwidth]{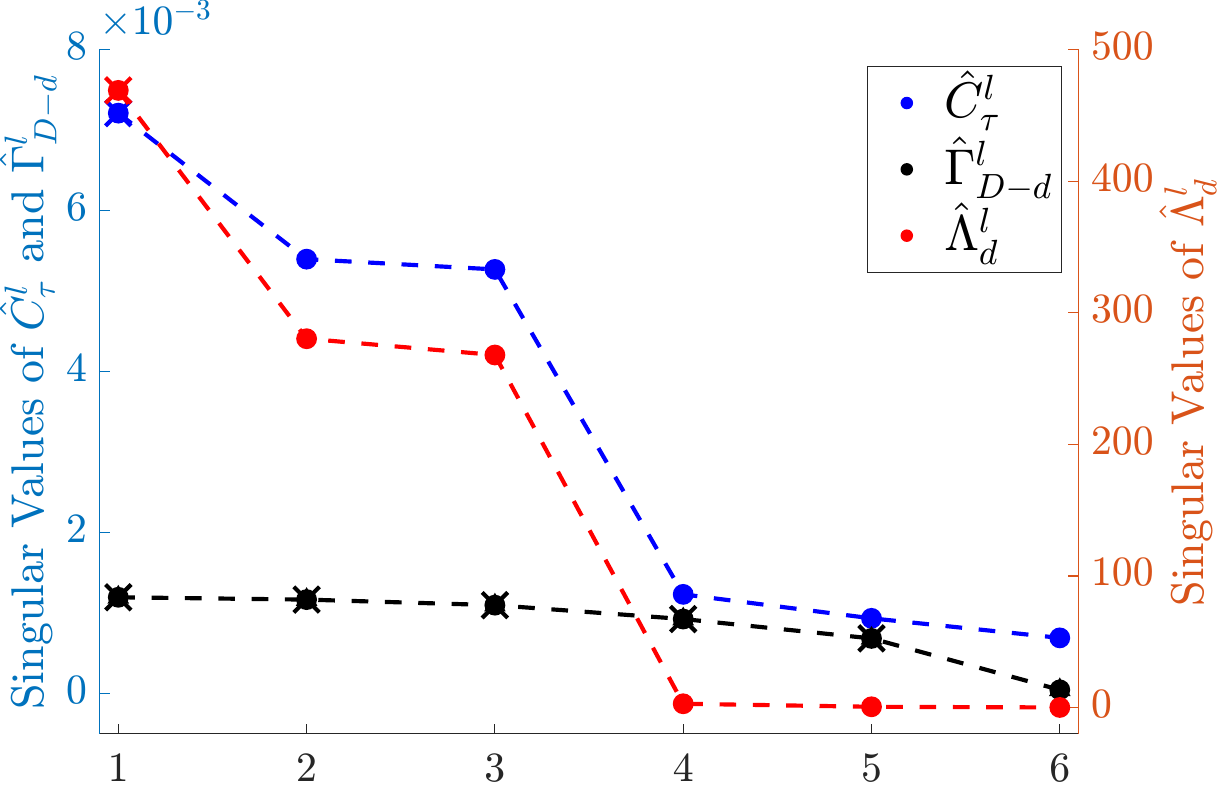}}
{\includegraphics[width=0.33\textwidth, height=0.203\textwidth]{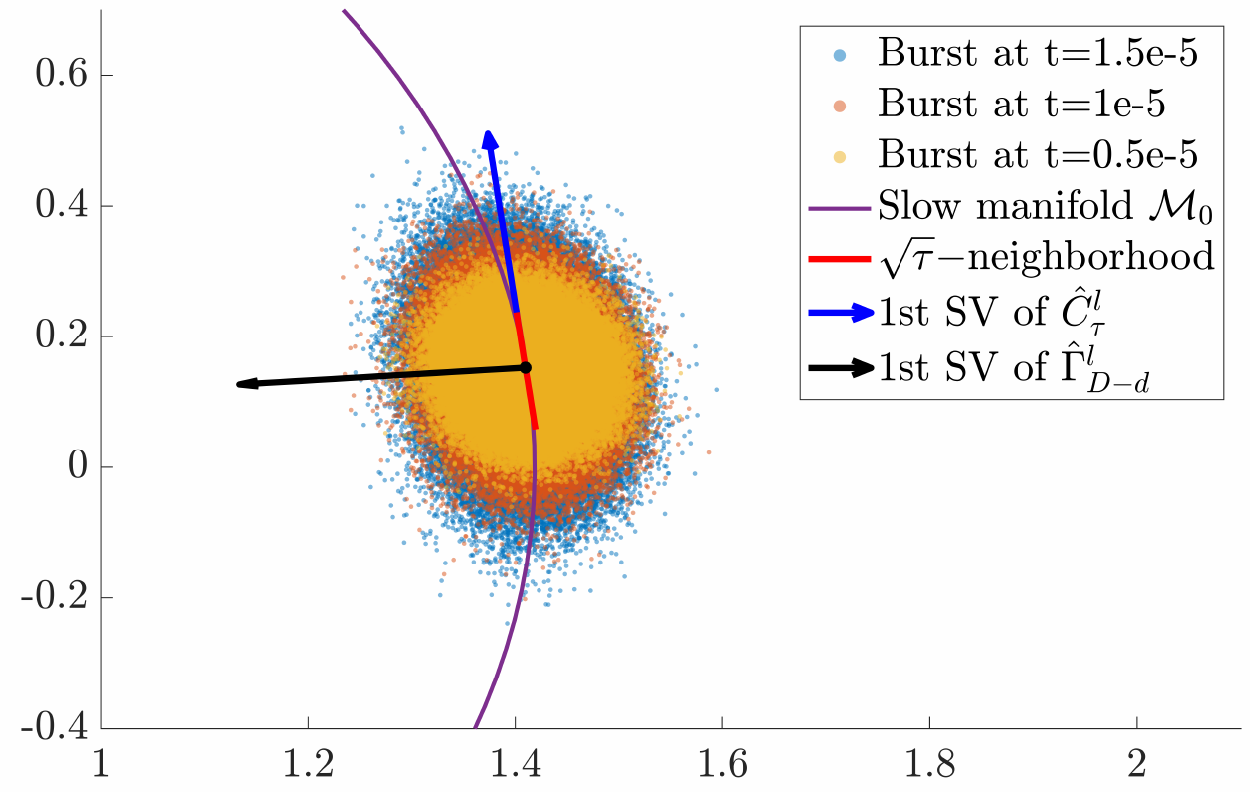}}
{\includegraphics[width=0.33\textwidth, height=0.203\textwidth]{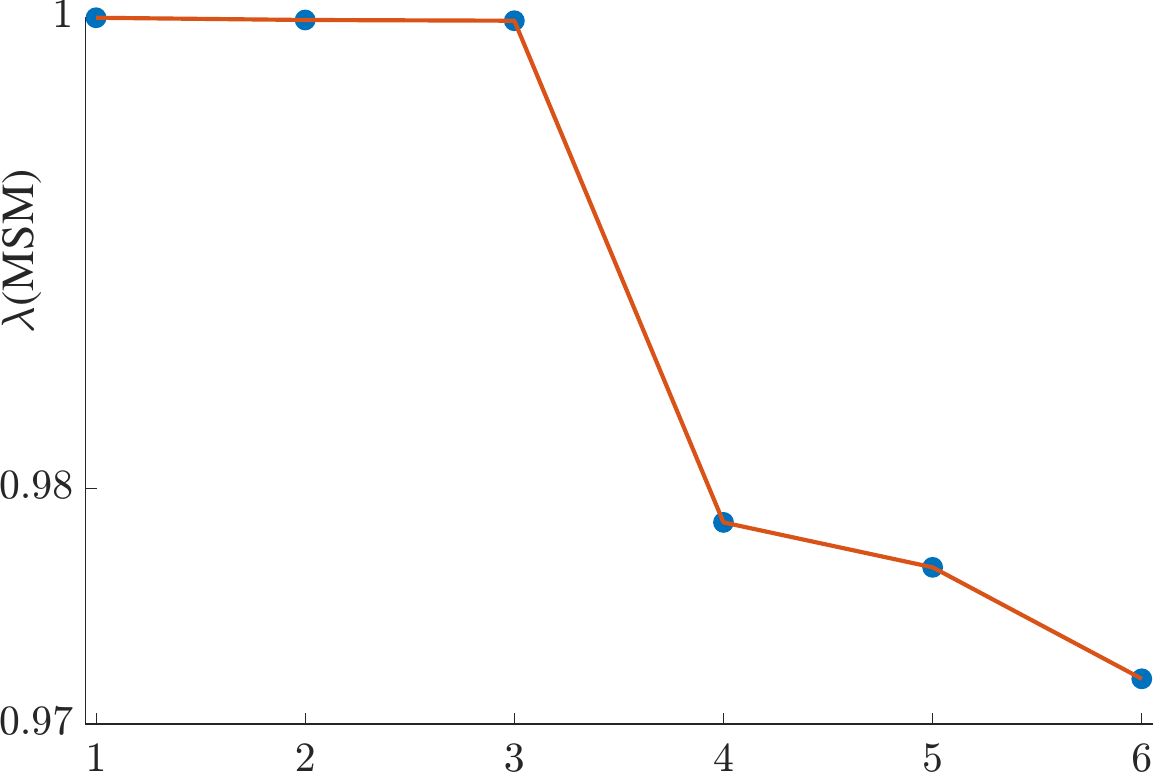}}
\caption{butane: {\bf Top left:}   plot of trace of empirical covariance and norm of empirical mean, $\tr(\hat{C}_t^l)$ and $\|\hat{\mm}_t^l\|$ verse time at one landmark. {\bf Top center:} plot the fourth and sixth coordinates of the empirical mean $ \hat{\mm}_t^l$ verse time at one landmark. 
{\bf Top right:} singular values of $\hat{C}_\tau^l, \hat{\Gamma}^{l}_{D-d}$ and $\LambdaldN$ at one landmark,  the dominant singular values are marked as cross labels. {\bf Bottom left:} scatter of the burst in the $(x_4, z_4)$ coordinate at $t=\num{0.5e-5}, \num{1e-5}, \num{1.5e-5}$, together with  the \SM, $\sqrt{\tau}$-neighborhood and dominant singular vectors of $\hat{C}_\tau^l, \hat{\Gamma}^{l}_{D-d}$.   {\bf Bottom right:}   top six eigenvalues of the Markov transition matrix. }
\label{f:butane}
\end{figure}

\noindent{\textbf{Estimating the relevant time scale $\tau$}}.
 Based on the strength of the bond angle and the largest parameter in the torsion potential, the contribution from the bond and bond angle part will be quickly relaxed and  the scale of separation is approximately a factor of 20, which is not very large, both in absolute terms and when compared to the other examples we considered. It is therefore necessary to perform multiple rounds of refinement to ensure the initial condition is close enough to \IM. Initially, we use the time window $[\tau_{\min{}}, \tau_{\max{}}]=[\num{4e-5}, \num{5e-5}]$ to learn the parameters in each landmark and proceeds with rounds of refinement until the relative differences of estimated parameters within 5\% (this resulted in no more than $10$ rounds), as discussed in \cref{s:ATLASconstruction}.
In \cref{f:butane}, both $\tr(\hat{C}_t^l)$ and $\|\hat{\mm}_t^l\|$ reach the linear regime at $t=\num{5e-6}$, as well as the fourth and sixth coordinates of $\hat{\mm}_t^l$, which are TICA coordinates. We thus ran a last round of refinement with the true training interval  $[\num{1e-5}, \num{1.5e-5}]$ and $\tau=\num{1e-5}$.

\noindent{\textbf{Estimating dimension and tangent spaces of \SMmth, and direction of the fast modes}}.
  We report in \cref{f:butane} the behavior of the singular values of $\hat C_\tau^l$, $\hat\Gamma^l_{D-d}$ and $\LambdaldN$ at one landmark in descending order. In this example, the dominant singular vector of covariance matrix $\hat{C}_t^l$ matches with the one of the diffusivity matrix $\LambdaldN$ and the first dominant singular vector of covariance matrix $\hat{\Gamma}^l_{D-d}$ is almost orthogonal with the slow direction. There are 5 dominant singular values for the covariance matrix $\hat{\Gamma}^l_{D-d}$, then it shows the fast variable has 5 dimensions. 
Similar to the oscillating half-moon example, the part of \SM{} that these data cloud covers clearly shows the nonlinear effect. Therefore, $\tau$ is chosen from the lower bound of the training interval and the special procedure to modify the landmark during the last round of refinement is necessary.  {On average, it requires 77 charts to fully describe the invariant manifold. }

\noindent{\textbf{Identifying metastable states, and estimation of large-time properties of the process}}.
From the top six eigenvalues of the transition matrix of a Markov state model constructed from ATLAS, we observe a significant gap between the third and fourth eigenvalues, correctly indicating that the system has three metastable states. 
As in the half-moon example, we express the metastable states in the latent (dihedral, in this case) angle $\phi$: trans:=$\{\phi\in(-\frac{\pi}{3}, \frac{\pi}{3})\}$, top-cis:=$\{\phi\in(\frac{\pi}{3},\pi)\}$ and bot-cis:=$\{\phi\in(-\pi, -\frac{\pi}{3})\}$. We simulate trajectories of time $\num{5e2}$ with the original simulator and with the ATLAS simulator,  and from those we plot both invariant distributions in the dihedral angle $\phi$. The standard bin width is $\sigma\sqrt{\tau}= 0.07$ and we use it for the plot and calculation of the difference of two distributions. 
The $L^1$-norm of the difference of two approximated densities is $ 0.060\pm 0.013$ and $L^2$-norm of the difference is $0.013\pm0.003$. 

\noindent{\textbf{Estimation of residence times}}. 
In the stage of simulating residence time, we generate a single long trajectory of time $\num{5e2}$ by the original simulator and uniformly sample $N_{\text{IC}}$ initial conditions in each metastable state. The boundary of the residence set is the same as the region of the metastable states. In this example, we say the point is in the residence set if its dihedral angle $\phi$ is in the corresponding metastable state. 

\section{Error analysis}\label{appx:error}
The equations of the relative error of Euclidean norm of the estimated drift term $\bA(\mz)$,  estimated diffusivity matrix $\LambdaA(\mz)$, and between the estimated and theoretical tangent space, are defined, respectively, as
\begin{align}
\relerrb = \frac{\|\bA(\mz) - \mb(\mz)\|}{\|\mb(\mz)\|}\,,\,\relerrLambda=\frac{\|\LambdaA_d(\mz) - \Lambda(\mz)\|}{\|\Lambda(\mz)\|} \,,\,\abserrTangle = \arccos\left(\left\|(\Uslowl)^T \Uslowtruel\right\|\right)\,.
\end{align}
For the estimation error of the invariant manifold, as discussed above we compared points generated by long ATLAS trajectories, which lie on $\hatIMmth$ by definition (of $\hatIMmth$) with points on the slow manifold obtained by averaging the equations in the limit as $\eps\rightarrow0$; this corresponds to the following:
\begin{itemize}
\item in the \Peanut example, from the calculations above we let $\abserrIMmth =  \left|r-r^\star(\theta)\right |$. 
\item In the oscillating half-moon example, we use the distance of the point to the \SM, 
$$\abserrIMmth = \sqrt{(\sqrt{z_1^2+z_2^2}-\bar{r})^2+\sum_{i=3}^{20}(z_i-1)^2}$$
\item In the butane example, we use the distance of the point to the \SM,
$$\abserrIMmth=\sqrt{ \left(x_1+ l\sin(\theta_{eq})\right)^2 + \left(y_1 - l\cos(\theta_{eq})\right)^2 +\left(y_3- l \right)^2 +\left(y_4+  l\cos(\theta_{eq})- l \right)^2 +\left(\sqrt{x_4^2 + z_4^2 } -  l\sin(\theta_{eq})\right)^2}.$$
\end{itemize}

\bibliographystyle{alpha}
\let\oldbibliography\thebibliography
\renewcommand{\thebibliography}[1]{%
  \oldbibliography{#1}%
  \setlength{\itemsep}{0pt}%
}
\bibliography{MasterA,MM_Publications}

\end{document}